%% file: main.tex
\documentclass[11pt,letterpaper]{article}
\usepackage{cogsys}
\usepackage[T1]{fontenc}
\usepackage{times}
\usepackage[pdftex]{graphicx} 
\usepackage{listings}
\usepackage{latexsym}
\usepackage{multirow}
\usepackage{amssymb}
\usepackage{amsmath}
\usepackage{amsthm} 
\usepackage{booktabs}
\usepackage{enumitem}
\usepackage{graphicx}
\usepackage{color}
\usepackage{subcaption}
\usepackage{newfloat}
\usepackage{listings}
\usepackage{algorithm}
\usepackage{tabularx}
\usepackage{xcolor}
\usepackage{inconsolata} 
\usepackage{svg}
\usepackage{tcolorbox}
\definecolor{brown}{rgb}{0.6, 0.4, 0.2}
\usepackage{algpseudocode}

\usepackage{natbib}
\cogsysheading{Twelfth}{2025}{1-18}{X/20XX}{X/20XX}

\ShortHeadings{Formatting Instructions}
              {V.\ Khandelwal, V.\ Pallagani, B.\ Srivastava, and F.\ Rossi}

\begin{document} 

\title{A Neurosymbolic Fast and Slow Architecture for Graph Coloring}
 
\author{Vedant Khandelwal\textsuperscript{1}}{vedant@email.sc.edu}
\author{Vishal Pallagani\textsuperscript{1}}{vishalp@mailbox.sc.edu}
\author{Biplav Srivastava\textsuperscript{1}}{biplav.s@sc.edu}
         \author{Francesca Rossi\textsuperscript{2}}{francesca.rossi2@ibm.com}

         \address{\textsuperscript{1}University of South Carolina, Columbia, SC, USA.}
\address{\textsuperscript{2}IBM Research, NY, USA.} 
\begin{abstract}
Constraint Satisfaction Problems (CSPs) present significant challenges to artificial intelligence due to their intricate constraints and the necessity for precise solutions. Existing symbolic solvers are often slow, and prior research has shown that Large Language Models (LLMs) alone struggle with CSPs because of their complexity. To bridge this gap, we build upon the existing SOFAI architecture (SOFAI-v1), which adapts Daniel Kahneman’s ``Thinking, Fast and Slow'' cognitive model to AI. Our enhanced architecture, SOFAI-v2, integrates refined metacognitive governance mechanisms to improve adaptability across complex domains, specifically tailored here for solving the graph coloring problem, a specific type of CSP. SOFAI-v2 combines a fast System 1 (S1), leveraging LLMs, with a deliberative System 2 (S2), governed by a metacognition module. S1's initial solutions, often limited by constraint adherence issues, are improved through targeted feedback and examples from metacognition, aligning S1 more closely with CSP requirements. If S1 fails to resolve the problem, metacognition strategically invokes S2, ensuring accurate and reliable solutions. Our empirical results demonstrate that SOFAI-v2 achieves a 10.5\% higher success rate and is up to 30\% faster than a traditional symbolic solver in solving graph coloring problems.
\end{abstract}

\input{content_files/introduction}
\input{content_files/background}
\input{content_files/related_works}

\input{content_files/methods}
\input{content_files/experimental_setup}
\input{content_files/results}
\input{content_files/conclusion}

\vspace{-0.25in}

{\parindent -10pt\leftskip 10pt\noindent
\bibliographystyle{cogsysapa}
\bibliography{format}

}

\newpage
\input{content_files/appendix}

\end{document}

%% file: content_files/introduction.tex
\section{Introduction}

Constraint Satisfaction Problems are a core challenge in artificial intelligence (AI) due to their demand for correctness adhering to strict constraints in static \citep{csp-tutorial-Kumar-1992} or uncertain, dynamic environments \citep{csp-tutorial-uncertain-dynamic}. 
These problems are prominent in applications such as scheduling, boolean satisfiability, resource allocation, temporal reasoning, and planning \citep{welsh1967upper, chaitin1982register}. 
Traditional approaches to solve CSPs typically rely on constraint propagation or search \citep{haralick1980increasing}. The symbolic approaches are known for their accuracy but often struggle with scalability as the complexity of CSPs increases, 
 hindered by high computational overhead and slow processing times \citep{dechter2003constraint, rossi2006handbook}. 

Recent advancements in LLMs have shown promise in rapidly processing complex information \citep{ruoss2024grandmaster}, but they are fundamentally limited in solving sequential decision-making tasks like planning \citep{valmeekam2022large}, and reasoning  \citep{stechly2024self}. Due to their probabilistic, retrieval-based nature, LLMs approximate solutions rather than generating definitive answers, often leading to partial or inconsistent adherence to constraints \citep{jiang2023followbench}. Even with state-of-the-art prompting strategies and self-verification techniques, LLMs fall short in handling strict requirements of CSPs, making their outputs unreliable for applications demanding high precision. We study feasibility under fixed resource limits in the graph coloring decision problem, a core constraint satisfaction setting that supports clear verification and iterative guidance.

The disparity between the precision of traditional symbolic methods and the adaptability and speed of LLMs underscores the need for a new paradigm in tackling complex tasks like graph coloring \citep{jensen2011graph}, a canonical CSP. Drawing inspiration from Daniel Kahneman's cognitive theory of ``Thinking, Fast and Slow'' \citep{kahneman2011thinking} and its adaptation to AI as SOFAI-v1 \citep{fabiano2025thinking, booch2021thinking}, we introduce SOFAI-v2 which extends the earlier architecture by refining metacognitive mechanisms that oversee a fast, experience-based S1, powered by an LLM, and a slow, deliberative S2 using the DSATUR algorithm for graph coloring. This governance mechanism continuously monitors and enhances S1’s outputs, offering targeted feedback and constraint adherence examples; when S1’s solutions fall short, S2 is strategically invoked to guarantee accuracy. In this setup, S1 swiftly produces initial solutions, while S2 provides a reliable fallback for achieving precise constraint satisfaction. This work studies problem solving with a fast S1, a deliberate S2, and an MC that monitors and guides iterations, with memory for reuse. Beyond the S1/S2 framing, SOFAI-v2 is positioned as a \emph{neurosymbolic} cognitive architecture: S1 (neural) proposes, symbolic constraints verify, and MC enacts control. This aligns with computational accounts of automatic versus controlled processing and supervisory control (e.g., contention scheduling and supervisory attention) \citep{norman1986attention, cooper2000contention}, providing a bridge from cognitive theory to an AI system.

The implemented system targets the graph coloring decision task and evaluates success rate, time, and iteration behavior. Our contributions are as follows: 
\begin{itemize}
    \item Introduced SOFAI-v2, a neurosymbolic fast and slow architecture to enhance the performance of CSP solvers on graph coloring problems through adaptive metacognitive governance.
    \item Created a comprehensive benchmark of graph coloring problems, featuring variations in graph size, edge probability, and solvable and unsolvable instances to test adaptability.
    \item Conducted extensive empirical evaluations of various solver configurations for graph coloring, demonstrating advantages of SOFAI-v2 in terms of success rate, time efficiency.
\end{itemize}

The rest of the paper is organized as follows: we first provide the necessary background and discuss prior work. Next, we present the SOFAI-v2 architecture, followed by the experimental setup and results. Finally, we conclude with a summary of our findings.

%% file: content_files/background.tex
\section{Background}

\subsection{Graph Coloring Decision Problem}

The graph coloring decision problem is a specific type of constraint satisfaction problem (CSP), where the objective is to determine whether a given graph can be colored using at most \( k \) colors such that no two adjacent nodes share the same color. Formally, the problem is defined as follows:

\begin{itemize}
    \item An undirected, unweighted graph \( G = (V, E) \)
    \item A positive integer \( k \) representing the maximum number of colors
    \item Determine whether there exists a function \( f: V \rightarrow \{1, 2, \dots, k\} \) such that for every edge \( (u, v) \in E \), \( f(u) \neq f(v) \).
\end{itemize}

Unlike the classical graph coloring problem, which aims to minimize the number of colors used (an optimization problem), the decision version focuses solely on satisfiability: whether a valid coloring exists for a given \( k \). We use the DIMACS representation \citep{johnson1996cliques} to describe the graph coloring problem. The decision problem asks a clear yes/no question for a fixed $k$. 
This matches many practical settings where a fixed resource limit is given and the goal is feasibility, e.g., schedules with a fixed number of time slots, 
register allocation with a fixed number of registers, or frequency assignment with a fixed set of bands. 
It provides MC with a simple accept/reject signal and makes it straightforward to generate targeted feedback when constraints fail, which we interpret as a cue for supervisory control in the architecture \citep{norman1986attention, cooper2000contention}.

\subsection{Degree of Saturation Algorithm with Backtracking}

The Degree of Saturation (DSATUR) algorithm assigns colors to graph vertices based on their saturation degree, prioritizing vertices that are the most constrained—those having adjacent vertices with the greatest variety of colors already assigned \citep{brelaz1979new}. DSATUR remains competitive for graph coloring due to its heuristic efficiency, especially for sparse and mid-sized graphs, and its direct operation on graph structures without requiring transformations typical of SAT solvers. Recent studies \citep{yekezare2024optimality, schidler2023sat} highlight its effectiveness, making it a suitable choice for our symbolic solver. However, the traditional DSATUR algorithm targets the optimization version of the graph coloring problem, minimizing the number of colors. For our work, we adapt DSATUR into a satisfiability solver by constraining it with a fixed maximum number of colors. A detailed description of the modified algorithm is provided in Supplementary Material Section \ref{appendix:dsatur}. Unlike other solvers, DSATUR can be straightforwardly modified for the decision version of the problem, making it a practical and adaptable choice for our symbolic backend. 
In SOFAI-v2, DSATUR with backtracking plays the role of slow, controlled processing that is selectively invoked by MC under conflict or low confidence, consistent with supervisory control accounts in cognitive systems \citep{norman1986attention, cooper2000contention}.

%% file: content_files/related_works.tex
\section{Related Works}

\begin{figure*}[ht!]
    \centering
    \includegraphics[width=0.80\linewidth]{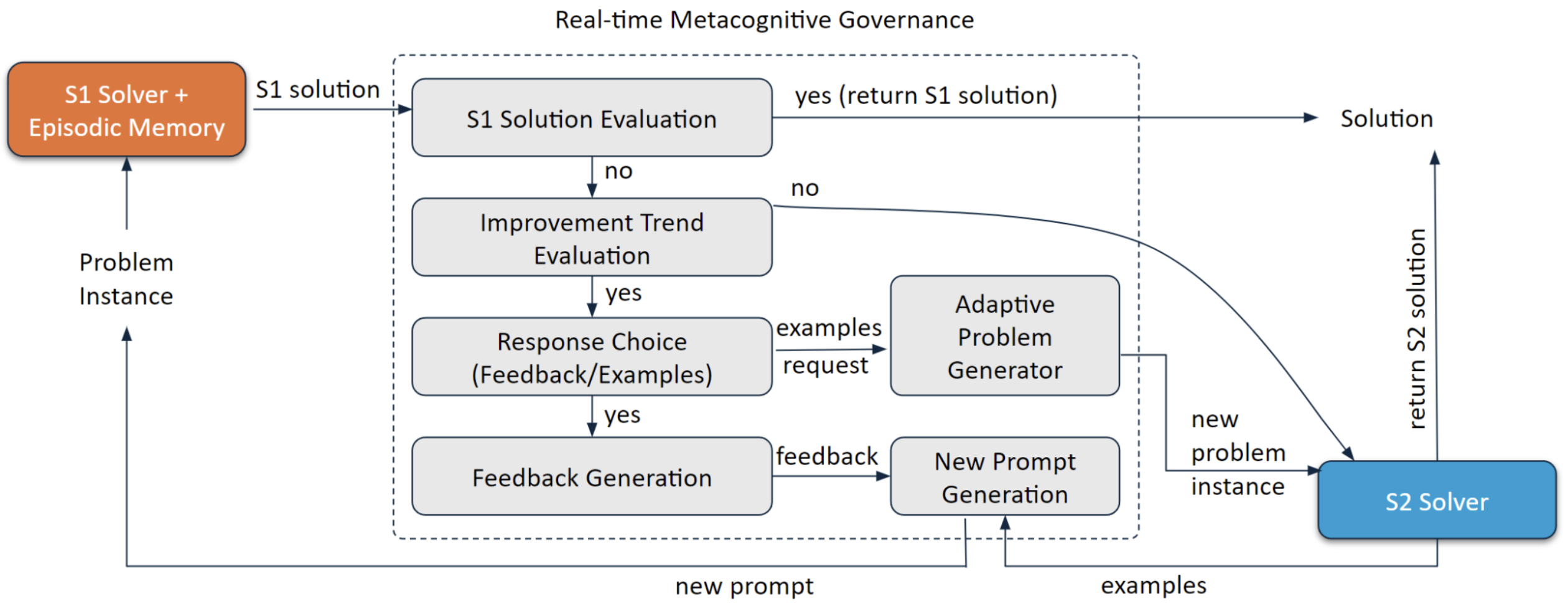}
    \caption{SOFAI-v2 architecture with real-time metacognitive governance (supervisory control over fast/automatic and slow/controlled processing)}
    \label{fig:sofai-arch}
\end{figure*}

This section reviews the evolution of solvers for graph coloring, the application of LLMs in sequential decision-making tasks, and the emergence of neurosymbolic approaches for enhancing decision-making processes. \textit{We also situate SOFAI-v2 within cognitive control theories that distinguish routine (automatic) processing from controlled, supervised processing \citep{norman1986attention, cooper2000contention}.}

\subsection{Solvers for Graph Coloring}
Traditional symbolic solvers have been extensively developed for this problem, leveraging algorithms that systematically explore the search space. Early works like \citep{brelaz1979new} introduced efficient heuristics such as the DSATUR algorithm, significantly improving practical performance. \citep{chaitin1982register} applied graph coloring to compiler optimization, demonstrating the versatility of symbolic methods. \citep{golumbic2004algorithmic} provided a comprehensive treatment of graph algorithms, including coloring techniques. Despite their accuracy, symbolic approaches often struggle with scalability as CSP complexity increases, hindered by high computational overhead and slow processing times \citep{dechter2003constraint}. Techniques like constraint propagation \citep{mackworth1977consistency} and backtracking algorithms \citep{bitner1975backtrack} have been employed to enhance efficiency but are limited by the exponential growth of the search space.

 DSATUR selects at each step the uncolored vertex with the highest \emph{saturation degree} (the number of distinct colors among its colored neighbors), breaking ties by degree \citep{brelaz1979new}. This ordering aggressively focuses search on the most constrained vertices, yielding strong practical performance. With appropriate data structures, per-step selection can be done in near 
$O(logn)$ update time, but the overall search with backtracking remains exponential in the worst case due to the NP-completeness of graph coloring. In our setting, we couple DSATUR with backtracking under a fixed color budget 
$k$: the procedure attempts to extend a partial assignment using at most 
$k$ colors; failure to extend yields a certified \textsc{UNSAT} outcome, while success produces a constructive 
$k$-coloring witness.

To address these limitations, neural solvers have been explored, utilizing machine learning techniques to approximate solutions for graph coloring. For instance, \citep{li2018combinatorial} proposed using graph neural networks to tackle combinatorial optimization problems. Similarly, \citep{khalil2017learning} introduced a framework that learns heuristics directly from data, improving scalability. However, neural approaches often face challenges in strictly adhering to the constraints inherent in CSPs. In contrast, our SOFAI setup directly targets the \emph{decision} form of graph coloring with an explicit 
$k$, integrating DSATUR+backtracking as S2 and using S1/SOFAI components to propose assignments that are subsequently verified. Many end-to-end neural solvers are not readily adaptable to this satisfiability variant, which limits their comparability to our setting; we therefore retain them here primarily for historical context and contrast, while focusing our technical discussion on DSATUR and its role in our pipeline.

\subsection{LLMs for Reasoning Tasks}
Recent developments in LLMs have opened new avenues for addressing sequential decision-making tasks, including CSPs. Various prompt engineering techniques have been proposed to enhance the reasoning capabilities of LLMs \citep{brown2020language, wei2022chain, kojima2022large, nye2021show, wang2022self}. \citep{han2023pive} demonstrated that LLMs can arrive at correct answers for graph-based problems through iterative prompting. However, the reasoning capabilities of LLMs have been found to be fundamentally approximate retrieval in nature \citet{kambhampati2024can}, limiting their effectiveness in precise reasoning tasks. This limitation is evidenced by studies showing LLMs' inability to solve planning problems \citep{valmeekam2022large} and graph coloring problems \citep{stechly2024self}, where strict adherence to constraints is required \citep{jiang2023followbench}. \textit{In cognitive-systems terms, LLM-only routines resemble contention among habitual schemas without sufficient supervisory control \citep{norman1986attention}.}

\subsection{Neurosymbolic Approaches for Reasoning Tasks}
Neurosymbolic approaches have gained considerable attention in addressing complex sequential decision-making tasks such as automated planning \citep{fabiano2023plan}, constrained grid navigation \citep{ganapini2022combining}, and puzzle solving \citep{lin2024swiftsage}. These studies demonstrate that integrating neural methods, such as LLMs, with symbolic solvers enhances performance on sequential decision-making tasks. To the best of our knowledge, neurosymbolic techniques have not been explored in the context of graph coloring problems. To our knowledge, prior neurosymbolic studies have not targeted the graph coloring decision problem, nor integrated meta-level feedback with a decision-oriented S2; SOFAI-v2 explicitly addresses this gap. \textit{Conceptually, SOFAI-v2 instantiates a neurosymbolic cognitive architecture in which S1 (neural) handles routine proposals while MC+S2 provide supervisory/controlled processing aligned with Norman--Shallice-style accounts \citep{norman1986attention, cooper2000contention}.}

%% file: content_files/methods.tex
\section{The SOFAI Architecture} \label{sec:sofai_arch}

The SOFAI architecture, also referred to as SOFAI-v1 here \citep{booch2021thinking}, leverages a rule-based metacognitive control mechanism 
that dynamically chooses between S1 and S2 solvers based on a confidence threshold associated with S1's outputs. S1 solvers are experience-based (and usually data-driven) solvers, while S2 solvers are deliberative (and usually symbolic and rule-based) solvers. 
SOFAI-v1 has been instantiated to planning \citep{fabiano2023plan} and constrained-grid navigation \citep{ganapini2022combining}, showing improvement on the performance compared to either symbolic or data-driven solvers for the same class of problems.

In SOFAI-v1, the metacognitive component only chooses between S1 and S2 solvers, but does not exploit the possible collaboration between these two kinds of solvers, nor does it provide feedback to the solvers if they cannot solve the given problem instance. 
In this paper we consider a generalized version of this architecture, called SOFAI-v2, shown in Figure \ref{fig:sofai-arch}, where the metacognitive governance (called MC) can provide feedback and examples to a failing S1 solver (that for this paper is an LLM), calling the S1 solver more than once, until either a correct solution is returned or a maximum number of iterations is reached, which triggers the activation of an S2 solver. 
\textit{This implements a neurosymbolic control loop: S1 produces routine proposals (automatic), MC performs verification and trend monitoring (supervisory attention), and S2 executes controlled search when conflicts persist, echoing contention scheduling with a supervisory system \citep{norman1986attention, cooper2000contention}.}

Formally, let \( f_{\text{S1}}(x, M) \) denote the solution produced by S1 for a problem instance \( x \) with access to an episodic memory \( M \). Here, \( M \) consists of previously encountered problem-solution pairs generated by SOFAI-v2, i.e., pairs \( (x_j, f_{\text{SOFAI-v2}}(x_j)) \), where \( f_{\text{SOFAI-v2}}(x_j) \) represents the final solution that could have been generated by either S1 or S2. In the context of this paper, \( x \) corresponds to a graph coloring decision problem instance defined by a graph \( G \) and a color bound \( k \).

When a new problem instance \( x \) is presented to S1, a similarity function \( \sigma(x, M) \) is used to retrieve a subset \( M_x \subset M \) of similar past instances, defined as:
\[
M_x = \{ (x_j, f_{\text{SOFAI-v2}}(x_j)) \mid \sigma(x, x_j) \geq \alpha \},
\]
where \( x_j \in M \) and \( \alpha \) is a predefined similarity threshold. This subset, \( M_x \), provides S1 with additional contextual information from past solutions, enabling it to leverage SOFAI-v2’s previous outputs for experience-guided context.

The correctness of solution for the graph coloring decision problem can be checked and evaluated in polynomial time by MC. The correctness score used by MC for graph coloring solutions is:
\[
C(f_{S1}(x, M_x)) = \frac{\sum_{(v,u)\in E} \mathbf{1}[f(v)\neq f(u)]}{|E|}
\]

where:
\begin{itemize}
    \item \( f(v) \) denotes the assigned color of vertex \( v \),
    \item \( E \) denotes the set of edges in the graph.
\end{itemize}

 If this correctness meets or exceeds a threshold \( \theta_{\text{S1}} \), solution is accepted. Otherwise, MC iteratively provides structured feedback \(\mathcal{F}(f_{S1}(x, M_x))\) or simplified examples \(\mathcal{E}(x)\). We set the correctness threshold to 1, since 1 indicates full constraint adherence for the graph-coloring decision task; this ensures S1 outputs are only accepted when a complete, validator-checked coloring is produced. MC also monitors S1’s solution correctness across iterative feedbacks, identifying improvement trends defined by:
\[
\forall i \in \{1, \dots, m-1\}, \quad C_{i+1} > C_i
\]

Failure to show improvement within \( m \) iterations results in MC invoking the fallback S2 solver, ensuring robustness. Thus, the iterative feedback mechanism and episodic memory advances SOFAI-v2 over SOFAI-v1, enhancing solver effectiveness via structured feedback and experience-based learning. Our theoretical commitments are: (i) dual-solver with metacognitive governance, (ii) episodic memory for similarity-based retrieval, and (iii) verify-and-feedback loops grounded in polynomial-time checks. 
The concrete instantiation (Mistral-7B as S1, DSATUR with backtracking as S2) is an implementation detail; same principles transfer to other domains. Detailed algorithm for the architecture is in Supplementary Material Section~\ref{appendix:algosfv2}.

%% file: content_files/experimental_setup.tex
\section{Experimental Setup and Results}\label{sec:expsetup}

This section details methodology and experimental results evaluating various solver configurations on graph coloring problems, emphasizing efficiency and success rate under systematic constraints. \textit{We also report how iterative feedback and selective escalation can be interpreted as supervisory control improving routine policy performance under increasing constraint conflict \citep{norman1986attention}.}

\subsection{Data Generation and Problem Classification}
Graph coloring instances were generated using the Erdős–Rényi model, controlling problem complexity through three primary parameters:

\noindent\textbf{Graph Size ($n$):} Varies between $n \in [5, 50]$ vertices, encompassing a diverse spectrum of problem sizes.

\noindent\textbf{Edge Probability ($p$):} Adjusted within $p \in [0.1, 0.9]$, influencing edge density and problem difficulty.

\noindent\textbf{Solvability Mix ($m$):} The chromatic number ($\chi(G)$) of each generated graph was computed using the DSATUR algorithm. Problems were then categorized based on solvability, represented as \( m = (a,b) \), where \(a\) and \(b\) denote percentages of solvable and unsolvable instances respectively:

\begin{itemize}
    \item \( m = (100, 0) \): All problems solvable within constraints.
    \item \( m = (50, 50) \): Equal distribution of solvable and unsolvable problems.
    \item \( m = (0, 100) \): All problems unsolvable; set by choosing the number of available colors \( k < \chi(G) \), ensuring infeasibility.
\end{itemize}

For each graph size and edge probability pair 100 graph problems were generated and stored in standardized DIMACS format for reproducibility. \textit{Github Repository: \url{https://github.com/khvedant02/CSP-SOFAI_v2} }.

\subsection{SOFAI-v2 Implementation for Graph Coloring}
SOFAI-v2 integrates iterative feedback mechanisms and episodic memory to dynamically resolve graph coloring problems. Figures \ref{fig:graph_coloring_prompt} and \ref{fig:episodic_memory_prompt} illustrate example prompts.

\begin{itemize}
    \item \textbf{Episodic Memory (\(M\)):} Stores historical graph instances with attributes—vertices, edges, edge density, chromatic number—and solutions. A similarity function \( \sigma(x, M) \) retrieves relevant past solutions:
    \[
    M_x = \{ (x_j, f_{\text{v2}}(x_j)) : x_j \in M, \, d(x, x_j) < \alpha \}
    \]
    where \(d(x, x_j)=|attributes(x)-attributes(x_j)|\), and \(\alpha\) is a similarity threshold.

    \item \textbf{Iterative Feedback Mechanism:} If the solver's correctness score (defined in Section \ref{sec:sofai_arch}) is insufficient, a two-step feedback process occurs:
    \begin{enumerate}
        \item \textit{Detect Conflict Pairs}: Identify adjacent vertices assigned identical colors.
        \item \textit{Template-Based Feedback}: Algorithmically generate structured corrections (e.g., ``Error: Vertices A and B are adjacent but have the same color''). Example of feedbacks is given in Supplementary Material Section \ref{appendix:inout}.
    \end{enumerate}

    \item \textbf{Improvement Evaluation and S2 Invocation:} If no solution improvement occurs over five iterations or if correctness remains below threshold, S2 is invoked as a fallback. If either solver identifies an instance as unsolvable, the instance is labeled as UNSAT. Exceeding the strict 200-second per-instance time limit results in failure, regardless of actual solvability.
    
    \item \textbf{Example Generation (\(\mathcal{E}(x)\)):} A greedy-based algorithm generates simpler subgraph instances to guide S1. Examples are detailed further in Supplementary Material Section \ref{appendix:inout}.
\end{itemize}

This comprehensive integration ensures SOFAI-v2 leverages both real-time metacognitive governance and experience-guided context for enhanced problem-solving effectiveness. \textit{Operationally, this mirrors a supervisory system that monitors conflict and recruits controlled processing when routine proposals fail \citep{norman1986attention, cooper2000contention}.}

\begin{figure}[h!]
\centering
\begin{tcolorbox}[
    colframe=black!0, colback=gray!5, coltitle=black,
    boxrule=0.5pt, title=Graph Coloring Problem Prompt, fonttitle=\bfseries,
    width=\columnwidth, rounded corners, colbacktitle=gray!20
]
\textbf{New Problem to Solve:}\\
You are given an undirected graph with 2 colors available. Your task is to assign a color to each vertex such that no two adjacent vertices share the same color.\\

\vspace{0.2cm}
\textbf{Graph Representation:}\\
- Number of vertices and edges: \texttt{p edge 5 5}.\\
- Edges between vertices are listed as follows:\\
\texttt{e A B}\\
\texttt{e A C}\\
\texttt{e B C}\\
\texttt{e C D}\\
\texttt{e D E}

\vspace{0.2cm}
\textbf{Objective:}\\
Assign a unique color to each vertex, ensuring that no two vertices connected by an edge have the same color. Use no more than 2 distinct colors. Provide the color assignments for each vertex in the format:\\
(Vertex Color)

\vspace{0.2cm}
\textbf{Example Format:}\\
\texttt{(A 1)}\\
\texttt{(B 2)}\\
\texttt{(C 1)}

\vspace{0.2cm}
Please provide the color assignment for the new problem to solve, or respond with "NOT SOLVABLE" if it cannot be solved.
\end{tcolorbox}
\caption{LLM prompt template for the graph coloring decision problem without using episodic memory.}
\label{fig:graph_coloring_prompt}
\end{figure}

\begin{figure}[h!]
\centering
\begin{tcolorbox}[
    colframe=black!0, colback=gray!5, coltitle=black,
    boxrule=0.5pt, title=Episodic Memory Prompt, fonttitle=\bfseries,
    width=\columnwidth, rounded corners, colbacktitle=gray!20
]

\textbf{Problem:}\\
\texttt{p edge 4 4}\\
\texttt{e X Y}\\
\texttt{e Y Z}\\
\texttt{e Z A}\\
\texttt{e X A}

\vspace{0.2cm}
\textbf{Correct Solution:}\\
\texttt{(X 1)}\\
\texttt{(Y 2)}\\
\texttt{(Z 1)}\\
\texttt{(A 2)}

\vspace{0.2cm}
\textbf{End of Example}
\end{tcolorbox}
\vspace{-1em}
\caption{Example for episodic memory, which is added at the end of LLM prompt template in Supplementary Material Section \ref{appendix:inout}.}
\vspace{+1em}
\label{fig:episodic_memory_prompt}
\end{figure}

\noindent\textbf{SOFAI-v1 vs SOFAI-v2: }SOFAI-v2 introduces substantial advancements over previous works of SOFAI (SOFAI-v1), by incorporating iterative feedback mechanisms and episodic memory, which enhance the problem-solving capabilities of the S1 solver. Unlike SOFAI-v1, where problems not solved by S1 are directly escalated to S2 without iterative refinement, SOFAI-v2 systematically leverages historical data and adaptive feedback to optimize solutions before resorting to S2, thereby improving efficiency and efficacy. Supplementary Material includes exact prompts, feedback templates, and our DSATUR+BT pseudocode.

\subsection{Performance Metrics and Evaluation}
All solver configurations (Table \ref{table2}) were evaluated under a strict 200-second time limit per instance. Solver performance was measured using:

\begin{equation} \label{eq:success_rate}
\text{Success Rate} (\%) = \left(\frac{\text{Number of Correct Solutions}}{\text{Total Number of Problems}}\right) \times 100
\end{equation}

\begin{equation} \label{eq:average_time}
\text{Average Time Taken} = \frac{\sum_{i=1}^{n} t_i}{n}
\end{equation}
where \( t_i \) is the solving time for the \( i \)-th successfully solved instance and \( n \) the total instances solved within the time limit. Instances exceeding this limit are marked as failures, explaining occasional 0\% success rates even when solvable theoretically. Results reported are averaged over three randomized trials for reliability. To better understand solver efficiency beyond strict time constraints, we conducted supplementary experiments without time limits, solely for comparative runtime analysis. \textit{From a cognitive perspective, increased success with iterative MC reflects benefits of supervisory control over routine responding under higher constraint densities \citep{norman1986attention, cooper2000contention}.}

\begin{table}[h!]
\caption{Descriptions of solver configurations employed in experiments.}
\centering
\small
\begin{tabular}{@{}p{1.8cm}p{6.5cm}@{}}
    \toprule
    \textbf{ID} & \textbf{Description} \\
    \midrule
    S1 & Mistral-7B LLM \\
    S2 & DSATUR with Backtracking \\
    SOFAI-v1 & Combines S1 and S2 with rule-based MC \\
    SOFAI-v2 & Combines S1 and S2 with iterative MC \\
    MC-S1-I[1-5] & Metacognitive feedback (1–5 iterations) to refine S1 solutions \\
    \bottomrule
\end{tabular}
\label{table2}
\end{table}

%% file: content_files/results.tex
\subsection{Results}\label{sec:results}
This section presents an analysis of solver performance by systematically controlling key parameters. Unless mentioned, the edge formation probability is fixed to \( p = 0.5 \), allowing a focused evaluation across varying graph sizes \( n \) and solvability mixes \( m \) to capture a broad range of problem configurations. Tables \ref{table:performance_comparison} and \ref{table:time_comparison} provide the success rate and average time taken for each solver configuration under these controlled conditions. The Supplementary Material \ref{appendix:successrate} \& \ref{appendix:averagetime} includes results for additional \( p \) values to supplement this focused analysis, providing a comprehensive view of solver robustness across varying edge densities. From these results, we aim to address four key research questions (RQs) that provide deeper insights into SOFAI-v2 performance and robustness.

\begin{tcolorbox}[
    colframe=black!0, colback=gray!5, coltitle=black,
    boxrule=0.5pt, title=RQ 1, fonttitle=\bfseries,
    width=\columnwidth, rounded corners,
    before skip=10pt, after skip=10pt,
    colbacktitle=gray!20
]
\textit{How does SOFAI-v2 success rate compare to that of S1, S2, SOFAI-v1?}
\end{tcolorbox}

\begin{table}[t]
\centering
\scriptsize
\setlength{\tabcolsep}{2.2pt} 
\renewcommand{\arraystretch}{0.95} 
\caption{Success rates (\%) across solvers and problem configurations for graph sizes from 5 to 50 with edge probability $p=0.5$. Configurations $m=(100,0)$, $m=(0,100)$, and $m=(50,50)$ represent solvable, unsolvable, and a balanced mix of solvable and unsolvable instances, respectively.}
\label{table:performance_comparison}
\resizebox{\linewidth}{!}{%
\begin{tabular}{@{}l *{3}{>{\centering\arraybackslash}p{0.6cm} >{\centering\arraybackslash}p{0.6cm} >{\centering\arraybackslash}p{1.2cm} >{\centering\arraybackslash}p{1.2cm}}@{}}
    \toprule
    \multirow{2}{*}{Graph Size} & \multicolumn{4}{c}{m(100,0) [\%]} & \multicolumn{4}{c}{m(0,100) [\%]} & \multicolumn{4}{c}{m(50,50) [\%]} \\ 
    \cmidrule(lr){2-5} \cmidrule(lr){6-9} \cmidrule(lr){10-13}
    & S1 & S2 & SOFAI\_v1 & SOFAI-v2 & S1 & S2 & SOFAI\_v1 & SOFAI-v2 & S1 & S2 & SOFAI\_v1 & SOFAI-v2 \\ 
    \midrule
    5  & 9.41 & \textbf{100} & \textbf{100} & \textbf{100} & 75 & \textbf{100} & \textbf{100} & \textbf{100} & 45.36 & \textbf{100} & \textbf{100} & \textbf{100} \\ 
    10 & 0 & \textbf{100} & \textbf{100} & \textbf{100} & 60.38 & \textbf{100} & \textbf{100} & \textbf{100} & 39.39 & \textbf{100} & \textbf{100} & \textbf{100} \\ 
    15 & 0 & \textbf{80} & \textbf{80} & \textbf{80} & 37.50 & 77.08 & 89.58 & \textbf{93.75} & 17.65 & 77.45 & 83.33 & \textbf{84.31} \\ 
    20 & 0 & 5 & 5 & \textbf{7.29} & 45.65 & 4.35 & 47.83 & \textbf{76.09} & 22.58 & 3.06 & 24.73 & \textbf{38.95} \\ 
    30 & 0 & 0 & 0 & 0 & 54.17 & 0 & 62.50 & \textbf{72.92} & 28.89 & 0 & 33.71 & \textbf{38.89} \\ 
    40 & 0 & 0 & 0 & 0 & 33.33 & 0 & \textbf{47.37} & \textbf{47.37} & 17.43 & 0 & 24.55 & \textbf{24.77} \\ 
    50 & 0 & 0 & 0 & 0 & 3.85 & 0 & 3.85 & \textbf{53.85} & 2.11 & 0 & 2.11 & \textbf{27.72} \\ 
    \bottomrule
\end{tabular}%
}
\end{table}

\begin{table}[!ht]
\centering
\scriptsize
\setlength{\tabcolsep}{3pt}
\renewcommand{\arraystretch}{0.95}
\caption{Average time (seconds) for SOFAI\_v1, SOFAI\_v2, and S2 across graph sizes and configurations $m(100,0)$, $m(0,100)$, and $m(50,50)$.}
\label{table:time_comparison}
\resizebox{\linewidth}{!}{%
\begin{tabular}{lccc ccc ccc}
    \toprule
    \multirow{2}{*}{Graph Size} 
    & \multicolumn{3}{c}{$m(100,0)$ (s)} 
    & \multicolumn{3}{c}{$m(0,100)$ (s)} 
    & \multicolumn{3}{c}{$m(50,50)$ (s)} \\
    \cmidrule(lr){2-4} \cmidrule(lr){5-7} \cmidrule(lr){8-10}
    & SOFAI\_v1 & SOFAI\_v2 & S2 
    & SOFAI\_v1 & SOFAI\_v2 & S2 
    & SOFAI\_v1 & SOFAI\_v2 & S2 \\
    \midrule
     5  & 2   & \textbf{1}  & 0
        & 2   & \textbf{1}  & 0
        & 2   & \textbf{1}  & 0 \\
    10  & 4   & \textbf{2}  & 0
        & 4   & \textbf{2}  & 0
        & 4   & \textbf{3}  & 0 \\
    15  & 50  & \textbf{40} & 51
        & 32  & \textbf{19} & 59
        & 52  & \textbf{47} & 54 \\
    20  & 87  & \textbf{65} & 82
        & 50  & \textbf{20} & 97
        & 87  & \textbf{84} & 107 \\
    30  & 323 & \textbf{306} & 320
        & 403 & \textbf{299} & 400
        & 353 & \textbf{263} & 350 \\
    40  & \underline{5043.2} & \textbf{4512.3} & 4995.8
        & 5006.9 & \textbf{3511.7} & 4989.4
        & 5012.1 & \textbf{3524.6} & 4992.7 \\
    50  & \underline{20087.3} & \textbf{18034.9} & 19950.5
        & 20044.2 & \textbf{14102.7} & 19920.6
        & \underline{20051.8} & \textbf{14178.3} & 19965.0 \\
    \bottomrule
\end{tabular}%
}
\end{table}

\noindent\textbf{Answer to RQ 1:} 
SOFAI-v2 demonstrates substantial improvements in success rates over other solvers, particularly in challenging problem configurations involving unsolvable or mixed solvability instances. Table \ref{table:performance_comparison} shows that for clearly solvable scenarios ($m = (100, 0)$), SOFAI-v2 matches the optimal performance of SOFAI-v1 and S2 across smaller graph sizes. However, its key advantage becomes apparent in more complex scenarios:

\begin{itemize}
    \item For unsolvable scenarios ($m = (0, 100)$) at graph size 50, SOFAI-v2 achieves a success rate of 53.85\%, far surpassing the performance of SOFAI-v1 and S1, which each attain only 3.85\%. This translates to an improvement of approximately 1298\%.
    
    \item In mixed scenarios ($m = (50, 50)$), SOFAI-v2 maintains a success rate of 27.72\%, substantially higher compared to just 2.11\% by SOFAI-v1 and S1.
\end{itemize}

These performance gains arise primarily from SOFAI-v2’s enhanced metacognitive governance, which systematically leverages iterative solver feedback and episodic memory. This adaptive capability allows SOFAI-v2 to dynamically resolve constraints and effectively handle complex or initially challenging instances that other solvers cannot reliably solve.

For the above results we set a fixed time limit of 200 seconds per problem. If a solver does not finish within this time, it is counted as a failure. The average time taken results for this setting is reported separately in the Supplementary Material Section \ref{appendix:avgtime}.

\begin{tcolorbox}[
    colframe=black!0, colback=gray!5, coltitle=black,
    boxrule=0.5pt, title=RQ 2, fonttitle=\bfseries,
    width=\columnwidth, rounded corners,
    before skip=10pt, after skip=10pt,
    colbacktitle=gray!20
]
\textit{How does SOFAI-v2 compare in terms of average time relative to SOFAI-v1 and S2?}
\end{tcolorbox}

\noindent\textbf{Answer to RQ 2:} 
SOFAI-v2 demonstrates substantial improvements in time efficiency compared to SOFAI\_v1 and S2, especially for complex configurations involving unsolvable or mixed scenarios. As detailed in Table~\ref{table:time_comparison}, in the unsolvable scenario ($m = (0,100)$) at graph size $n=50$, SOFAI-v2 requires approximately 14,000 seconds, whereas both SOFAI\_v1 and S2 take around 20,000 seconds. This corresponds to a substantial 30\% reduction in average solving time. Similarly, for mixed solvability scenarios ($m = (50,50)$) at the same graph size ($n=50$), SOFAI-v2 achieves the same notable reduction of 30\%, requiring only about 14,000 seconds compared to 20,000 seconds for SOFAI\_v1 and S2. In fully solvable scenarios ($m = (100,0)$), the improvements remain consistent. For instance, at graph size 50, SOFAI-v2 completes solving tasks in roughly 18,000 seconds, a 10\% time reduction compared to SOFAI\_v1 and S2 (each requiring about 20,000 seconds).

These time-efficiency gains are attributed to SOFAI-v2's advanced metacognitive governance, which intelligently manages solver interactions and computational resources based on real-time feedback and memory-driven decision-making. This dynamic resource allocation enhances solver efficiency, especially in challenging problem settings.

To ensure a fair and meaningful comparison of solver speed, we conducted this experiment without imposing any time limit. This is appropriate because SOFAI-v1, SOFAI-v2, and S2 are all complete solvers—that is, they can eventually solve or correctly determine the unsolvability of a problem given enough time. In contrast, S1 alone does not satisfy this property, so it is excluded from this timing comparison. This no-time-limit evaluation allows us to isolate and understand the time-efficiency advantages introduced by metacognitive governance in SOFAI-v2.

\begin{tcolorbox}[
    colframe=black!0, colback=gray!5, coltitle=black,
    boxrule=0.5pt, title=RQ 3, fonttitle=\bfseries,
    width=\columnwidth, rounded corners,
    before skip=10pt, after skip=10pt,
    colbacktitle=gray!20
]
\textit{Does increasing the number of iterations in the feedback loop lead to an increased success rate in SOFAI-v2?}
\end{tcolorbox}

\begin{figure*}[ht!]
\centering
\begin{subfigure}[b]{0.32\textwidth}
\includegraphics[width=\textwidth]{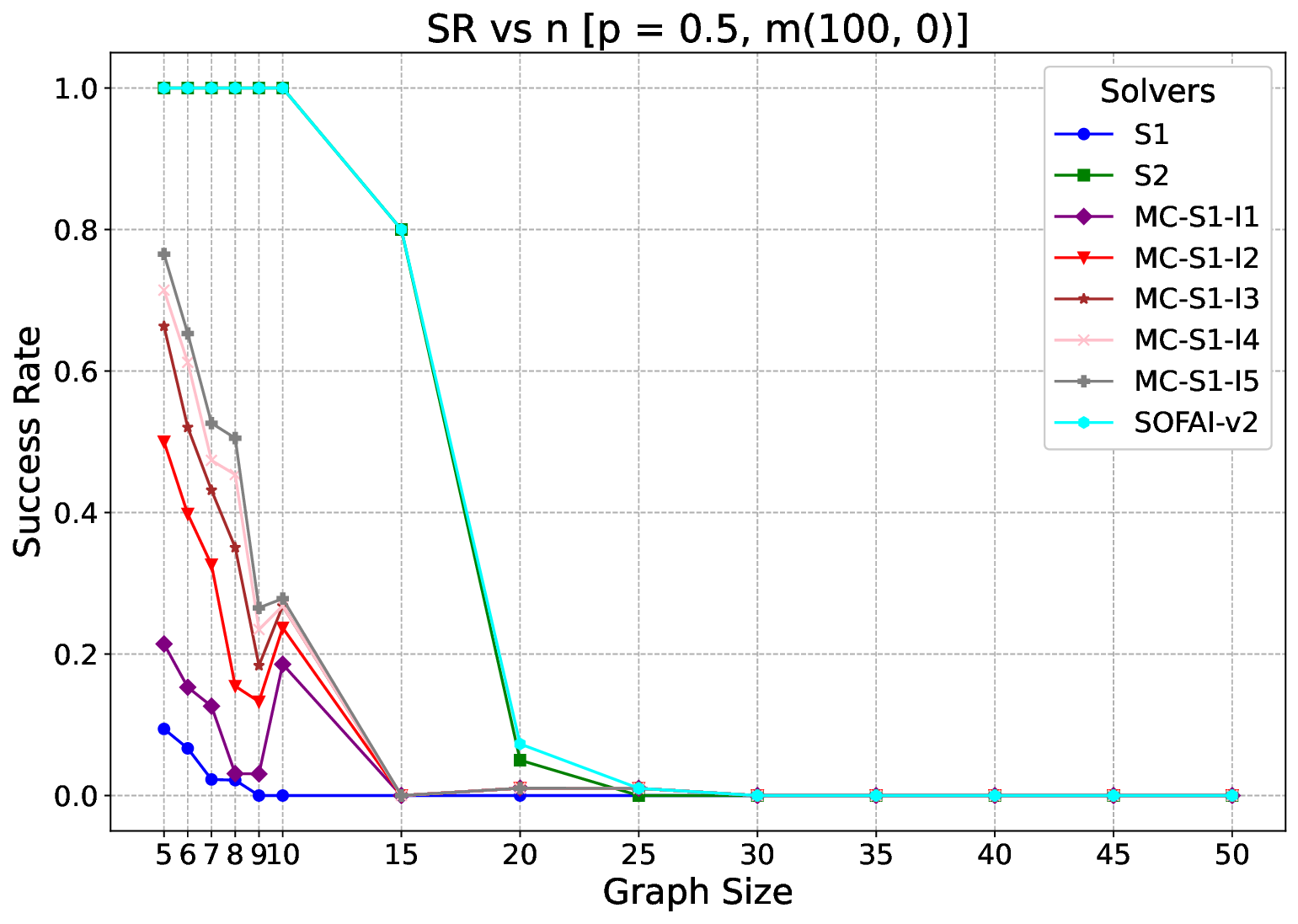}
\caption{All Solvable ($m = (100, 0)$)}
\label{fig:alls}
\end{subfigure}
\hfill
\begin{subfigure}[b]{0.32\textwidth}
\includegraphics[width=\textwidth]{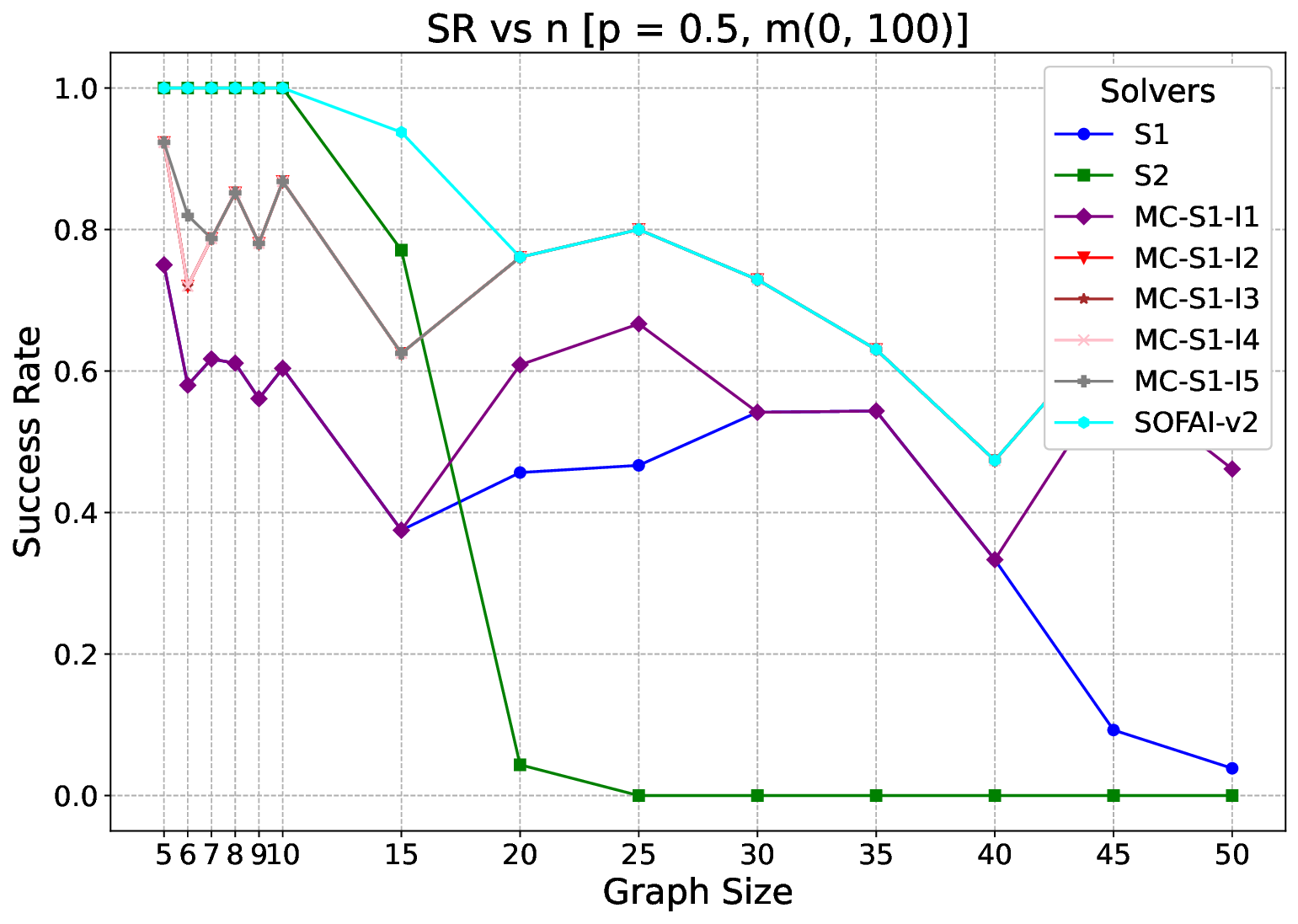}
\caption{All Unsolvable ($m = (0, 100)$)}
\label{fig:allus}
\end{subfigure}
\hfill
\begin{subfigure}[b]{0.32\textwidth}
\includegraphics[width=\textwidth]{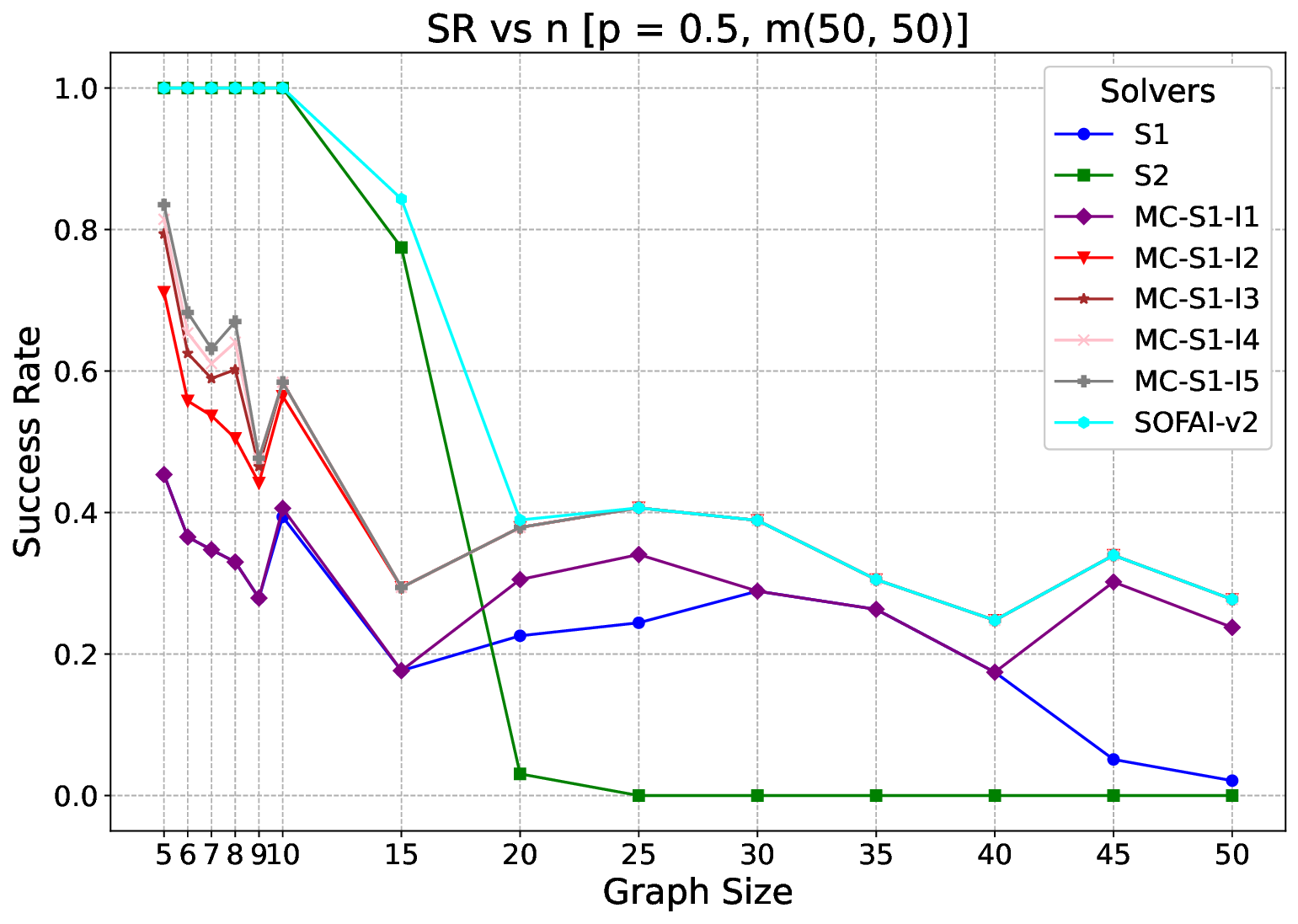}
\caption{Balanced Mix ($m = (50, 50)$)}
\label{fig:allhalf}
\end{subfigure}
\vspace{+1em}
\caption{Success rates of SOFAI-v2’s S1 with iterative metacognitive feedback (MC-S1-I1 to MC-S1-I5) across different graph sizes and problem configurations, showing the impact of iterations on solver performance.}
\label{fig:rq3}
\end{figure*}

\begin{figure*}[ht!]
\centering
\begin{subfigure}[b]{0.29\textwidth}
\includegraphics[width=\textwidth]{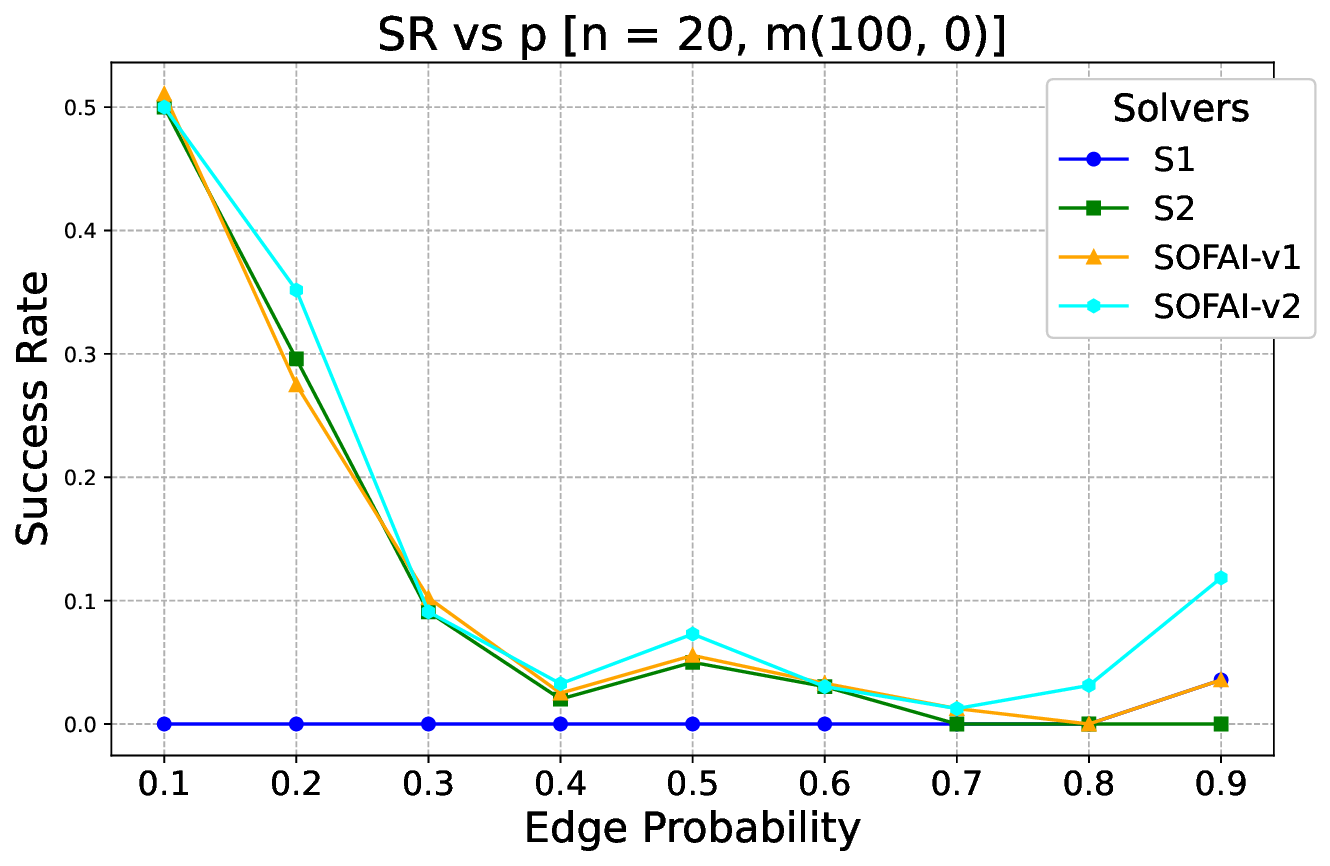}
\caption{All Solvable ($m = (100, 0)$)}
\label{fig:alls_edge}
\end{subfigure}
\hfill
\begin{subfigure}[b]{0.36\textwidth}
\includegraphics[width=\textwidth]{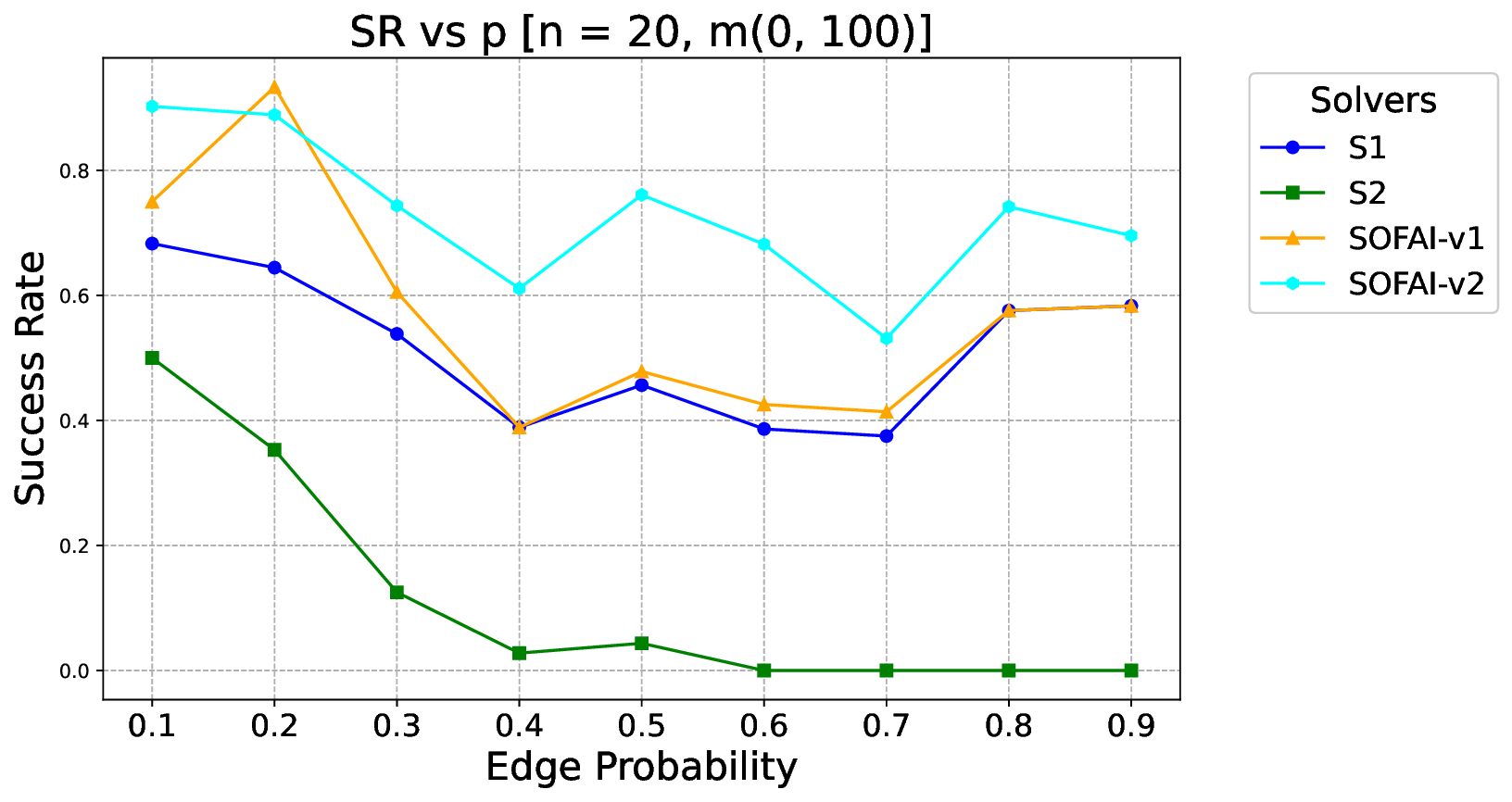}
\caption{All Unsolvable ($m = (0, 100)$)}
\label{fig:allus_edge}
\end{subfigure}
\hfill
\begin{subfigure}[b]{0.29\textwidth}
\includegraphics[width=\textwidth]{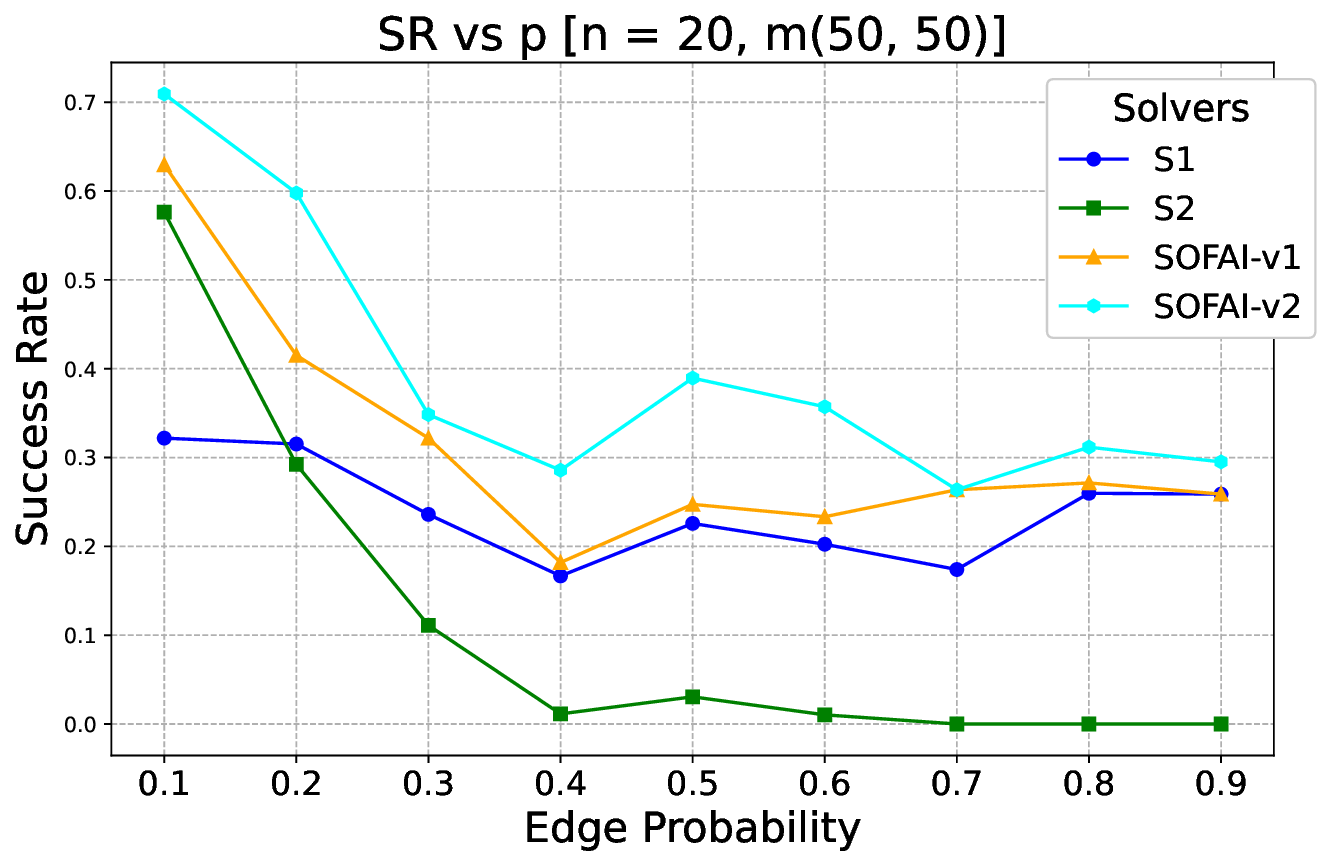}
\caption{Balanced Mix ($m = (50, 50)$)}
\label{fig:allhalf_edge}
\end{subfigure}
\vspace{+1em}
\caption{Success rates across solvers as a function of edge probability in different problem configurations, highlighting the adaptability of SOFAI-v2 compared to traditional solvers.}
\label{fig:rq4_1}
\end{figure*}

\begin{figure*}[ht!]
\centering
\begin{subfigure}[b]{0.32\textwidth}
\includegraphics[width=\textwidth]{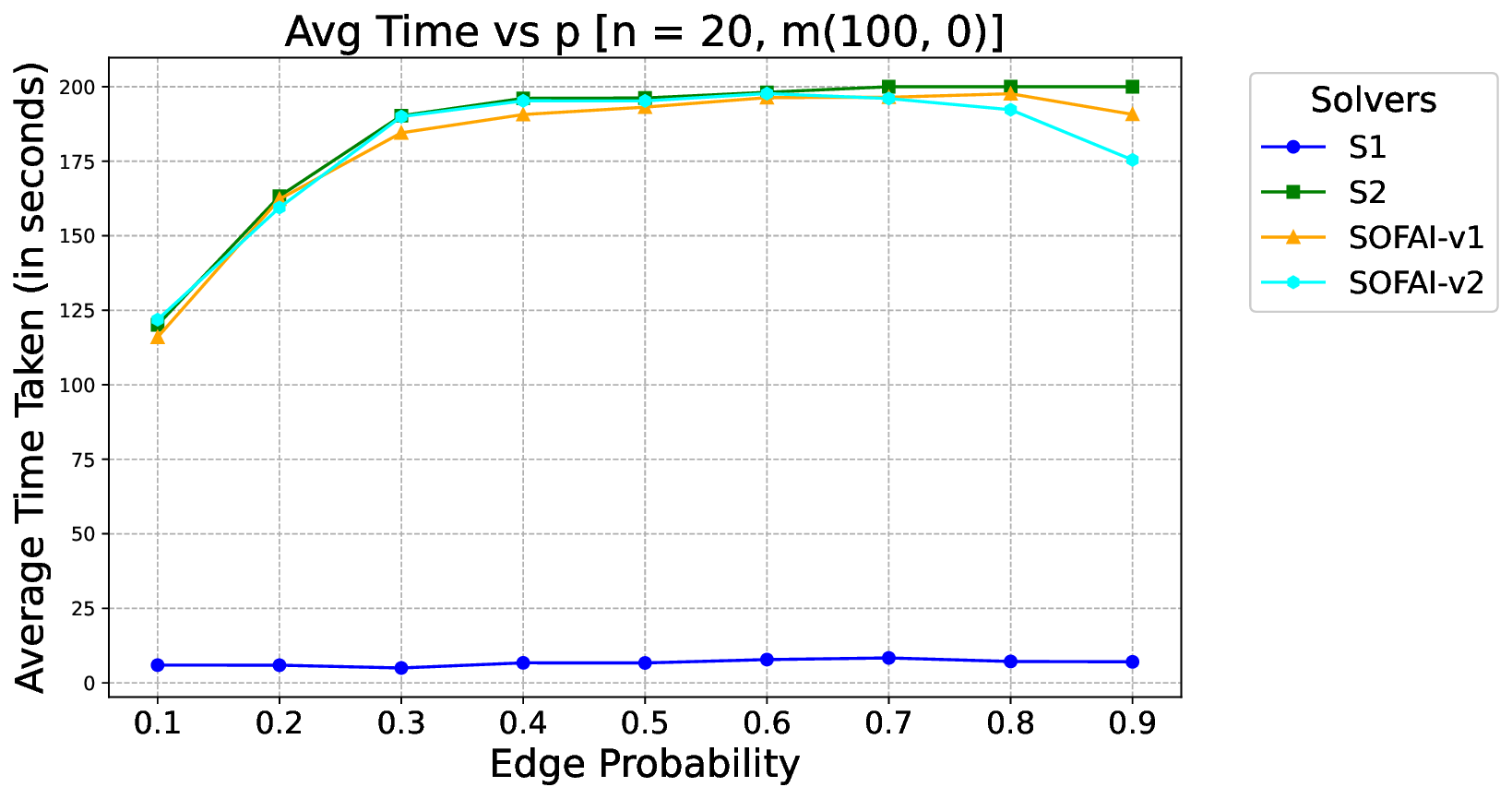}
\caption{All Solvable ($m = (100, 0)$)}
\label{fig:alls_edge_t}
\end{subfigure}
\hfill
\begin{subfigure}[b]{0.32\textwidth}
\includegraphics[width=\textwidth]{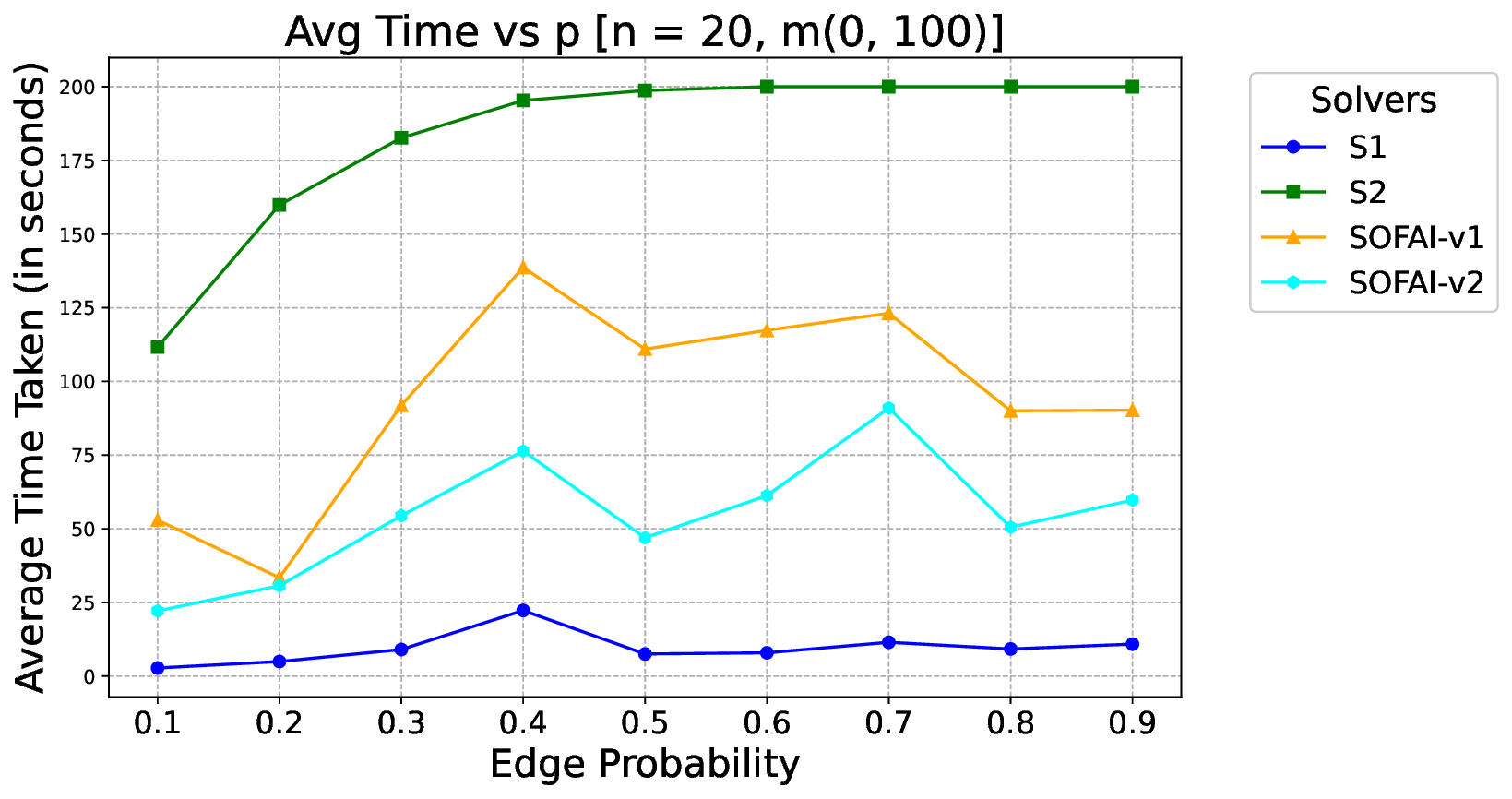}
\caption{All Unsolvable ($m = (0, 100)$)}
\label{fig:allus_edge_t}
\end{subfigure}
\hfill
\begin{subfigure}[b]{0.32\textwidth}
\includegraphics[width=\textwidth]{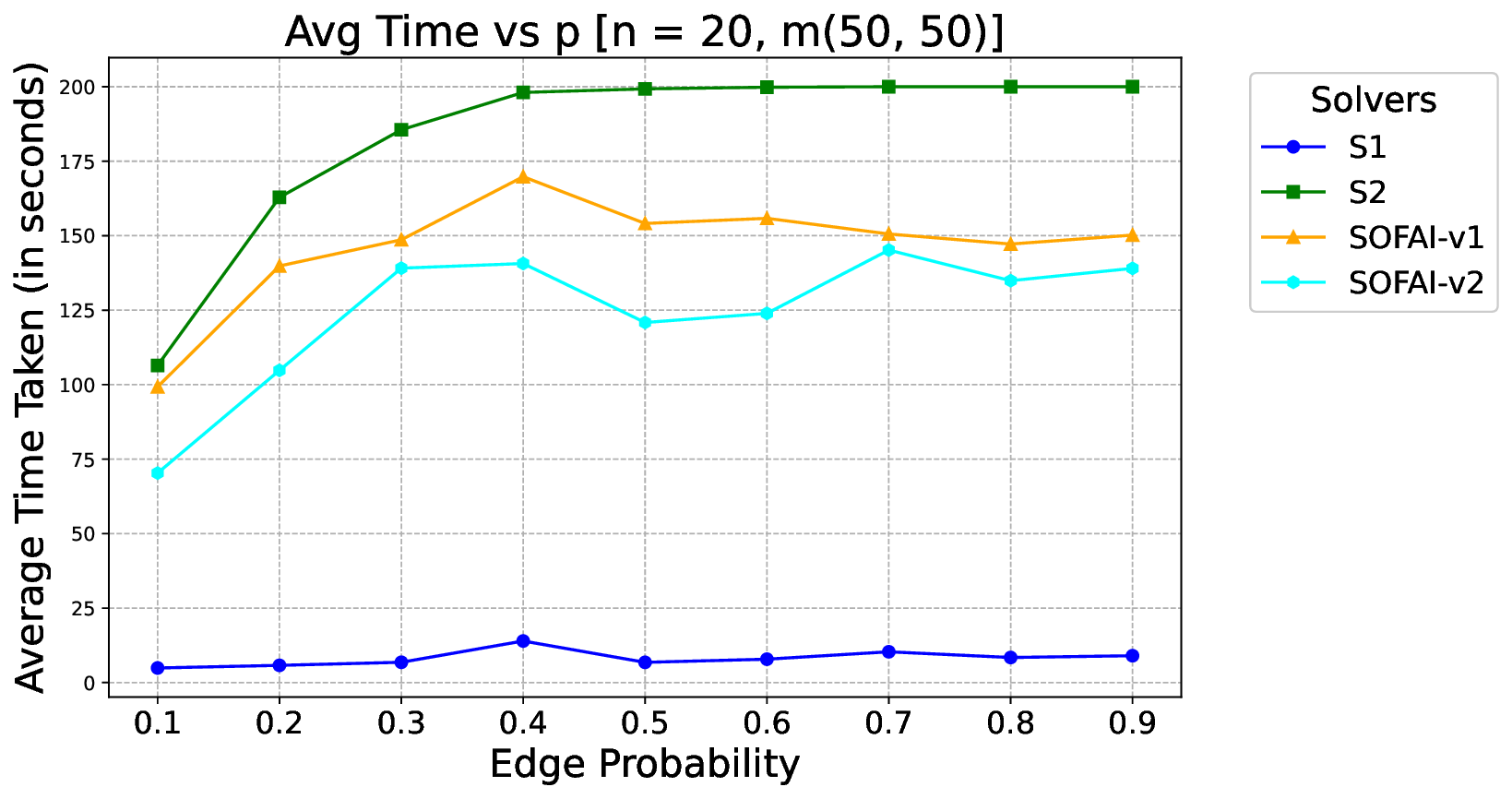}
\caption{Balanced Mix ($m = (50, 50)$)}
\label{fig:allhalf_edge_t}
\end{subfigure}
\vspace{+1em}
\caption{Time efficiency across solvers as a function of edge probability in different problem configurations, showcasing SOFAI-v2’s performance in reducing average time taken.}
\label{fig:rq4_2}
\end{figure*}

\noindent\textbf{Answer to RQ 3:}  
Yes, the iterative calling mechanism substantially enhances the performance of SOFAI-v2's S1 solver, driven by effective metacognitive governance. As illustrated in Figure~\ref{fig:rq3}, repeated solver invocations with structured feedback consistently improve success rates across all problem configurations:

\begin{itemize}
    \item In fully solvable problems (\(m=(100,0)\)), the initial success rate at iteration MC-S1-I1 (around 20\% for graph size \(n=5\)) improves substantially to nearly 80\% by iteration MC-S1-I5, a notable increase of 300\%.

    \item For completely unsolvable instances (\(m=(0,100)\)) at \(n=10\), success rates begin at approximately 60\% (MC-S1-I1) and increase to 85\% by MC-S1-I5, an improvement of 40\%.

    \item In balanced solvability scenarios (\(m=(50,50)\)) at graph size \(n=10\), success rates also see meaningful improvements, increasing from approximately 40\% (MC-S1-I1) to 60\% (MC-S1-I5), representing a 50\% improvement.
\end{itemize}

These consistent improvements underscore how SOFAI-v2’s metacognitive governance improves solver responses through iterative feedback, enhancing overall adaptability and efficiency across diverse problem settings. For the above results, we set a fixed time limit of 200 seconds per problem. If a solver does not finish within this time, it is counted as a failure.

\begin{tcolorbox}[
colframe=black!0, colback=gray!5, coltitle=black,
boxrule=0.5pt, title=RQ4, fonttitle=\bfseries,
width=\columnwidth, rounded corners,
before skip=10pt, after skip=10pt,
colbacktitle=gray!20
]
\textit{How does the density of graph influence success rate and time across the considered solvers?}
\end{tcolorbox}

\noindent\textbf{Answer to RQ 4:}  
Edge probability ($p$) substantially affects both solvability and solving time across graph coloring solvers, with SOFAI-v2 demonstrating superior adaptability and efficiency. Figure~\ref{fig:rq4_1} shows that increasing \(p\) generally reduces success rates due to higher complexity from additional edges. However, SOFAI-v2 consistently maintains higher success rates compared to other solvers:

\begin{itemize}
    \item In fully solvable problems (\(m=(100,0)\)), all solvers maintain high success rates as \(p\) increases, with SOFAI-v2 performing comparably to S2 and SOFAI-v1.

    \item In completely unsolvable scenarios (\(m=(0,100)\)), SOFAI-v2 notably outperforms others, particularly at lower edge probabilities. For instance, at \(p=0.1\), SOFAI-v2 achieves a 94\% success rate, considerably surpassing SOFAI-v1 (74\%), S1 (68\%), and S2 (51\%). This advantage remains pronounced (10-20\%) even as \(p\) increases.

    \item In mixed scenarios (\(m=(50,50)\)), SOFAI-v2 again excels, achieving a peak success rate of about 72\% at \(p=0.1\), approximately 15-25\% higher than other solvers.
\end{itemize}

Regarding time efficiency (Figure~\ref{fig:rq4_2}), as \(p\) increases, solving times rise initially and then stabilize. However, SOFAI-v2 consistently requires less time across all problem configurations:

\begin{itemize}
    \item In the unsolvable scenario (\(m=(0,100)\)), SOFAI-v2 solves problems substantially faster at lower and mid-range probabilities. For instance, at \(p=0.1\), SOFAI-v2 averages 20 seconds per problem, far less than SOFAI-v1 (55 seconds) and S2 (112 seconds).

    \item For mixed solvability scenarios (\(m=(50,50)\)), SOFAI-v2 again demonstrates superior efficiency, 40-50\% faster. Specifically, at \(p=0.1\), SOFAI-v2 solves problems in about 70 seconds, compared to SOFAI-v1’s 100 seconds and S2’s 108 seconds.
\end{itemize}

Overall, SOFAI-v2’s robust metacognitive governance enables it to adapt efficiently and effectively to varying complexities induced by changing edge probabilities, achieving higher success rates and superior time efficiency. For the above results we set a fixed time limit of 200 seconds per problem. If a solver does not finish within this time, it is counted as a failure.

%% file: content_files/conclusion.tex
\subsection{Other Considerations}
We also explicitly address concerns regarding context window limitations inherent in modern LLMs. Although iterative prompting in our method increases token consumption, our analyses confirm that token usage remains comfortably within the Mistral-7B model’s 32k-token context limit, even under extreme conditions (graph size 100, edge probability 0.9, and five iterations):

\begin{itemize}
    \item Graph Representation (DIMACS format): approximately 1500 tokens
    \item Initial Prompt \& Instructions: approximately 500 tokens
    \item Metacognitive Feedback per iteration (×5): approximately 2500 tokens (500 tokens each)
    \item Episodic Memory Retrieval per iteration (×5): approximately 5000 tokens (1000 tokens each)
    \item LLM Response per iteration (×5): approximately 5000 tokens (1000 tokens each)
    \item Accumulated Prompt per iteration (including previous responses, ×5): approximately 15{,}000 tokens (3000 tokens each iteration)
\end{itemize}

In total, even the worst-case scenario results in approximately 24{,}500 tokens, well within Mistral-7B’s 32k-token context window. Consequently, context-window overflow is not an issue in our experiments. \textit{Operationally, the metacognitive governance budgets information (problem, feedback, episodic examples) so that fast/automatic proposals remain verifiable within resource bounds, aligning with supervised control over routine processing \citep{norman1986attention, cooper2000contention}.}

Additionally, to account for the inherent variability in LLM output across runs, we repeat all experiments involving SOFAI-v1, SOFAI-v2, and S1 three times using the same set of problem instances. For success rate evaluation, a problem is considered a success if the solver succeeds in \textit{any} of the three runs. For average time taken, we report the \textit{minimum} time observed across the three trials. This choice reflects a conservative estimation of each system's upper potential in ideal conditions and avoids penalizing performance due to stochastic failures. Such a best-of-$n$ (here $n{=}3$) strategy is commonly used in LLM evaluations to mitigate randomness and highlight consistent solver capabilities \citep{kang2025scalable, chow2024inference}. \textit{Cognitively, this also controls for noise in routine responding while assessing whether supervisory interventions reliably reduce errors.}

We also note that SOFAI-based solvers (and S1) can declare \textsc{UNSAT} without a constructive certificate, whereas S2’s \textsc{UNSAT} arises from explicit search. This asymmetry can bias results on mixes containing unsatisfiable instances. However, the systematic variation in success rates across graph sizes and densities indicates the behavior is not arbitrary and reflects nontrivial problem sensitivity. A full proof-producing pipeline for \textsc{UNSAT} is orthogonal to our current scope; for automatic and quick evaluation, we accept such declarations, and we flag this as a limitation to be addressed in future work

\subsection{Implications and Comparisons with Prior Work}

This study broadens the scope of neurosymbolic systems by applying them to graph coloring problems. It demonstrates that integrating LLMs with traditional symbolic solvers, guided by metacognitive governance, enhances accuracy and efficiency. Unlike previous applications that primarily leveraged LLMs for sequential decision-making tasks \citep{valmeekam2022large, stechly2024self}, SOFAI-v2 directly addresses constraint-adherence limitations through iterative feedback and strategic use of symbolic reasoning. \textit{Viewed through cognitive control, S1 provides routine proposals, while MC and S2 supply supervisory/controlled processing when conflicts persist, consistent with contention scheduling and supervisory attention \citep{norman1986attention, cooper2000contention}.} This neurosymbolic framing clarifies why SOFAI-v2 improves success rate and time efficiency on constraint satisfaction problems.

\section{Conclusion and Future Work}

We introduced SOFAI-v2, a neurosymbolic architecture that integrates a fast, episodic memory-based LLM (S1) with a slow, deliberative symbolic solver (S2), enhanced by real-time metacognitive governance. Our empirical evaluations across various graph coloring problems confirm that SOFAI-v2 outperforms traditional symbolic solvers and its predecessor, SOFAI-v1, in both success rates and time efficiency, especially within complex problem configurations. SOFAI-v2 employs iterative metacognitive feedback and episodic memory-based iterations to enable S1 to refine its response. This effectively overcomes the limitations of LLMs in constraint \textit{adherence} for CSPs. By leveraging episodic memory and adaptive feedback, SOFAI-v2 improves the accuracy and efficiency of solving CSPs, indicating the value of cognitively informed, neurosymbolic architectures.

The observed improvements are consistent with dual-process theories: S1 generates a \emph{fast, potential solution}; the metacognitive controller checks it for constraint adherence and uncertainty, requests targeted revisions when needed, and escalates to S2 only when necessary; S2 then performs \emph{slow}, symbolic search as invoked. Episodic retrieval provides additional context from prior instances to support S1’s proposal and the controller’s checks. In this arrangement, SOFAI-v2 realizes a control loop that finds a trade-off between speed and reliability in a resource-constrained setting.

Future work will scale SOFAI-v2 to larger and more varied problem sets, explore different graph models, and apply the architecture to other CSPs and sequential decision-making tasks. We will also investigate learning-based MC policies and richer memory retrieval, and examine correspondences to cognitive-control signals (e.g., conflict costs) that could enable tighter links to computational cognitive neuroscience. \emph{Concretely, we will design human-aligned probes (e.g., speed–accuracy manipulations, conflict-cost perturbations, and memory cueing/interference) to test whether SOFAI-v2’s control dynamics mirror classic findings on supervisory attention and conflict-driven effort allocation.}

%% file: content_files/appendix.tex
\newpage
\clearpage
\section*{Appendix}
\section{Modified DSATUR Algorithm with Backtracking}
\label{appendix:dsatur}

Given the graph coloring problem $(G, C, k)$, with an undirected graph $G = (V, E)$, a set of available colors $C = {c_1, c_2, \dots, c_k}$, and the maximum allowed number of colors $k$, the modified DSATUR algorithm systematically explores assignments using backtracking. Below is a detailed, mathematically rigorous description:

\textbf{Definitions}:
\begin{itemize}
\item \textbf{Saturation degree} $d_{sat}(v)$ for a vertex $v \in V$ is the cardinality of the set of distinct colors assigned to its adjacent vertices:
\item \textbf{Vertex degree} $d(v)$ is the number of edges incident to $v$.
\end{itemize}

\textbf{Algorithm Steps}:
\begin{enumerate}
\item \textbf{Initialization}: Assign an initial coloring $f(v) = \emptyset$ for all $v \in V$ (all vertices uncolored).

\item \textbf{Compute Saturation Degree}: For each uncolored vertex $v$, compute its saturation degree $d_{sat}(v)$.

\item \textbf{Select Vertex}: Identify vertex $v^*$ that maximizes $d_{sat}(v)$. If multiple vertices have equal highest saturation degree, select the one with the highest vertex degree $d(v)$.

\item \textbf{Color Assignment}: Assign to $v^*$ the smallest color $c \in C$ that satisfies the condition:
\[ f(v^*) \neq f(u), \quad \forall u \text{ adjacent to } v^* \]

\item \textbf{Backtracking}: If no valid color can be assigned, recursively backtrack:
\begin{itemize}
    \item Undo the previous assignment and select the next smallest available color.
    \item Continue this recursive search until a valid color assignment is found or all possibilities are exhausted.
\end{itemize}

\item \textbf{Termination}: Repeat steps 2-5 until all vertices are successfully colored, or conclusively determine no valid coloring exists under the provided constraint of $k$ colors.

\end{enumerate}

This rigorous exploration via backtracking guarantees the determination of a feasible coloring if one exists within the constraint of using no more than $k$ colors.

\section{Algorithm SOFAI-v2}\label{appendix:algosfv2}
\noindent\textbf{Algorithm Description.}\\
Given a graph $G=(V,E)$ and a color bound $k$, SOFAI-v2 runs an S1--MC loop for at most $T$ iterations. 
In each iteration, S1 is prompted with the graph, $k$, and the accumulated history. 
S1 proposes a coloring assignment, which is checked for feasibility. 
If valid, the algorithm returns the assignment. 
If invalid, MC extracts the conflicts, generates minimal checkable feedback, and appends this to the history and trend signals. 
If repeated patterns or the iteration limit is reached, MC escalates to S2. 
S2 runs a DSATUR backtracking solver for the fixed-$k$ decision problem. 
If a valid coloring is found, it is returned and stored in memory; otherwise, the algorithm reports ``NOT SOLVABLE''.

\begin{algorithm}[H]
\caption{SOFAI-v2 for Graph Coloring (Decision, fixed $k$)}
\label{alg:sofaiv2-gc}
\begin{algorithmic}[1]
\Require Undirected graph $G=(V,E)$, color limit $k$, max S1 iterations $T$, memory $\mathcal{M}$
\Ensure Coloring assignment or ``NOT SOLVABLE''
\State $\textit{history} \gets [\,]$ \Comment{list of $(assign, feedback)$}
\State $\textit{trends} \gets [\,]$ \Comment{patterns for MC}
\State $\textit{seed} \gets \text{RetrieveMemory}(\mathcal{M}, G, k)$
\For{$t=1$ \textbf{to} $T$}
    \State $prompt \gets \text{GenerateGCPrompt}(G,k,seed,history)$
    \State $resp \gets \text{CallS1}(prompt)$
    \State $assign \gets \text{ParseColoring}(resp)$
    \State $(ok,viol) \gets \text{CheckFeasibility}(G,k,assign)$
    \If{$ok=\text{true}$}
        \State \Return $assign$
    \EndIf
    \State $C \gets \text{ExtractConflicts}(viol)$
    \State $H \gets \text{InducedSubgraph}(G,C)$
    \State $f \gets \text{FormatFeedback}(H,C,assign)$
    \State $history \gets history \cup \{(assign,f)\}$
    \State $trends \gets trends \cup \{\text{FeedbackSignature}(f)\}$
    \If{$\text{EscalationRule}(trends,t)$}
        \State \textbf{break} \Comment{escalate to S2}
    \EndIf
    \State $seed \gets \text{UpdateSeed}(assign,f)$
\EndFor
\Statex
\Statex \textbf{// S2 fallback (decision, fixed $k$)}
\State $(status, assign) \gets \text{DSATUR\_BT}(G,k)$
\If{$status=\text{SAT}$}
    \State $\text{UpdateMemory}(\mathcal{M},G,k,assign)$
    \State \Return $assign$
\Else
    \State \Return ``NOT SOLVABLE''
\EndIf
\end{algorithmic}
\end{algorithm}

\section{Prompt, Example, and Feedback Examples}
\label{appendix:inout}
This section discusses prompt type, feedback types, and generated examples. Feedback and generated example types are examples of the implementation of SOFAI-v2 iterative feedback mechanism with example generation. 

\begin{figure}[H]
\centering
\begin{tcolorbox}[
    colframe=black!0, colback=gray!5, coltitle=black,
    boxrule=0.5pt, title=Graph Coloring Problem Prompt, fonttitle=\bfseries,
    width=\columnwidth, rounded corners, colbacktitle=gray!20
]
\textbf{New Problem to Solve:}\\
You are given an undirected graph with 2 colors available. Your task is to assign a color to each vertex such that no two adjacent vertices share the same color.\\

\vspace{0.2cm}
\textbf{Graph Representation:}\\
- Number of vertices and edges: \texttt{p edge 5 5}.\\
- Edges between vertices are listed as follows:\\
\texttt{e A B}\\
\texttt{e A C}\\
\texttt{e B C}\\
\texttt{e C D}\\
\texttt{e D E}

\vspace{0.2cm}
\textbf{Objective:}\\
Assign a unique color to each vertex, ensuring that no two vertices connected by an edge have the same color. Use no more than 2 distinct colors. Provide the color assignments for each vertex in the format:\\
(Vertex Color)

\vspace{0.2cm}
\textbf{Example Format:}\\
\texttt{(A 1)}\\
\texttt{(B 2)}\\
\texttt{(C 1)}

\vspace{0.2cm}
Please provide the color assignment for the new problem to solve, or respond with "NOT SOLVABLE" if it cannot be solved.
\end{tcolorbox}
\caption{LLM prompt template for the graph coloring decision problem without using episodic memory.}
\label{fig:graph_coloring_prompt}
\end{figure}

\begin{figure}[H]
\centering
\begin{tcolorbox}[
    colframe=black!0, 
    colback=gray!5, 
    coltitle=black,
    boxrule=0.5pt, 
    title=Graph Coloring Problem Prompt with Episodic Memory Example, 
    fonttitle=\bfseries,
    width=\columnwidth, 
    rounded corners, 
    colbacktitle=gray!20
]
\textbf{New Problem to Solve:}\\
You are given an undirected graph with 2 colors available. Your task is to assign a color to each vertex such that no two adjacent vertices share the same color.

\vspace{0.2cm}
\textbf{Graph Representation:}\\
- Number of vertices and edges: \texttt{p edge 5 5}.\\
- Edges between vertices are listed as follows:\\
\texttt{e A B}\\
\texttt{e A C}\\
\texttt{e B C}\\
\texttt{e C D}\\
\texttt{e D E}

\vspace{0.2cm}
\textbf{Objective:}\\
Assign a unique color to each vertex, ensuring that no two vertices connected by an edge have the same color. Use no more than 2 distinct colors. Provide the color assignments for each vertex in the format:\\
\texttt{(Vertex Color)}

\vspace{0.2cm}
\textbf{Episodic Memory Example:}\\
\textbf{Graph:} \texttt{p edge 3 3}\\
\textbf{Edges:}\\
\texttt{e A B}\\
\texttt{e B C}\\
\texttt{e C A}
\textbf{Solution:}\\
\texttt{(A 1)}\\
\texttt{(B 2)}\\
\texttt{(C 3)}

\vspace{0.2cm}
This example is provided to demonstrate a previously successful coloring strategy for a similar subgraph, which may be helpful in solving the current problem.

\vspace{0.2cm}
Please provide the color assignment for the new problem to solve directly below, or respond with "NOT SOLVABLE" if it cannot be solved.
\end{tcolorbox}
\caption{LLM prompt template for $m(0,100)$ and $m(50,50)$ for the graph coloring problem, incorporating an episodic memory example to aid problem-solving.}
\label{fig:graph_coloring_prompt_with_memory}
\end{figure}

\begin{figure}[H]
\centering
\begin{tcolorbox}[
    colframe=black!0, colback=gray!5, coltitle=black,
    boxrule=0.5pt, title=Graph Coloring Feedback - Exceeding Limit, fonttitle=\bfseries,
    width=\columnwidth, rounded corners, colbacktitle=gray!20
]
An attempt was made to solve the graph coloring using three colors, exceeding the maximum allowed.

\vspace{0.2cm}
\textbf{Over-Colored Submission:}\\
\texttt{(A 1)}\\
\texttt{(B 2)}\\
\texttt{(C 3)}\\
\texttt{(D 1)}\\
\texttt{(E 2)}

\vspace{0.2cm}
\textbf{Feedback Provided:}\\
\texttt{Error: Only 2 colors are allowed. 3 colors were used.}
\end{tcolorbox}
\caption{Examples of feedback provided to coloring assignment, exceeding the number of colors limit.}
\label{fig:incorrect_coloring_feedback}
\end{figure}

\begin{figure}[H]
\centering
\begin{tcolorbox}[
    colframe=black!0, colback=gray!5, coltitle=black,
    boxrule=0.5pt, title=Graph Coloring Feedback -- Incorrect Coloring, fonttitle=\bfseries,
    width=\columnwidth, rounded corners, colbacktitle=gray!20
]
Given an undirected graph with 2 colors available, the vertices were incorrectly colored as follows:

\vspace{0.2cm}
\textbf{Graph Representation:}\\
- Number of vertices and edges: \texttt{p edge 5 5}.\\
- Edges between vertices are listed as follows:\\
\texttt{e A B}\\
\texttt{e A C}\\
\texttt{e B C}\\
\texttt{e C D}\\
\texttt{e D E}

\vspace{0.2cm}
\textbf{Incorrect Coloring Submitted:}\\
\texttt{(A 1)}\\
\texttt{(B 1)}\\
\texttt{(C 2)}\\
\texttt{(D 2)}\\
\texttt{(E 1)}

\vspace{0.2cm}
\textbf{Feedback Provided:}\\
\texttt{Error: Vertices A and B are adjacent but have the same color.}\\
\texttt{Error: Vertices B and C are adjacent but have the same color.}
\end{tcolorbox}
\caption{Examples of feedback provided to incorrect coloring assignment.}
\label{fig:incorrect_coloring_feedback}
\end{figure}

\begin{figure}[H]
\centering
\begin{tcolorbox}[
    colframe=black!0, colback=gray!5, coltitle=black,
    boxrule=0.5pt, title=Graph Coloring Generated Example, fonttitle=\bfseries,
    width=\columnwidth, rounded corners, colbacktitle=gray!20
]
To generate an example, a subgraph is extracted from the problem graph to demonstrate a valid coloring with fewer vertices.

\vspace{0.2cm}
\textbf{Subgraph Representation:}\\
- Subgraph vertices and edges: \texttt{p edge 3 3}.\\
- Edges between subgraph vertices are listed as follows:\\
\texttt{e A B}\\
\texttt{e A C}\\
\texttt{e B C}

\vspace{0.2cm}
\textbf{Correct Coloring of Subgraph:}\\
\texttt{(A 1)}\\
\texttt{(B 2)}\\
\texttt{(C 3)}

\vspace{0.2cm}
This example shows how to correctly assign colors in a smaller scope of the main problem. Scaling this approach for larger graphs while maintaining the constraints is crucial.

\end{tcolorbox}
\caption{Example of a correctly colored subgraph is generated as part of the feedback to illustrate the correct coloring assignment for less number of nodes.}
\label{fig:subgraph_example}
\end{figure}

\section{Results - Average time Solver Comparison}
\label{appendix:avgtime}

Table~\ref{table:times} reports the average time taken by different solver configurations (S1, S2, SOFAI\_v1, and SOFAI-v2) under a fixed time limit of 200 seconds. These results are obtained from the same experiment used to evaluate success rates in Section~\ref{sec:results}. Instances not solved within the allotted 200 seconds are excluded from the average time calculation. Cells with a dash (\texttt{--}) indicate that the solver failed to solve any instance in that configuration within the time limit, resulting in no valid average time. These values help assess not only whether solvers succeed but also how quickly they do so under realistic constraints.

\begin{table}[!ht]
\centering
\scriptsize
\setlength{\tabcolsep}{3pt} 
\renewcommand{\arraystretch}{0.95} 
\caption{
Average time taken (in seconds) by solvers under a 200-second time limit. Results are reported for each graph size and solvability configuration. Missing values (dashes) indicate that no problem was solved within the time limit for that solver and configuration. These results correspond to the same experiments used in the success rate analysis in RQ1-Section~\ref{sec:results}.
}
\label{table:times}
\resizebox{\linewidth}{!}{%
\begin{tabular}{lcccc cccc cccc}
    \toprule
    \multirow{2}{*}{Graph Size} 
    & \multicolumn{4}{c}{$m(100,0)$ (s)} 
    & \multicolumn{4}{c}{$m(0,100)$ (s)} 
    & \multicolumn{4}{c}{$m(50,50)$ (s)} \\
    \cmidrule(lr){2-5} \cmidrule(lr){6-9} \cmidrule(lr){10-13}
     & S1 & S2 & SOFAI\_v1 & SOFAI-v2 
     & S1 & S2 & SOFAI\_v1 & SOFAI-v2 
     & S1 & S2 & SOFAI\_v1 & SOFAI-v2 \\ 
    \midrule
     5  & 3.23 & {0.001} & 3.23 & 3.92 & 3.13 & {0.001} & 3.13 & 3.77 & 3.25 & {0.001} & 3.23 & 3.92 \\ 
     10 & 3.60 & {0.19} & 3.75 & 3.21 & 3.80 & {0.17} & 3.90 & 3.07 & 3.54 & {0.22} & 3.85 & 3.74 \\ 
     15 & 3.80 & 54.25 & 32.90 & 31.21 & 3.85 & 60.36 & 34.95 & 28.21 & 3.21 & 56.32 & 53.92 & 33.92 \\ 
     20 & 3.85 & 83.22 & 54.95 & 48.72 & 3.90 & 99.06 & 55.28 & 42.31 & 3.87 & 108.9 & 71.98 & 55.63 \\ 
     30 & - & - & - & - & 3.92 & - & 3.94 & 19.92 & 3.92 & - & 3.97 & 24.23 \\ 
     40 & - & - & - & - & 3.95 & - & 3.96 & 22.94 & 3.95 & - & 3.99 & 23.25 \\ 
     50 & - & - & - & - & 3.96 & - & 3.97 & 21.76 & 3.97 & - & 3.97 & 24.45 \\ 
    \bottomrule
\end{tabular}%
}
\end{table}
 \clearpage

\section{Results - Success Rate}
\label{appendix:successrate}
\begin{figure*}[!htbp]
\centering
\begin{subfigure}[b]{0.32\textwidth}
\includegraphics[width=\textwidth]{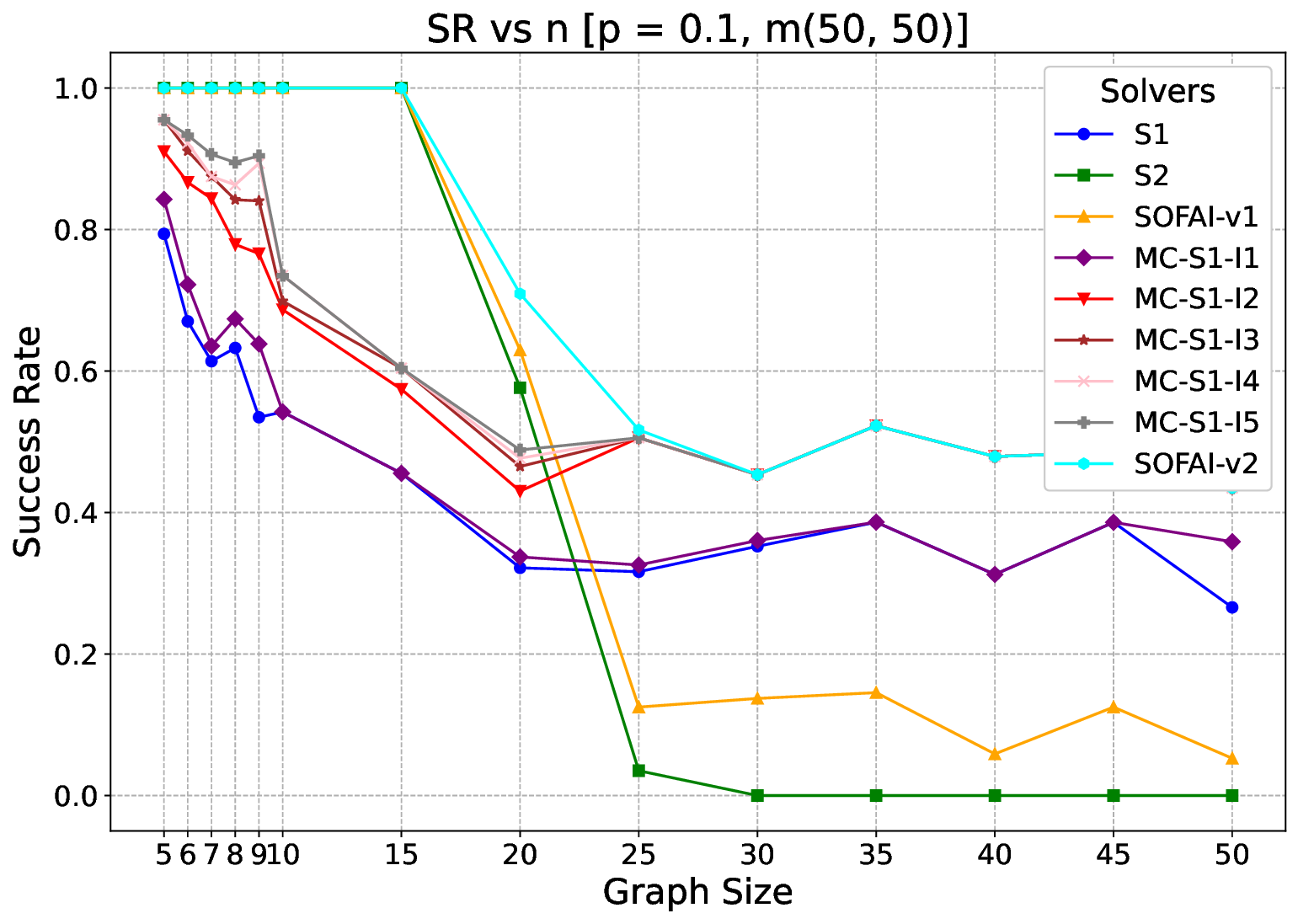}
\label{fig:alls}
\end{subfigure}
\hfill
\begin{subfigure}[b]{0.32\textwidth}
\includegraphics[width=\textwidth]{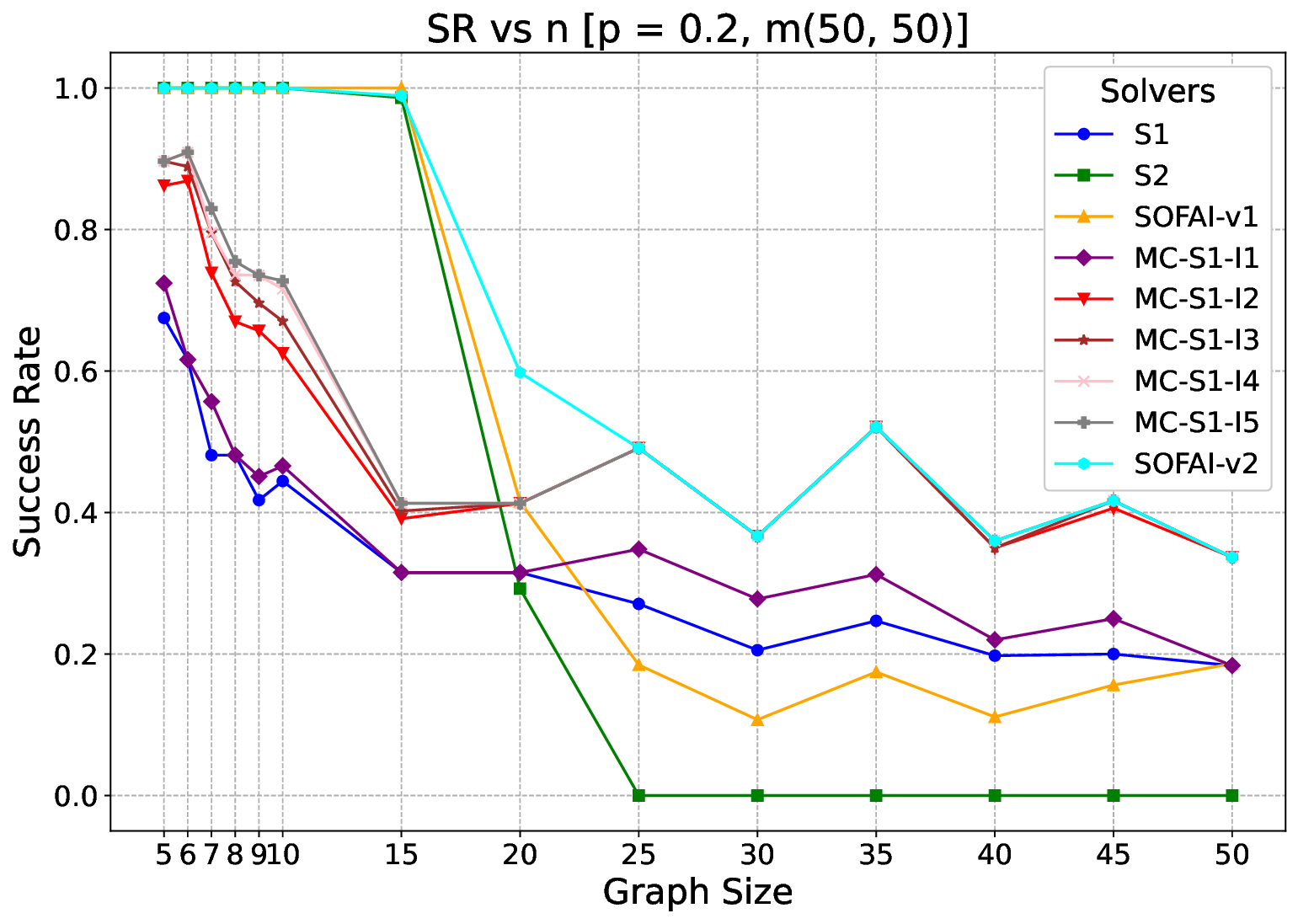}
\label{fig:allus}
\end{subfigure}
\hfill
\begin{subfigure}[b]{0.32\textwidth}
\includegraphics[width=\textwidth]{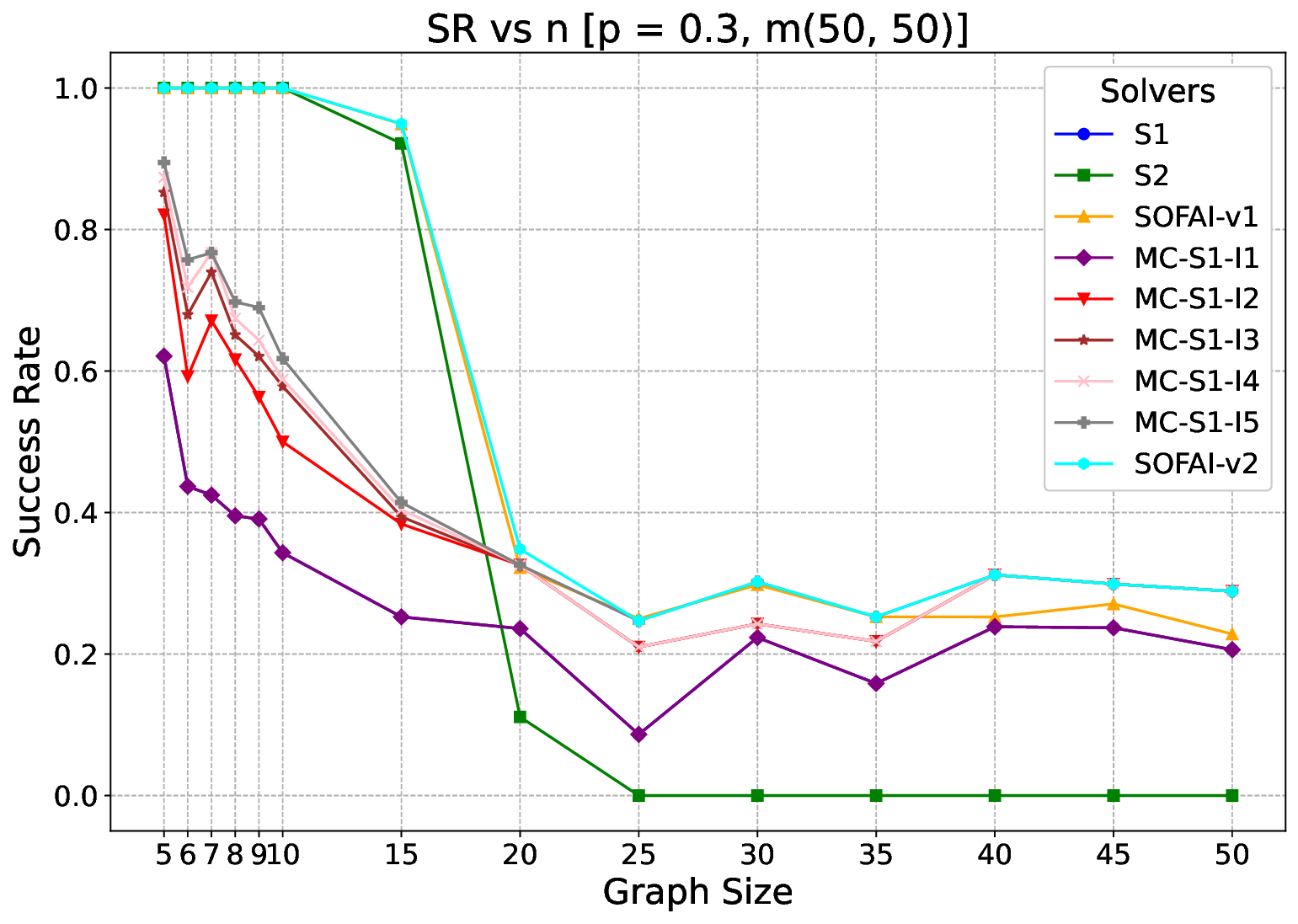}
\label{fig:allhalf}
\end{subfigure}
\begin{subfigure}[b]{0.32\textwidth}
\includegraphics[width=\textwidth]{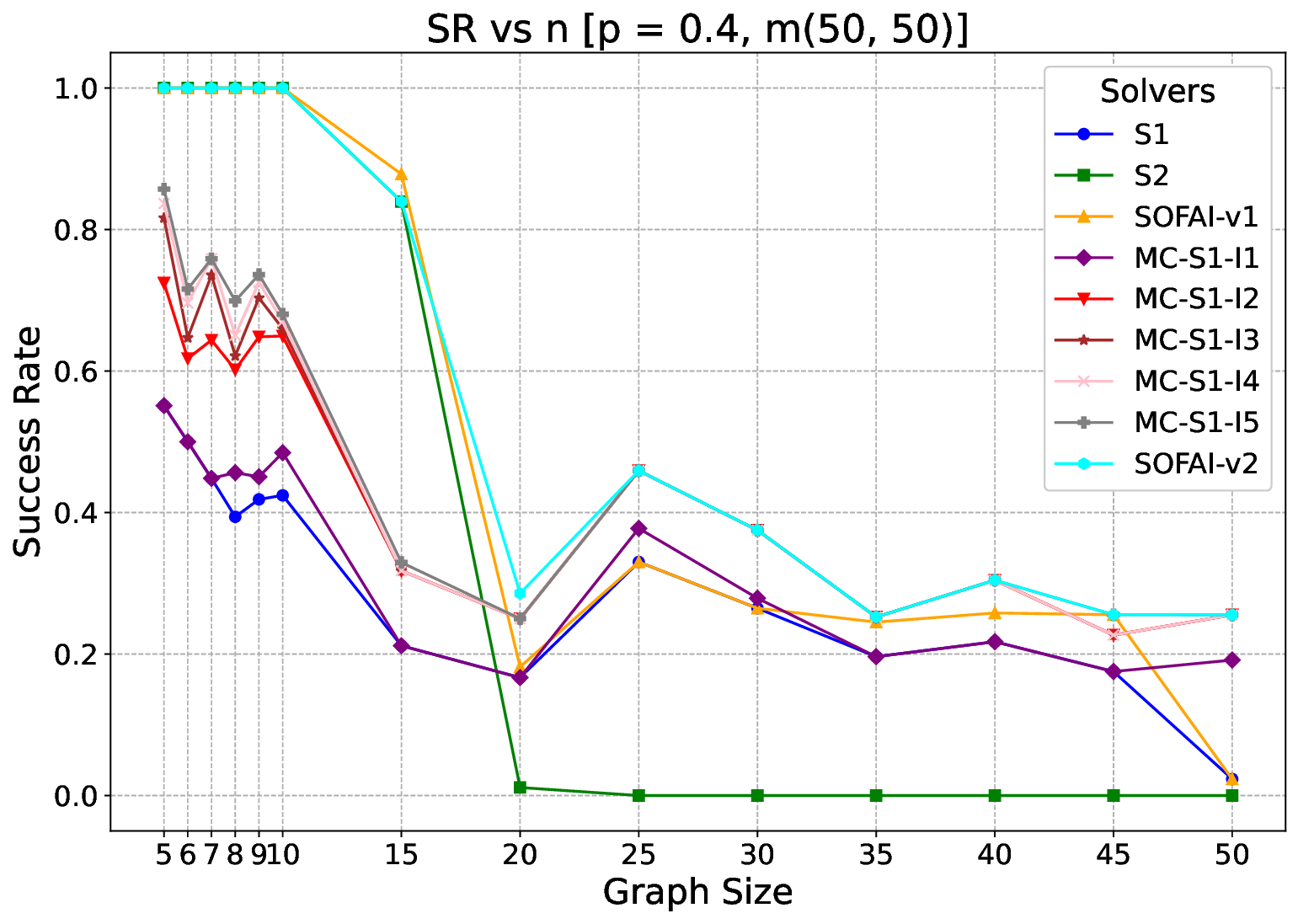}
\label{fig:alls}
\end{subfigure}
\hfill
\begin{subfigure}[b]{0.32\textwidth}
\includegraphics[width=\textwidth]{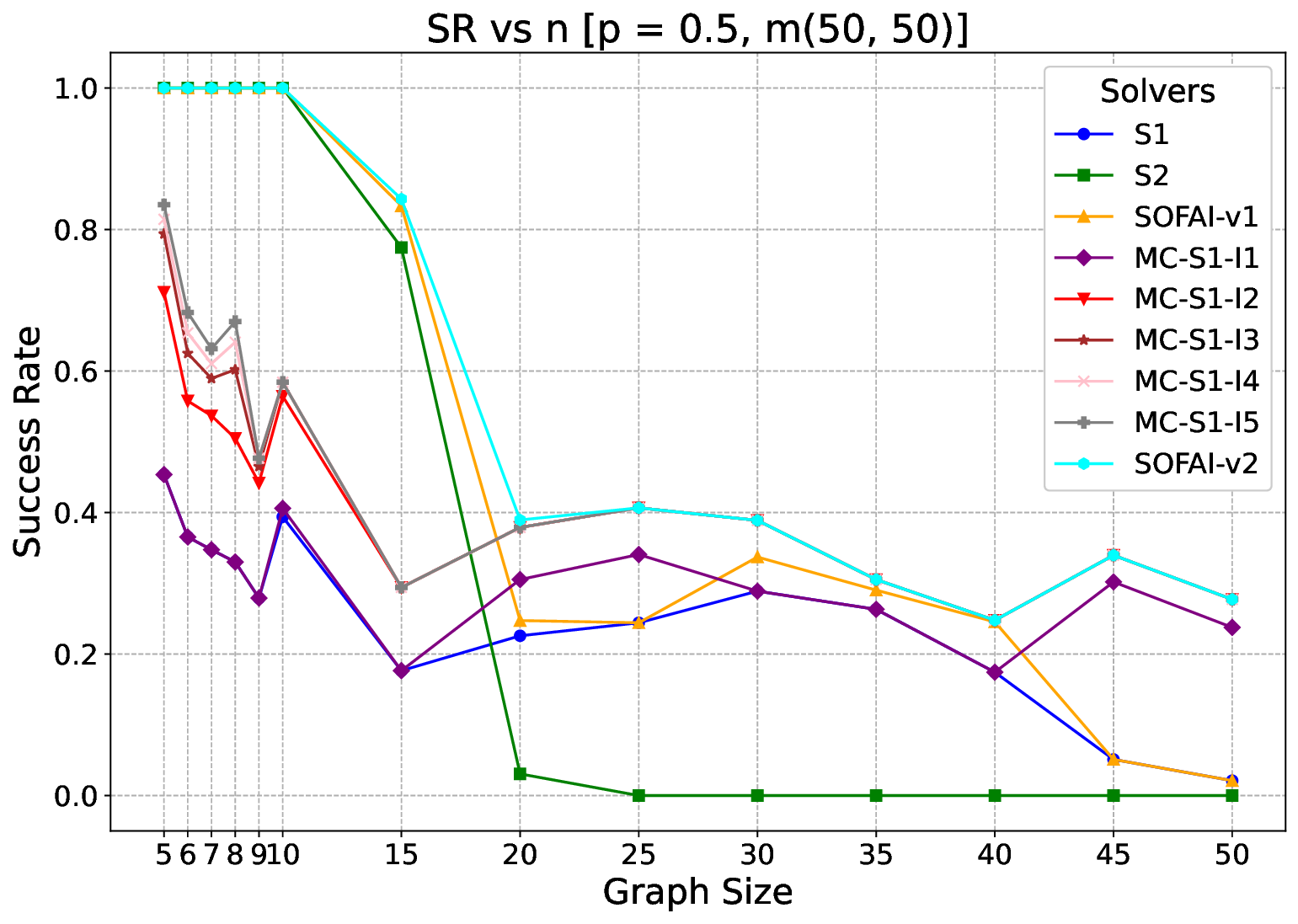}
\label{fig:allus}
\end{subfigure}
\hfill
\begin{subfigure}[b]{0.32\textwidth}
\includegraphics[width=\textwidth]{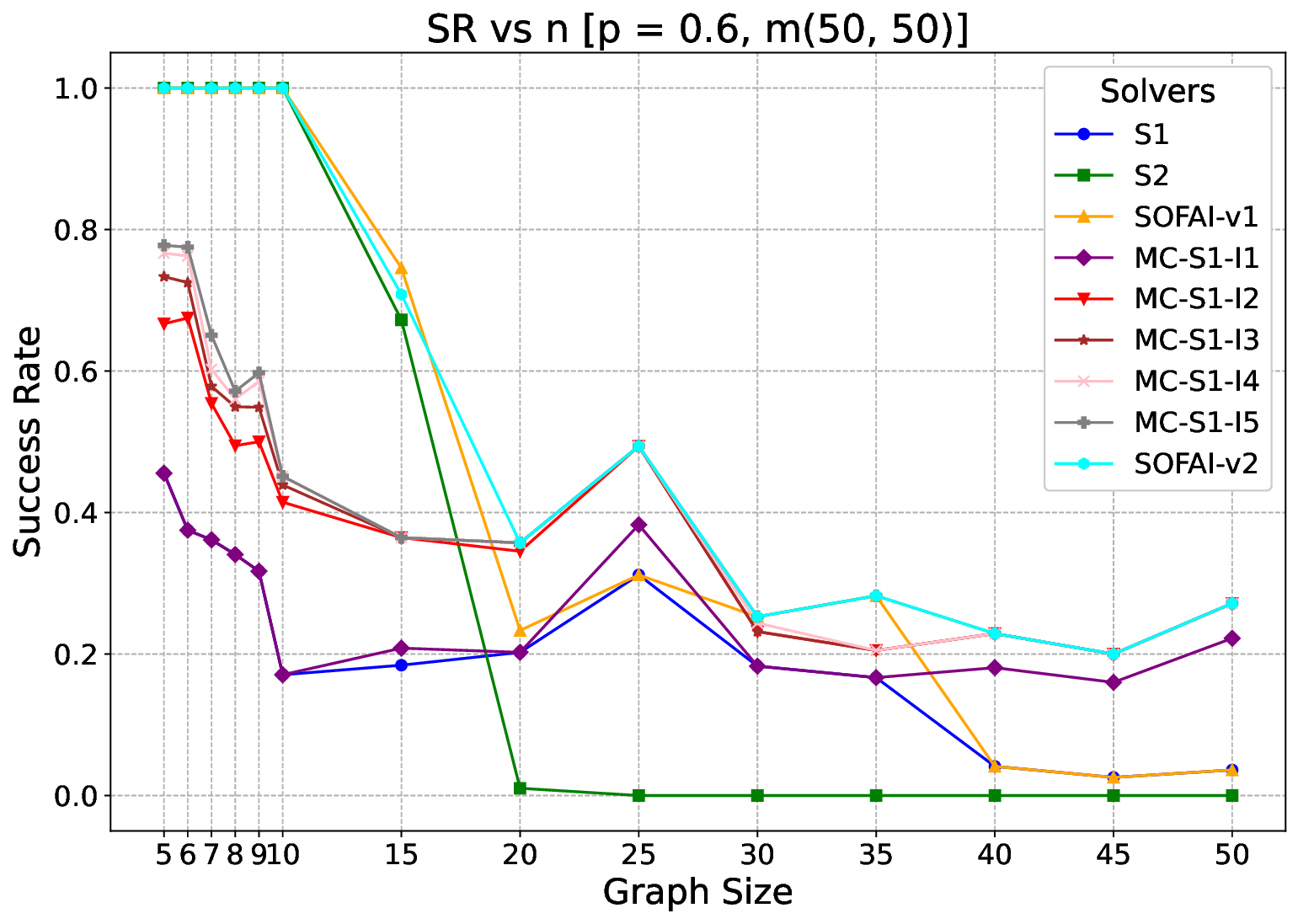}
\label{fig:allhalf}
\end{subfigure}
\begin{subfigure}[b]{0.32\textwidth}
\includegraphics[width=\textwidth]{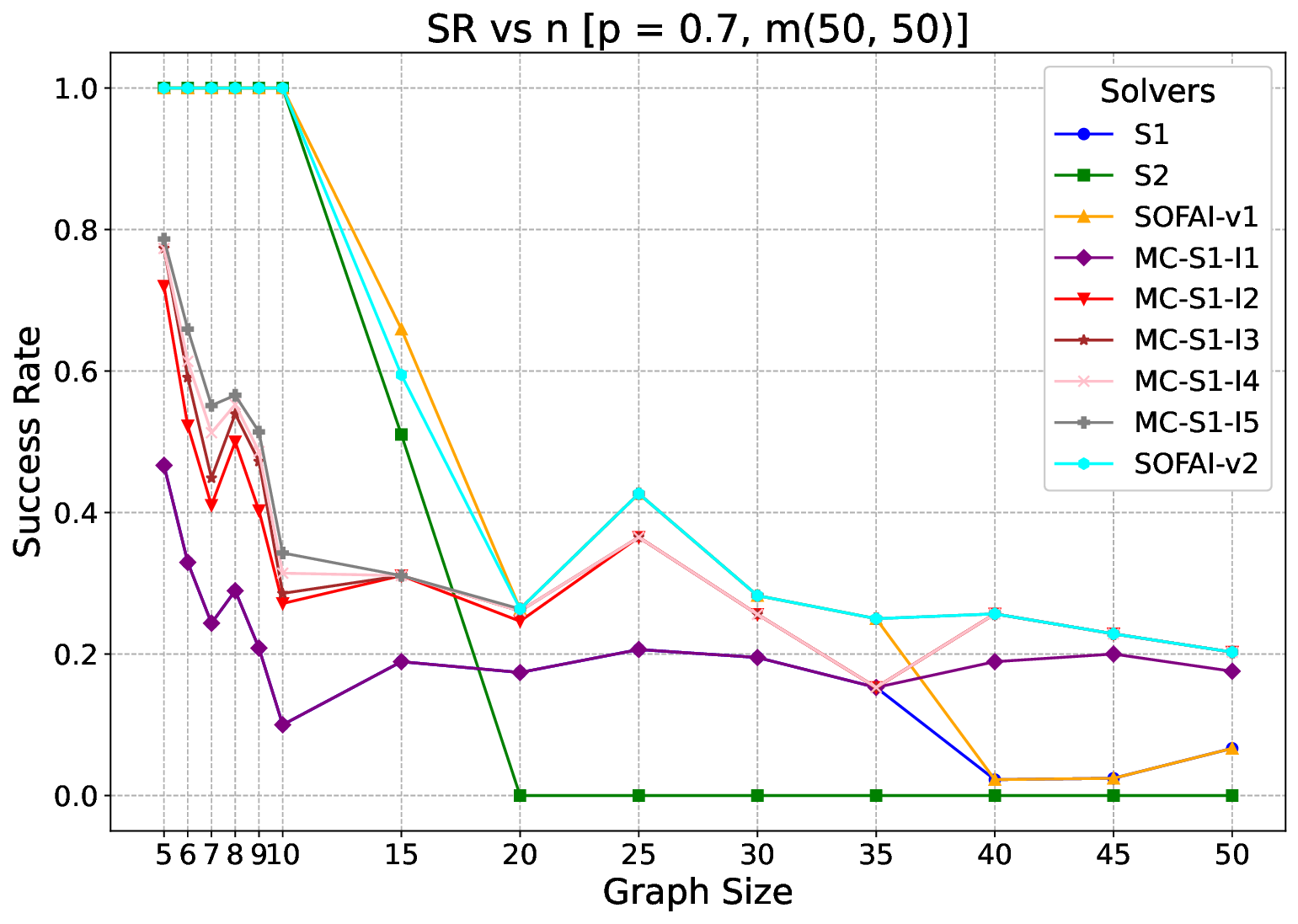}
\label{fig:alls}
\end{subfigure}
\hfill
\begin{subfigure}[b]{0.32\textwidth}
\includegraphics[width=\textwidth]{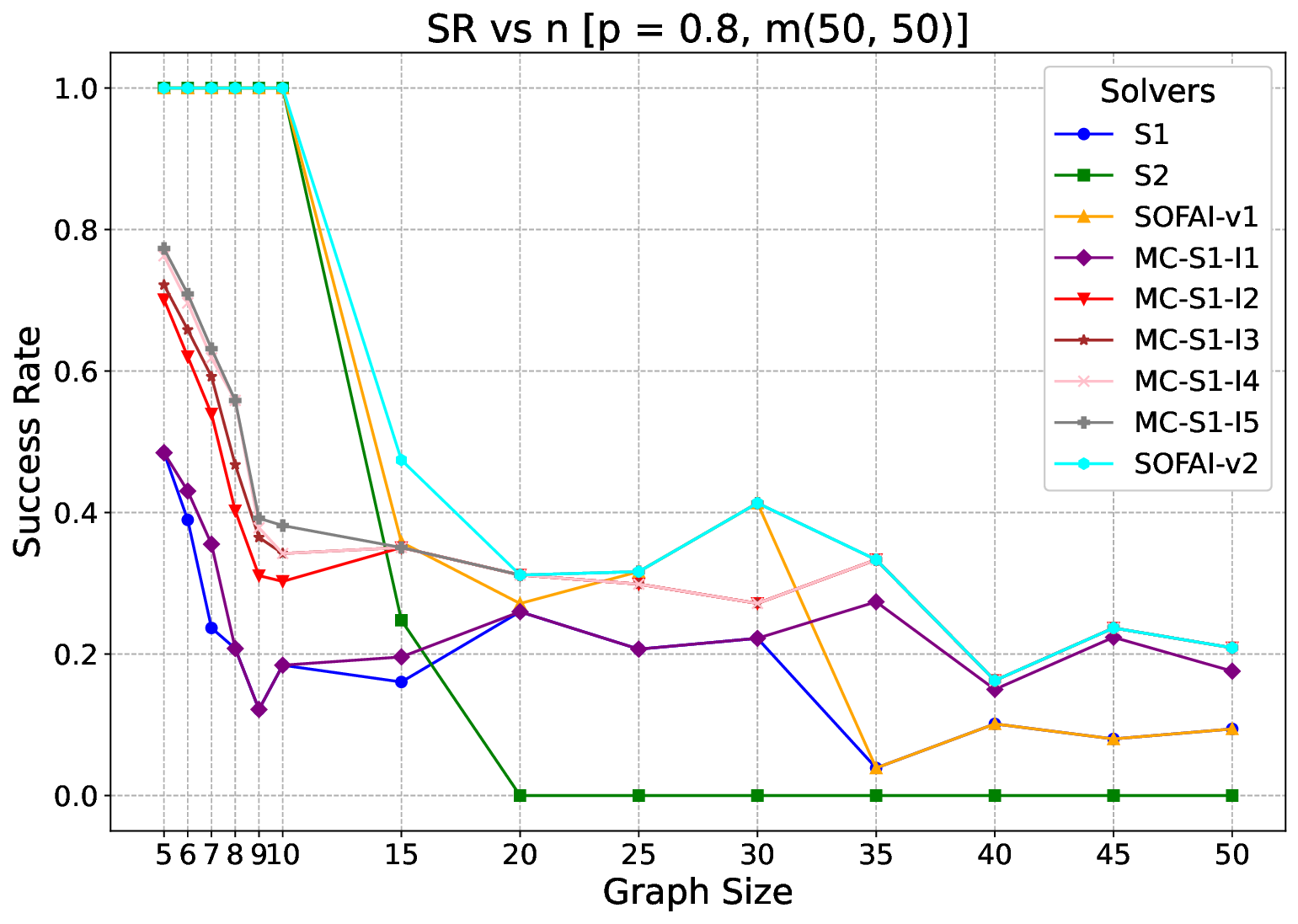}
\label{fig:allus}
\end{subfigure}
\hfill
\begin{subfigure}[b]{0.32\textwidth}
\includegraphics[width=\textwidth]{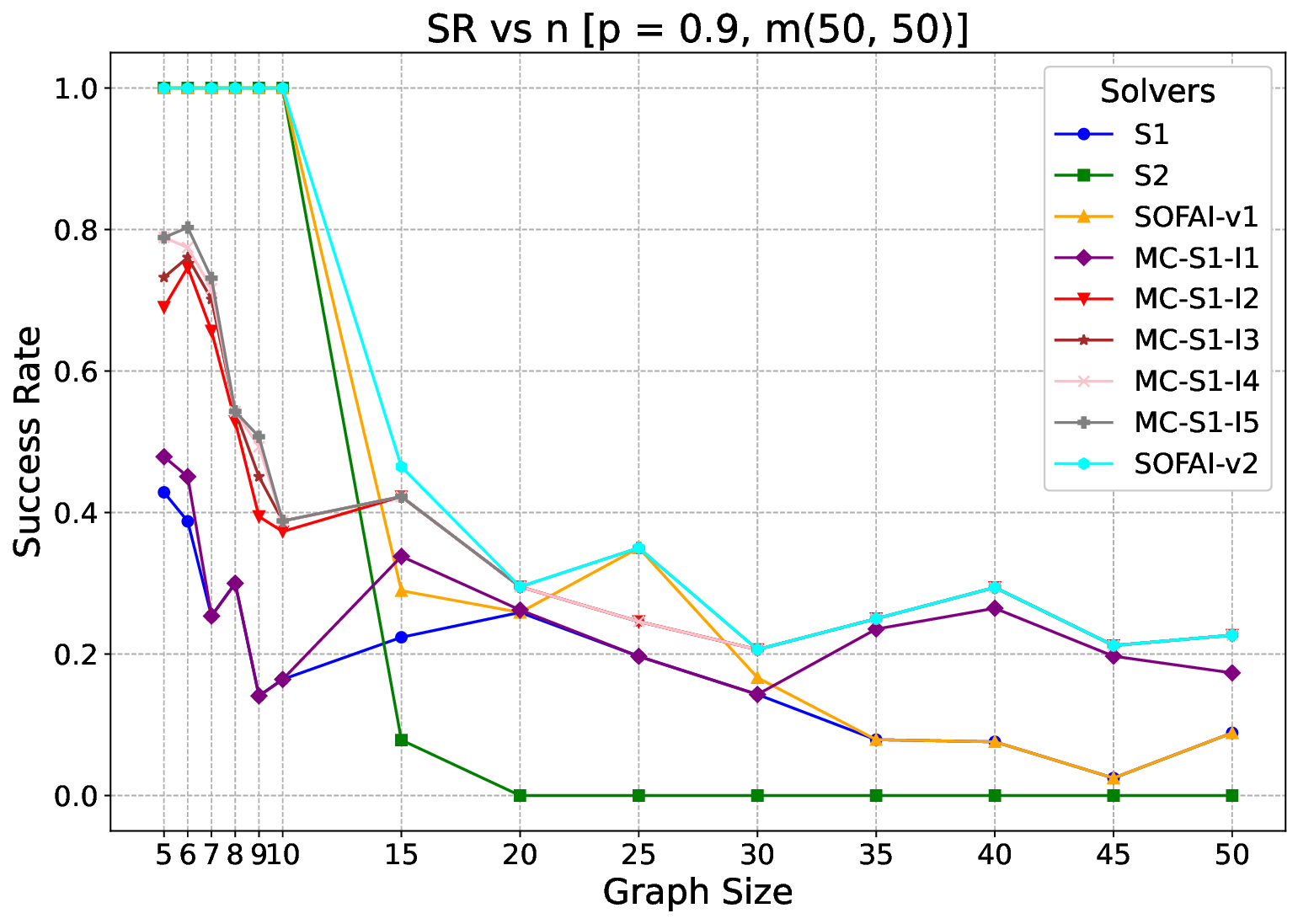}
\label{fig:allhalf}
\end{subfigure}
\caption{Success Rate (SR) of different solvers vs Graph size ($n$) across edge probabilities ($p$) for problem configuration ($m = (50, 50)$)}
\label{fig:sc_half}
\end{figure*}

\begin{figure*}[!htbp]
\centering
\begin{subfigure}[b]{0.32\textwidth}
\includegraphics[width=\textwidth]{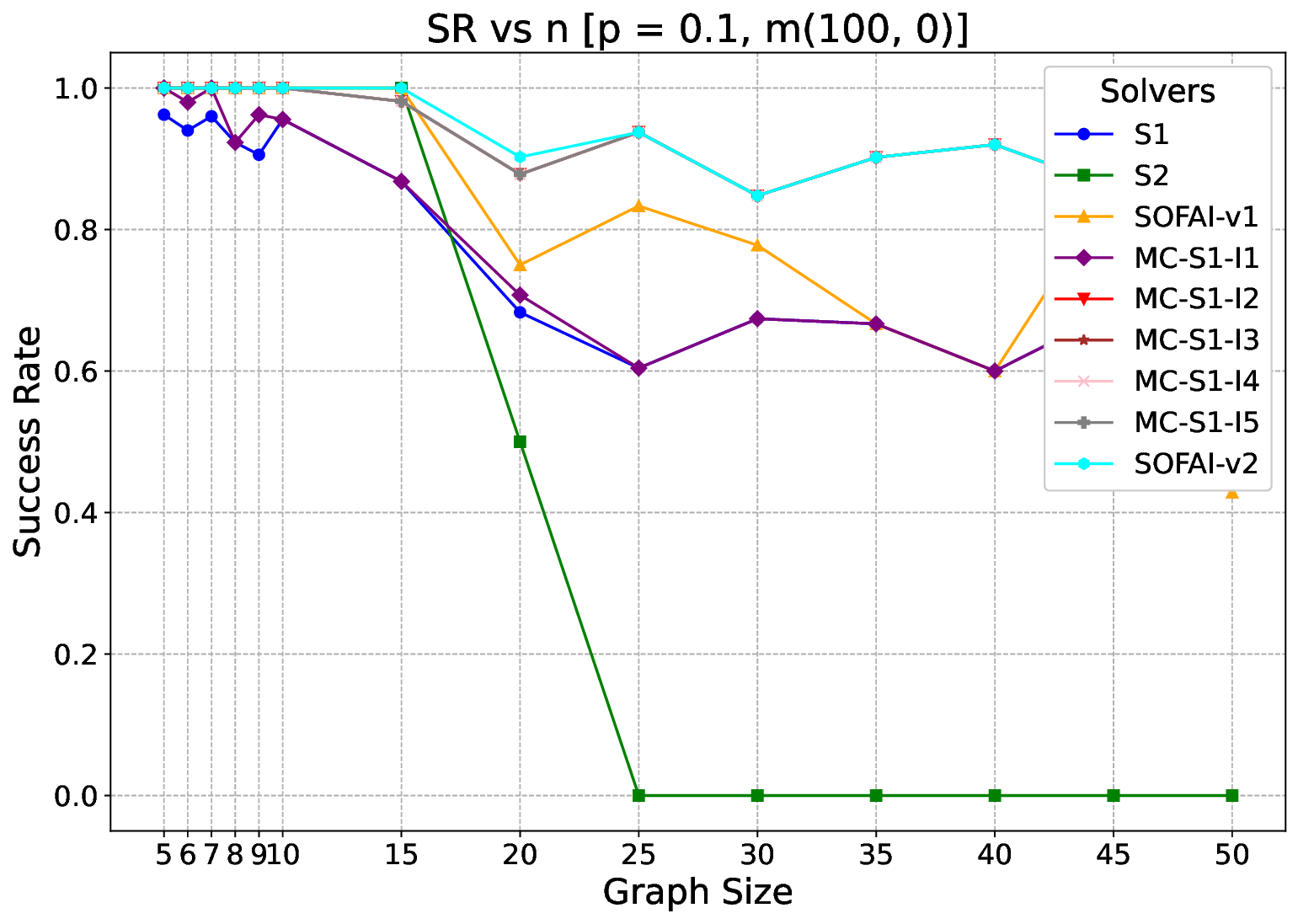}
\label{fig:alls}
\end{subfigure}
\hfill
\begin{subfigure}[b]{0.32\textwidth}
\includegraphics[width=\textwidth]{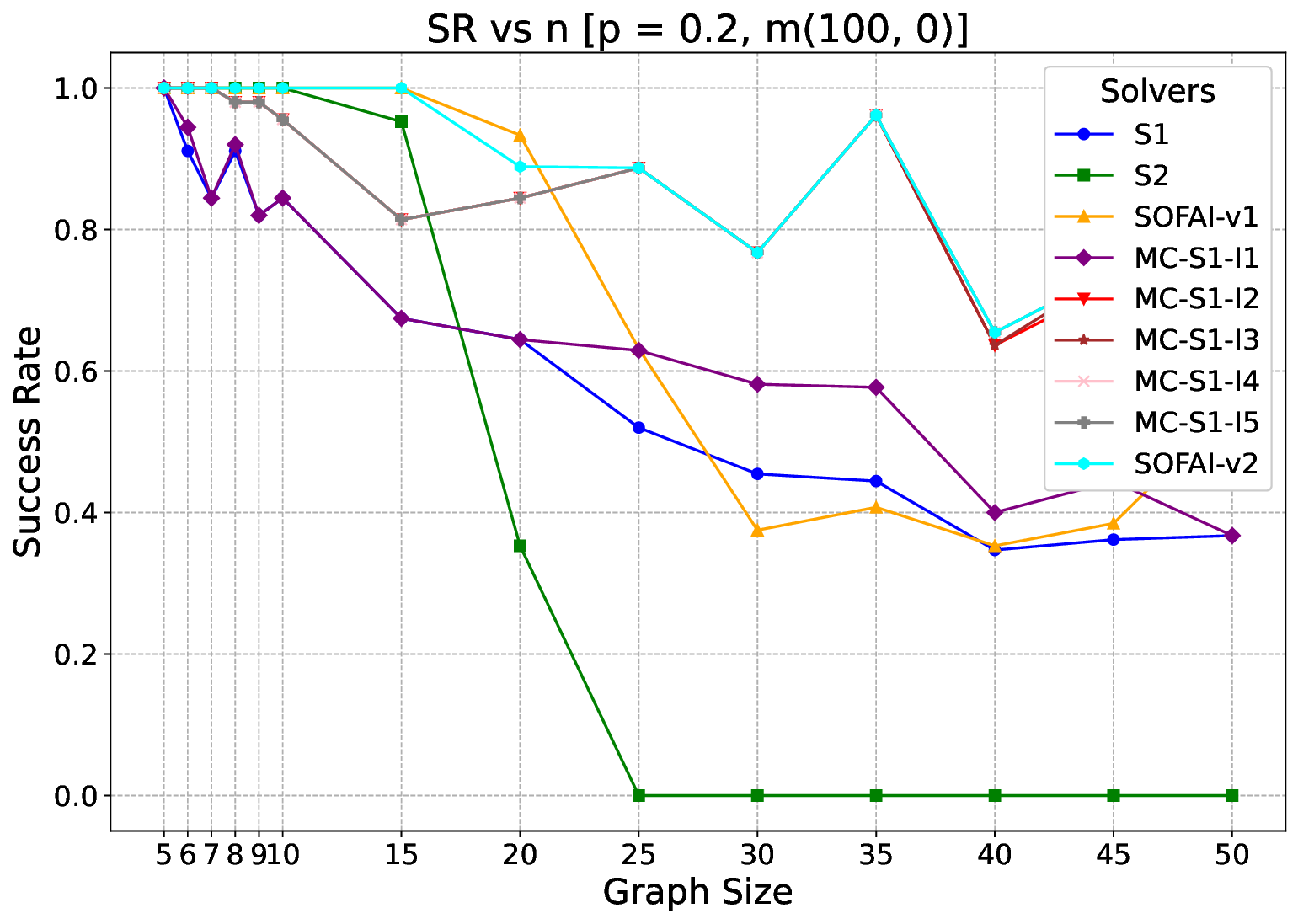}
\label{fig:allus}
\end{subfigure}
\hfill
\begin{subfigure}[b]{0.32\textwidth}
\includegraphics[width=\textwidth]{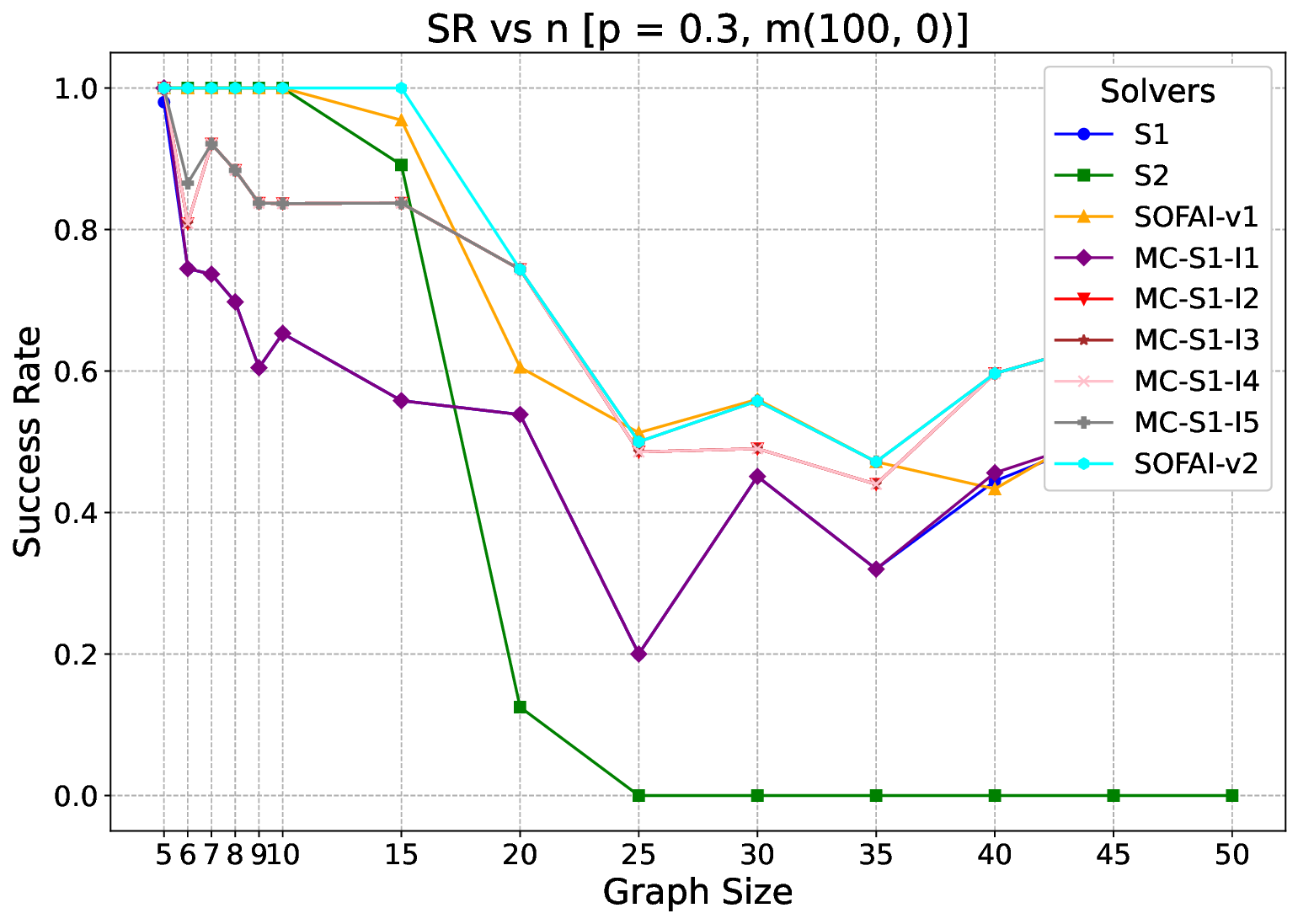}
\label{fig:allhalf}
\end{subfigure}
\begin{subfigure}[b]{0.32\textwidth}
\includegraphics[width=\textwidth]{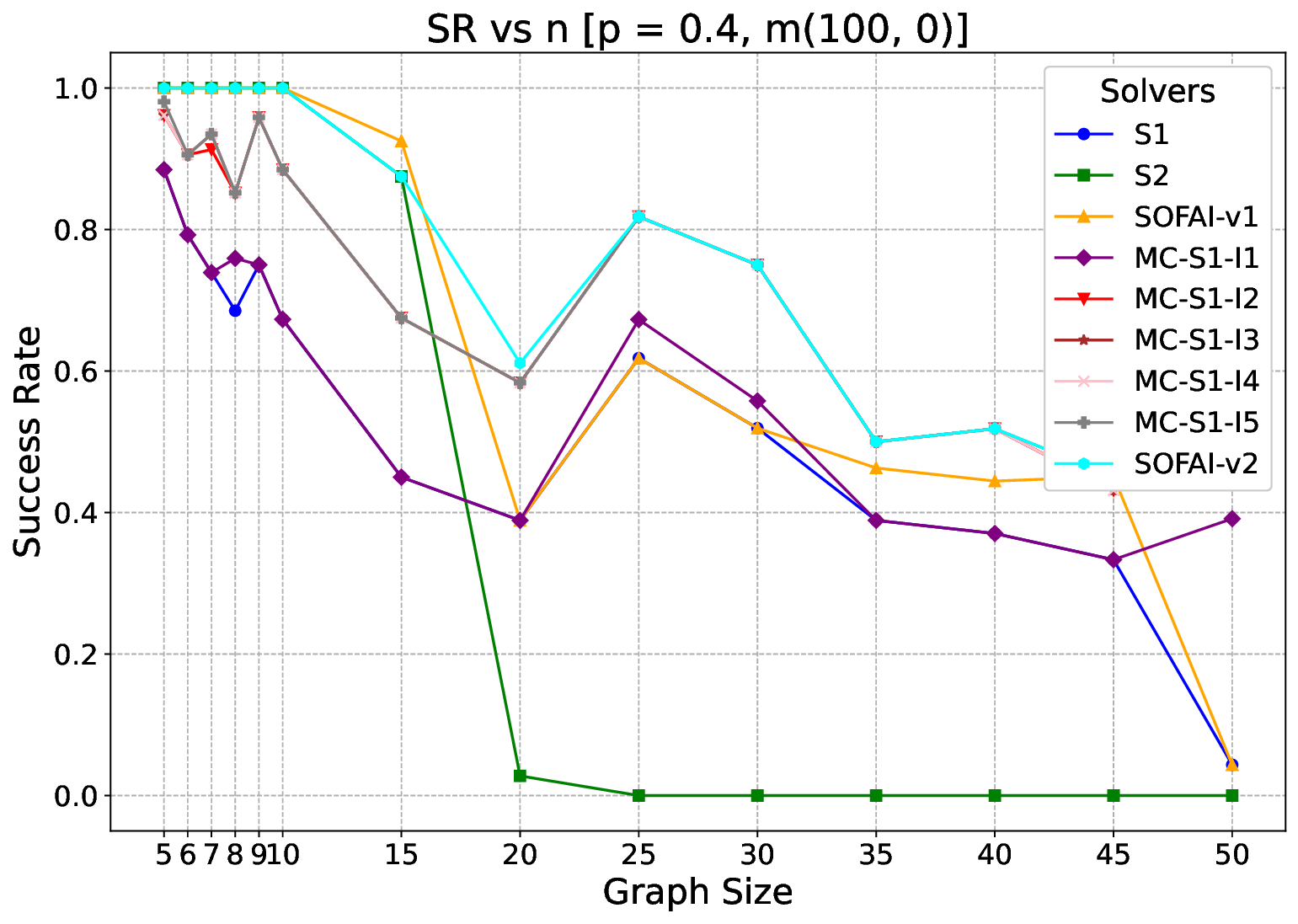}
\label{fig:alls}
\end{subfigure}
\hfill
\begin{subfigure}[b]{0.32\textwidth}
\includegraphics[width=\textwidth]{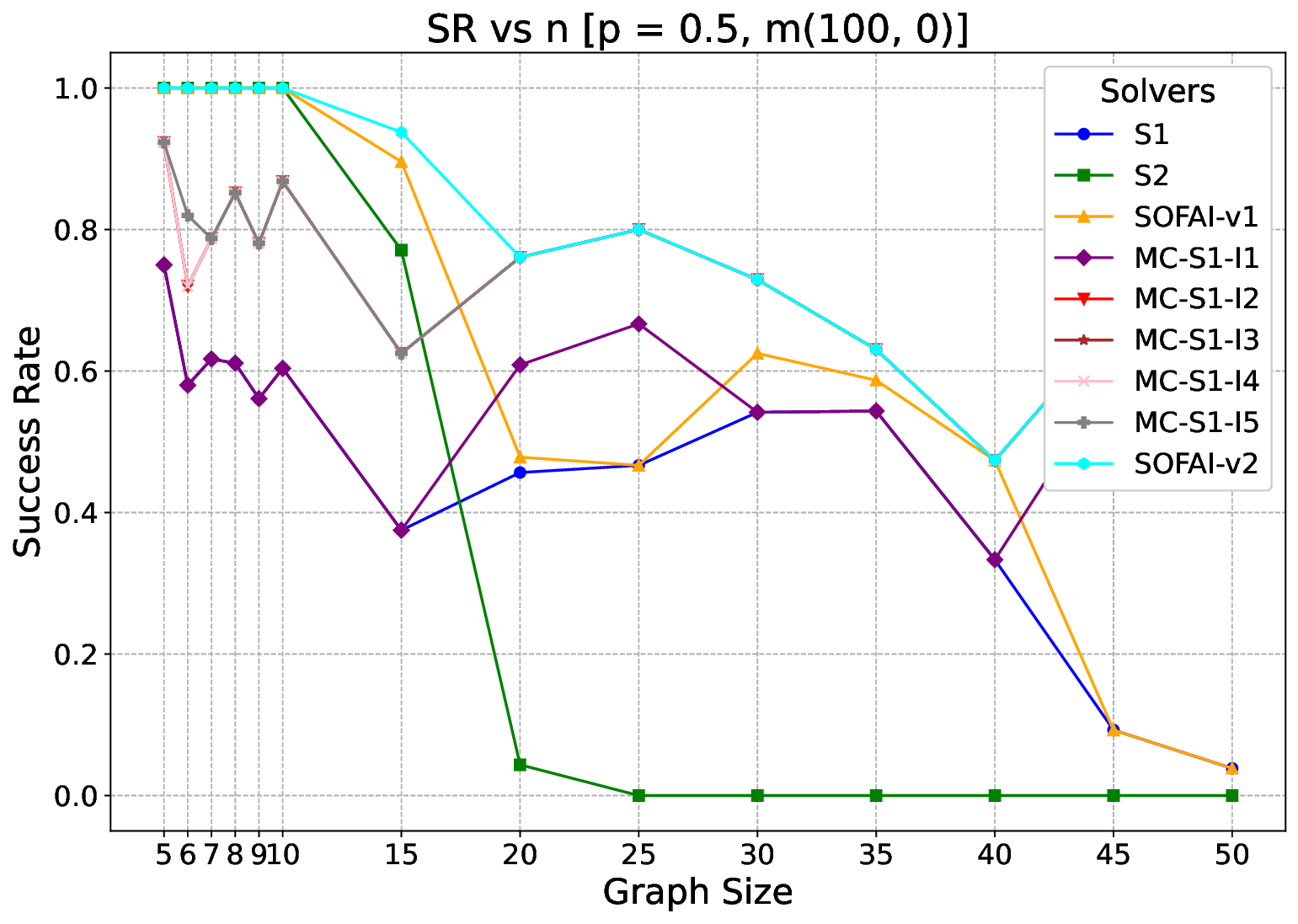}
\label{fig:allus}
\end{subfigure}
\hfill
\begin{subfigure}[b]{0.32\textwidth}
\includegraphics[width=\textwidth]{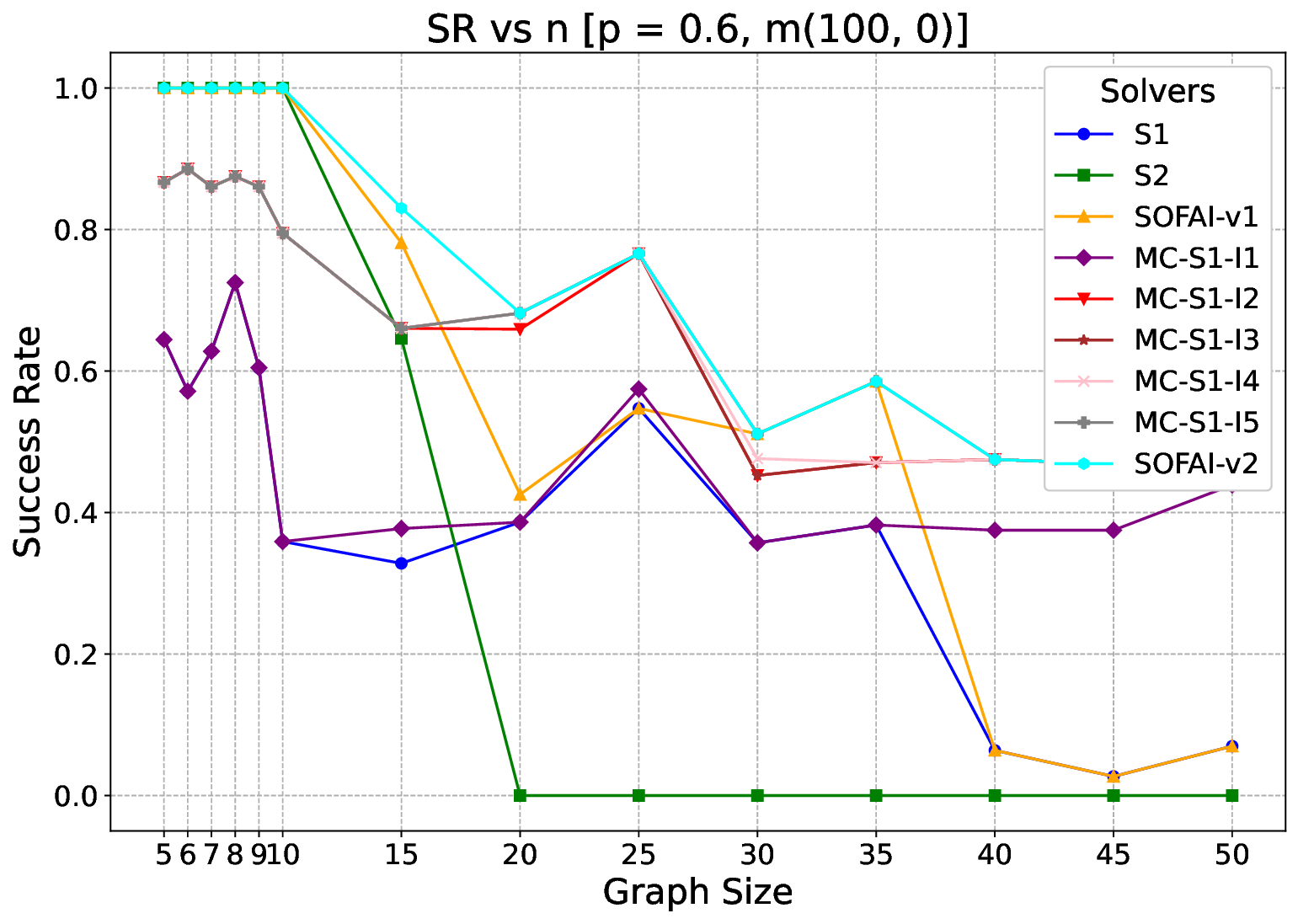}
\label{fig:allhalf}
\end{subfigure}
\begin{subfigure}[b]{0.32\textwidth}
\includegraphics[width=\textwidth]{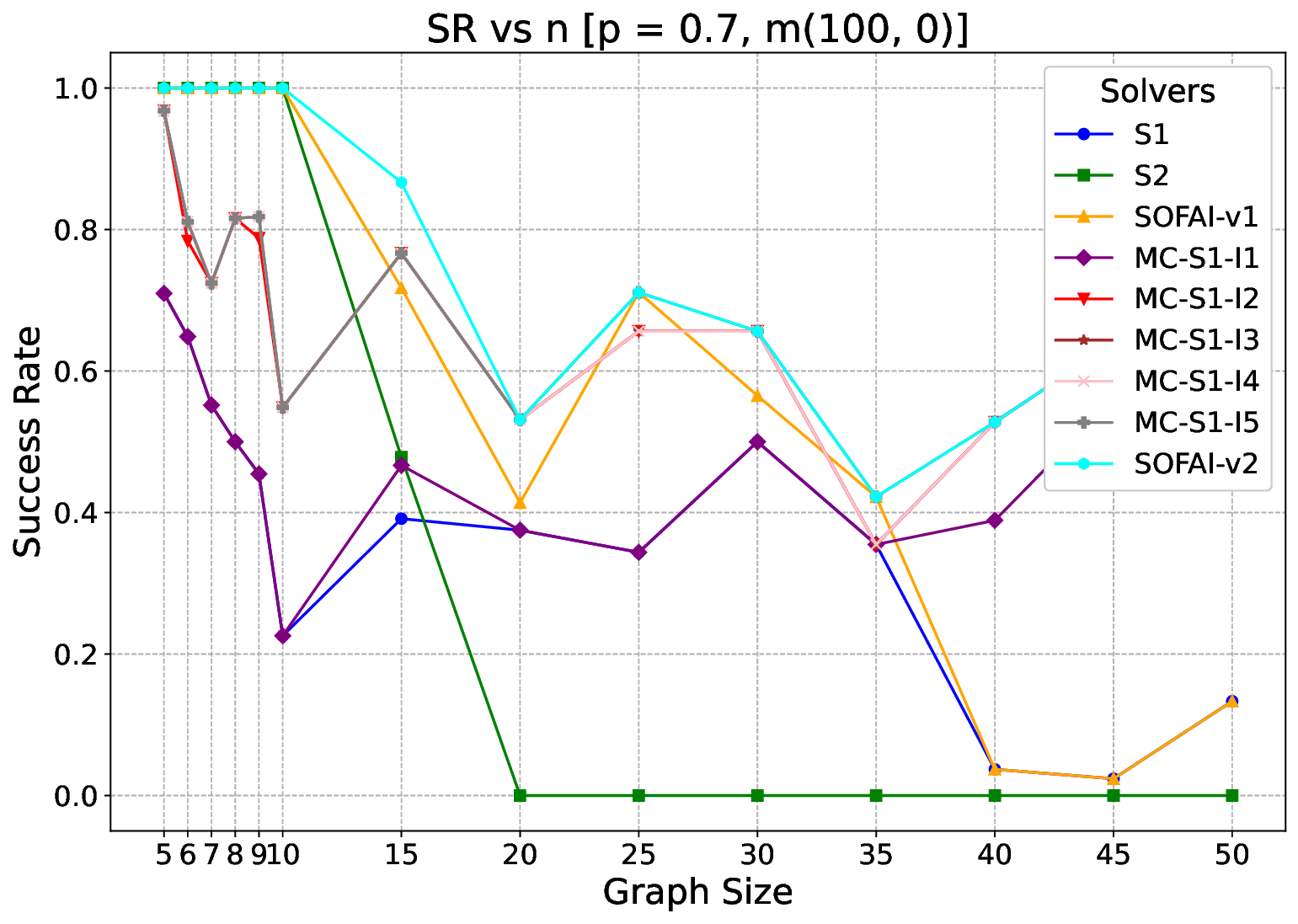}
\label{fig:alls}
\end{subfigure}
\hfill
\begin{subfigure}[b]{0.32\textwidth}
\includegraphics[width=\textwidth]{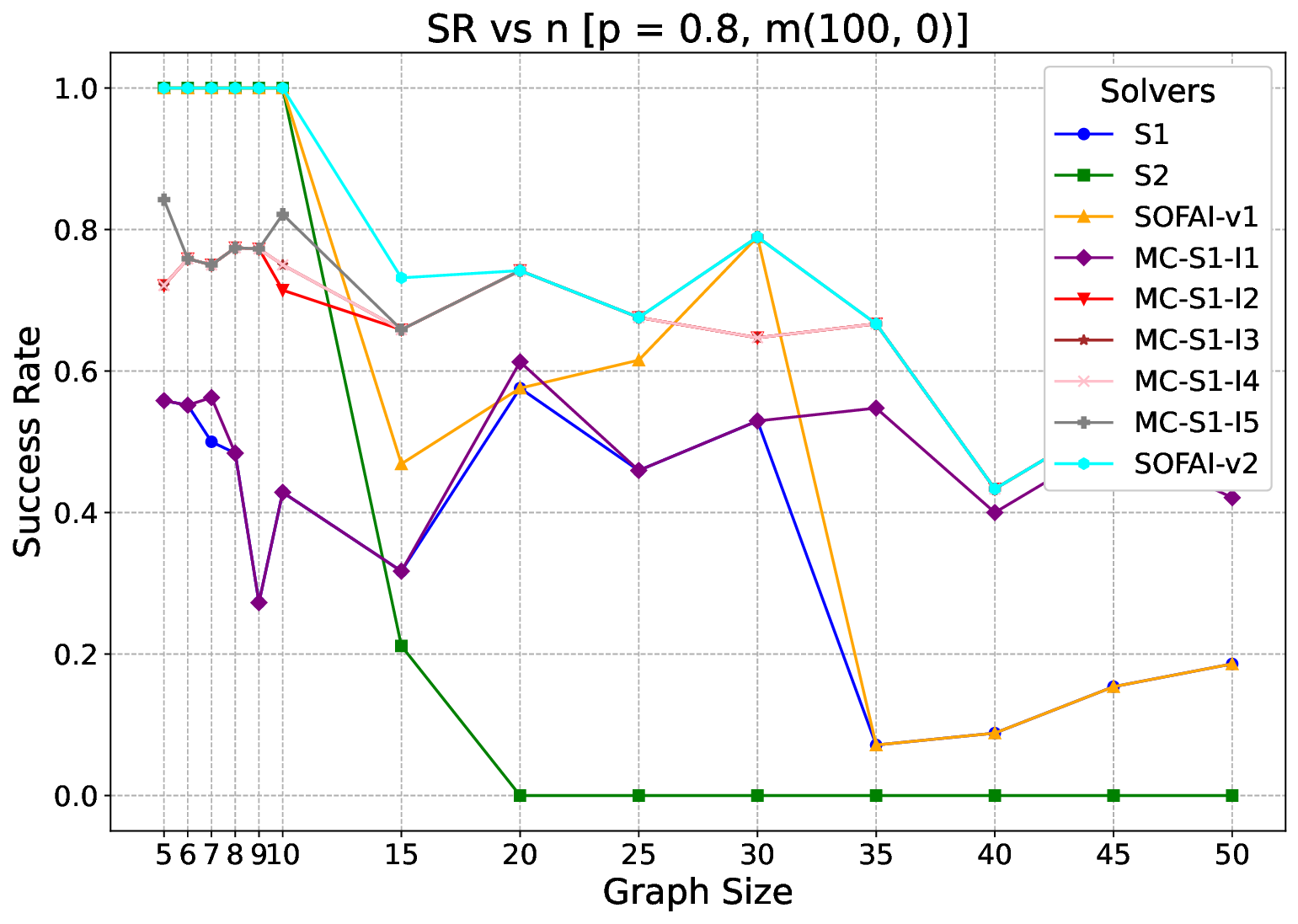}
\label{fig:allus}
\end{subfigure}
\hfill
\begin{subfigure}[b]{0.32\textwidth}
\includegraphics[width=\textwidth]{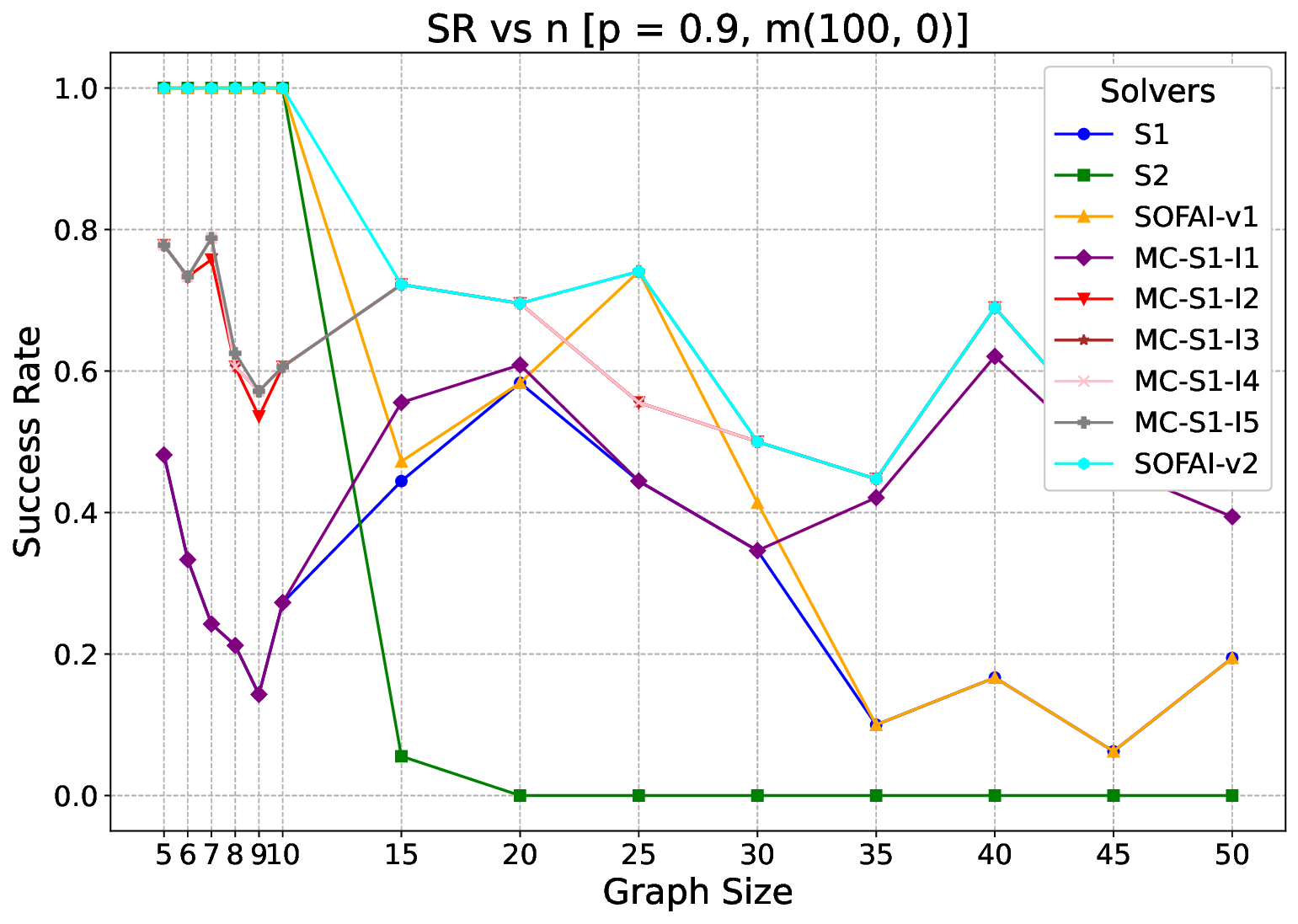}
\label{fig:allhalf}
\end{subfigure}
\caption{Success Rate (SR) of different solvers vs Graph size ($n$) across edge probabilities ($p$) for problem configuration ($m = (0, 100)$)}
\label{fig:sc_un}
\end{figure*}

\begin{figure*}[!htbp]
\centering
\begin{subfigure}[b]{0.32\textwidth}
\includegraphics[width=\textwidth]{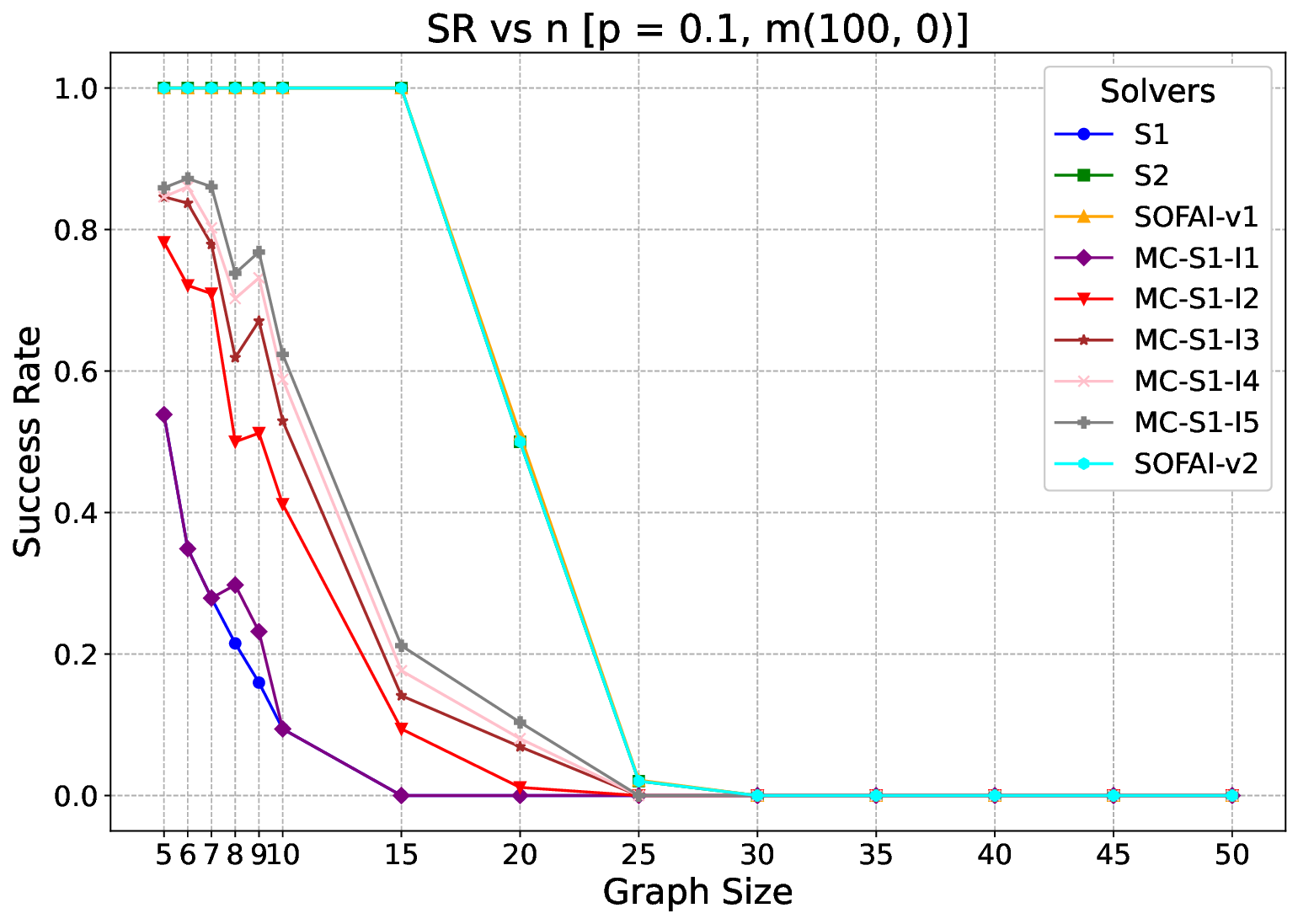}
\label{fig:alls}
\end{subfigure}
\hfill
\begin{subfigure}[b]{0.32\textwidth}
\includegraphics[width=\textwidth]{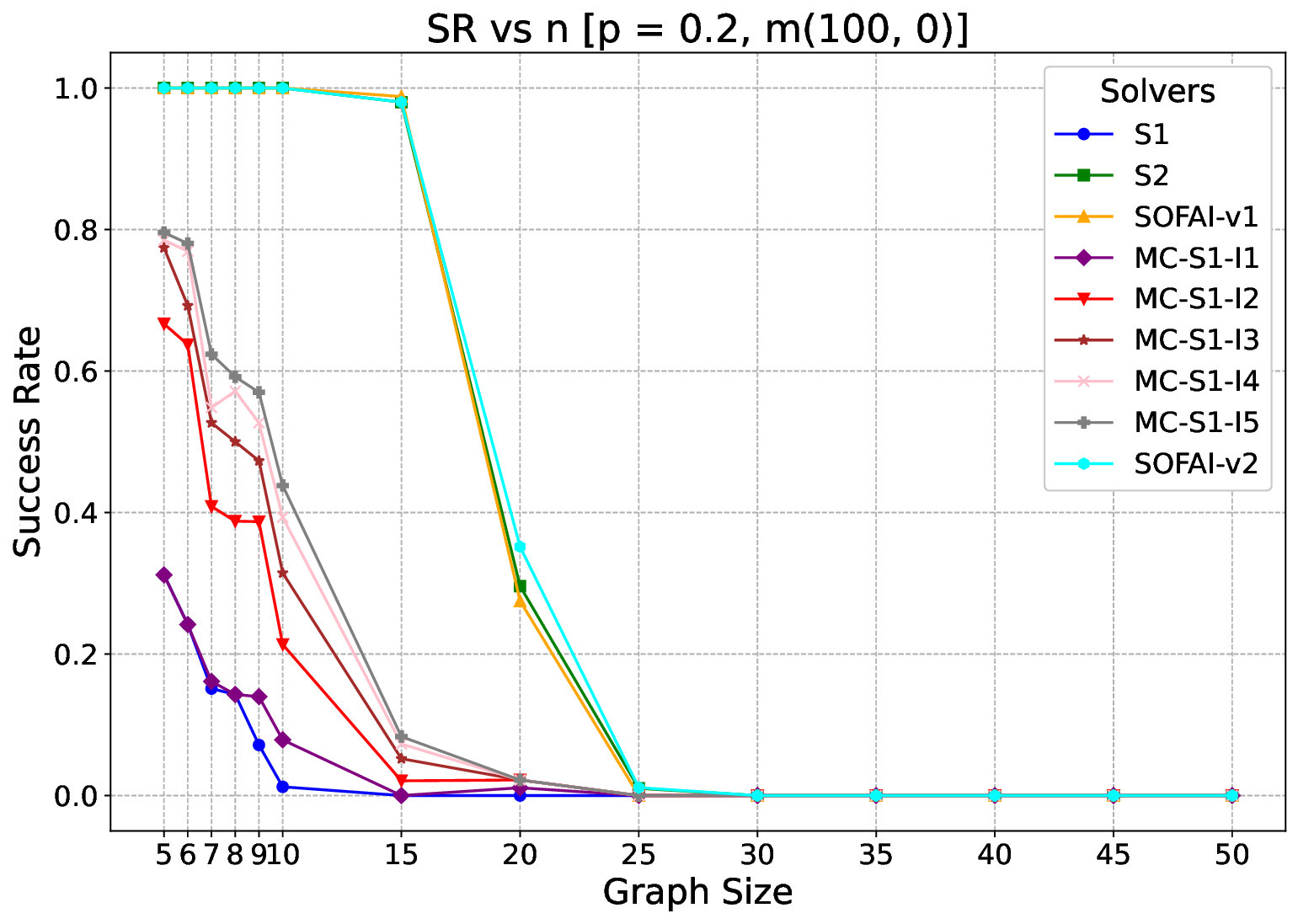}
\label{fig:allus}
\end{subfigure}
\hfill
\begin{subfigure}[b]{0.32\textwidth}
\includegraphics[width=\textwidth]{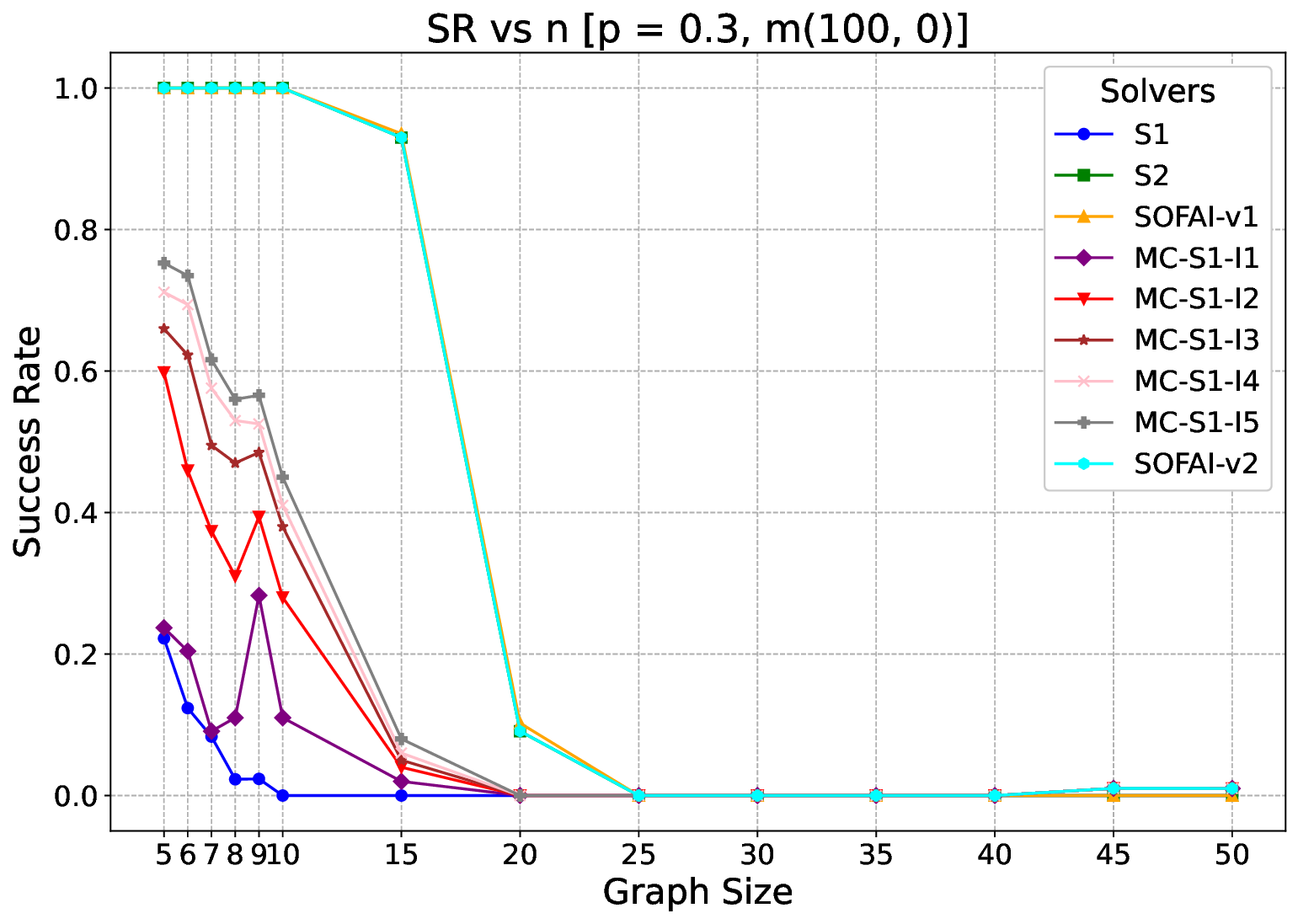}
\label{fig:allhalf}
\end{subfigure}
\begin{subfigure}[b]{0.32\textwidth}
\includegraphics[width=\textwidth]{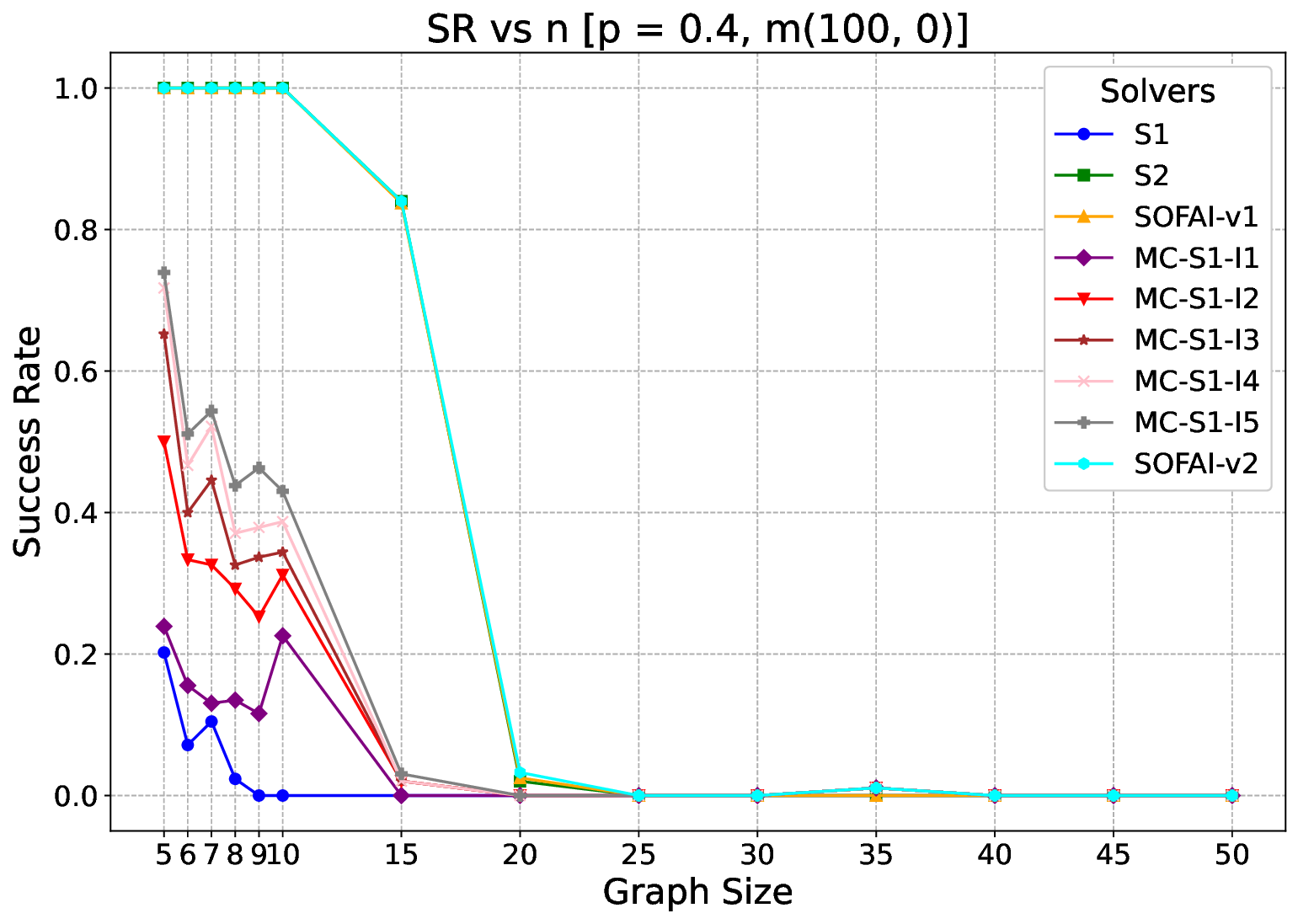}
\label{fig:alls}
\end{subfigure}
\hfill
\begin{subfigure}[b]{0.32\textwidth}
\includegraphics[width=\textwidth]{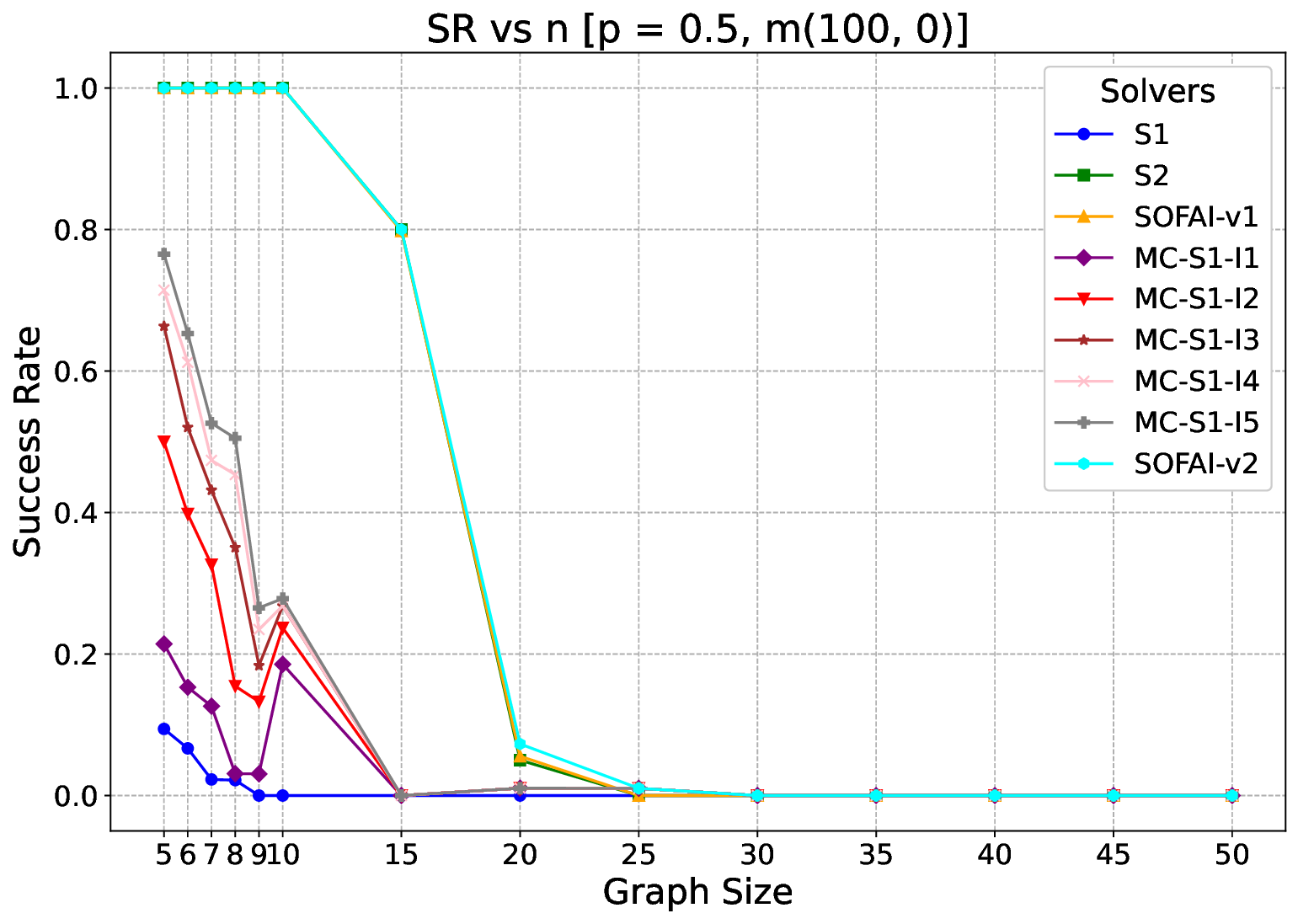}
\label{fig:allus}
\end{subfigure}
\hfill
\begin{subfigure}[b]{0.32\textwidth}
\includegraphics[width=\textwidth]{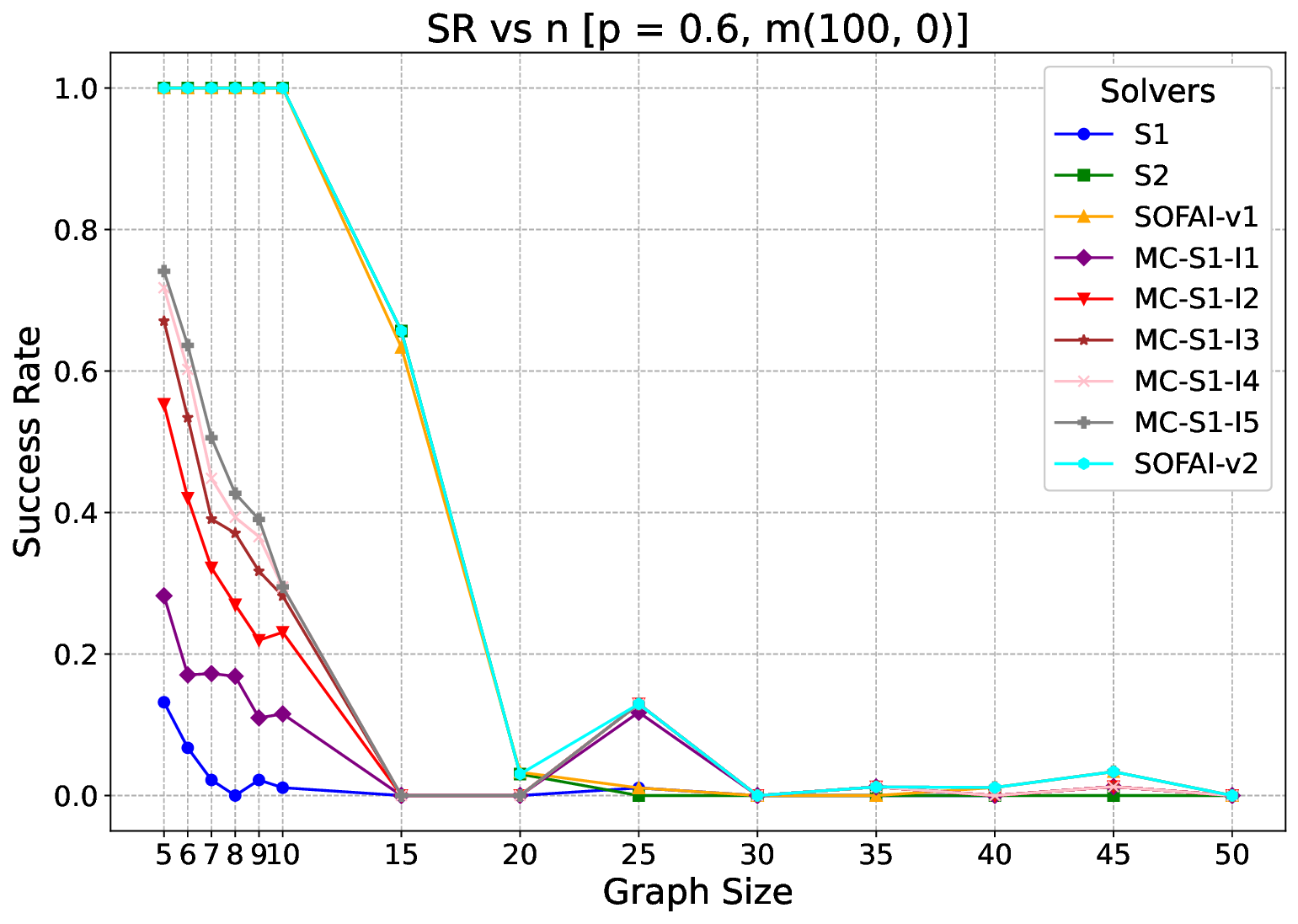}
\label{fig:allhalf}
\end{subfigure}
\begin{subfigure}[b]{0.32\textwidth}
\includegraphics[width=\textwidth]{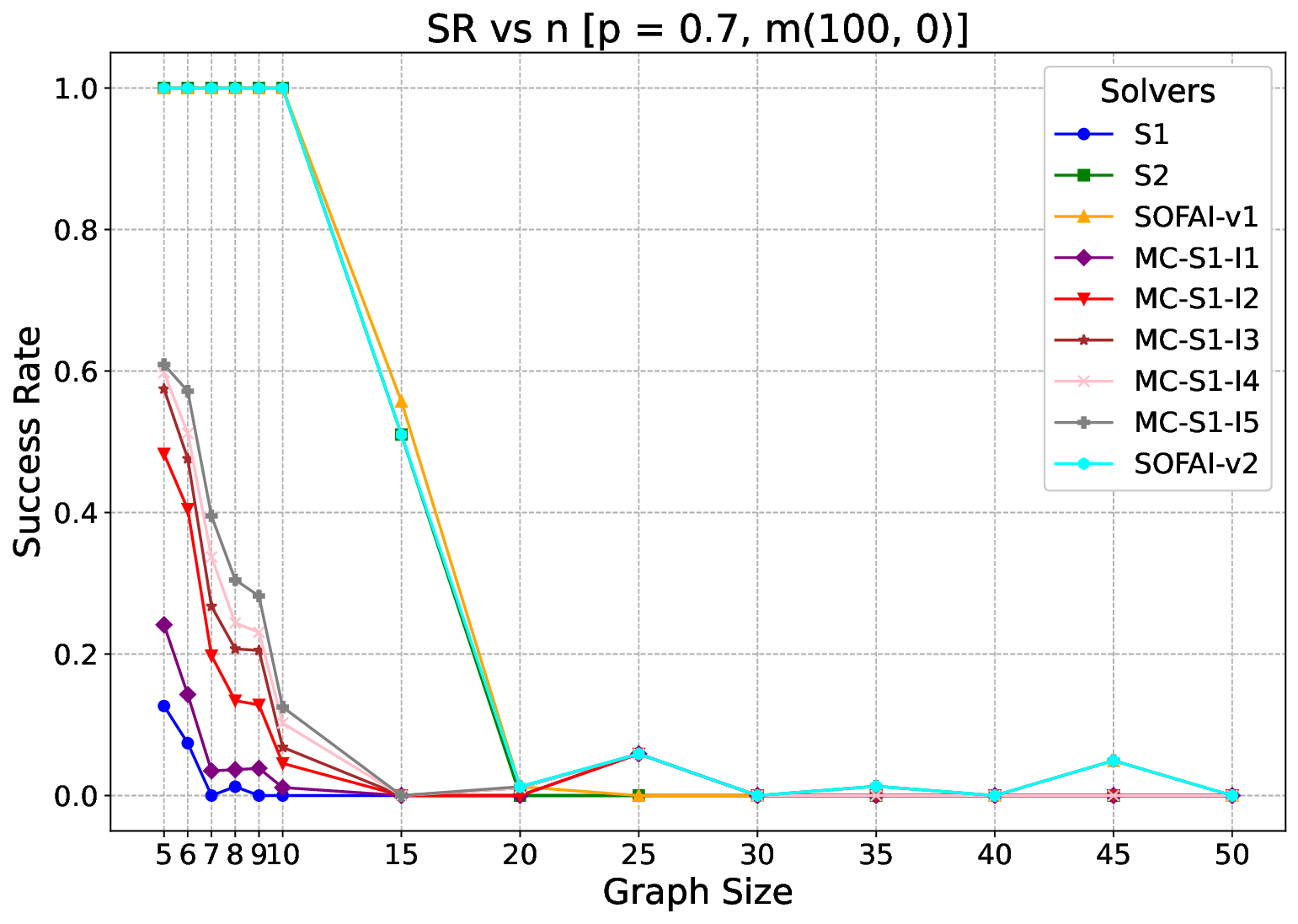}
\label{fig:alls}
\end{subfigure}
\hfill
\begin{subfigure}[b]{0.32\textwidth}
\includegraphics[width=\textwidth]{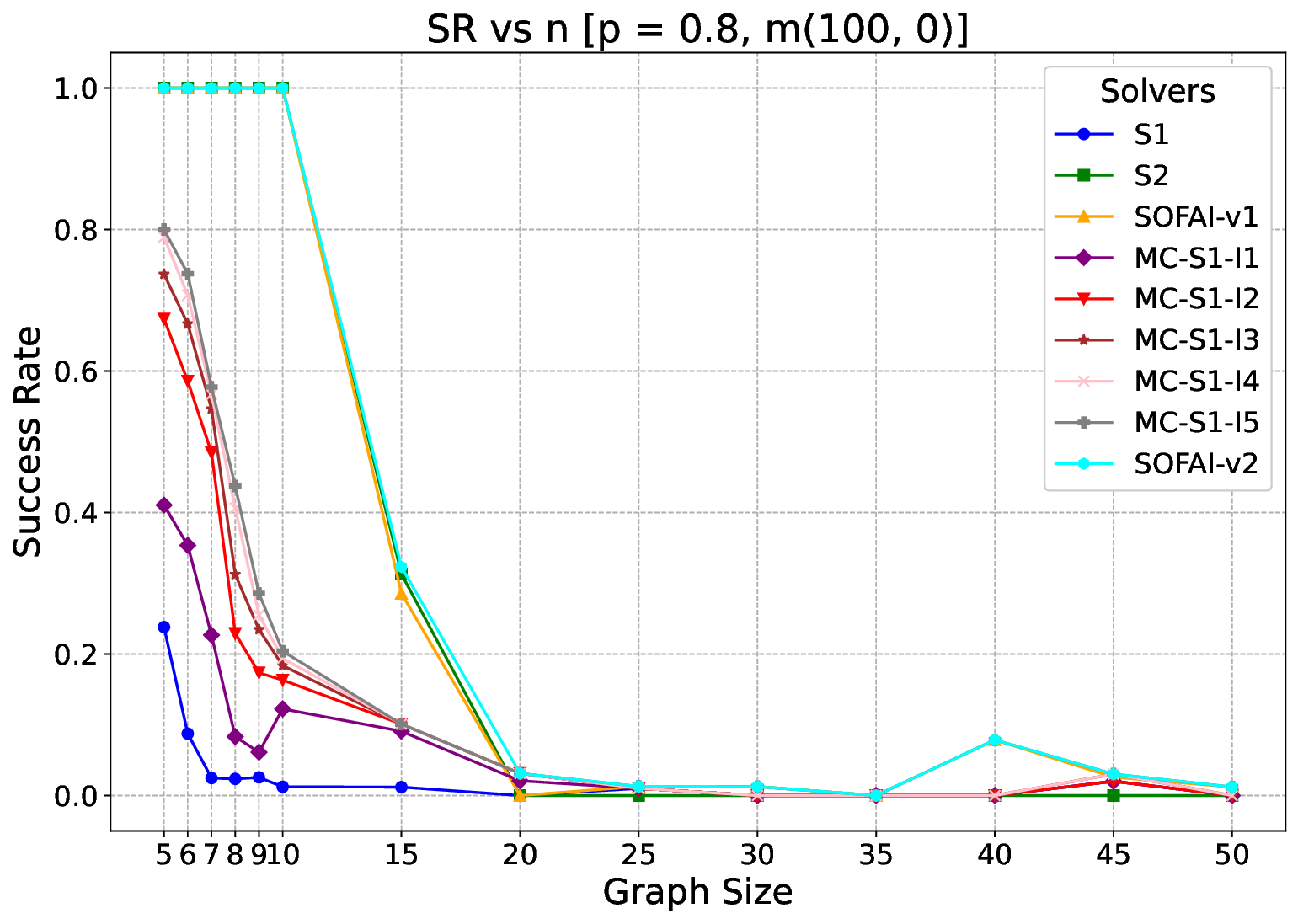}
\label{fig:allus}
\end{subfigure}
\hfill
\begin{subfigure}[b]{0.32\textwidth}
\includegraphics[width=\textwidth]{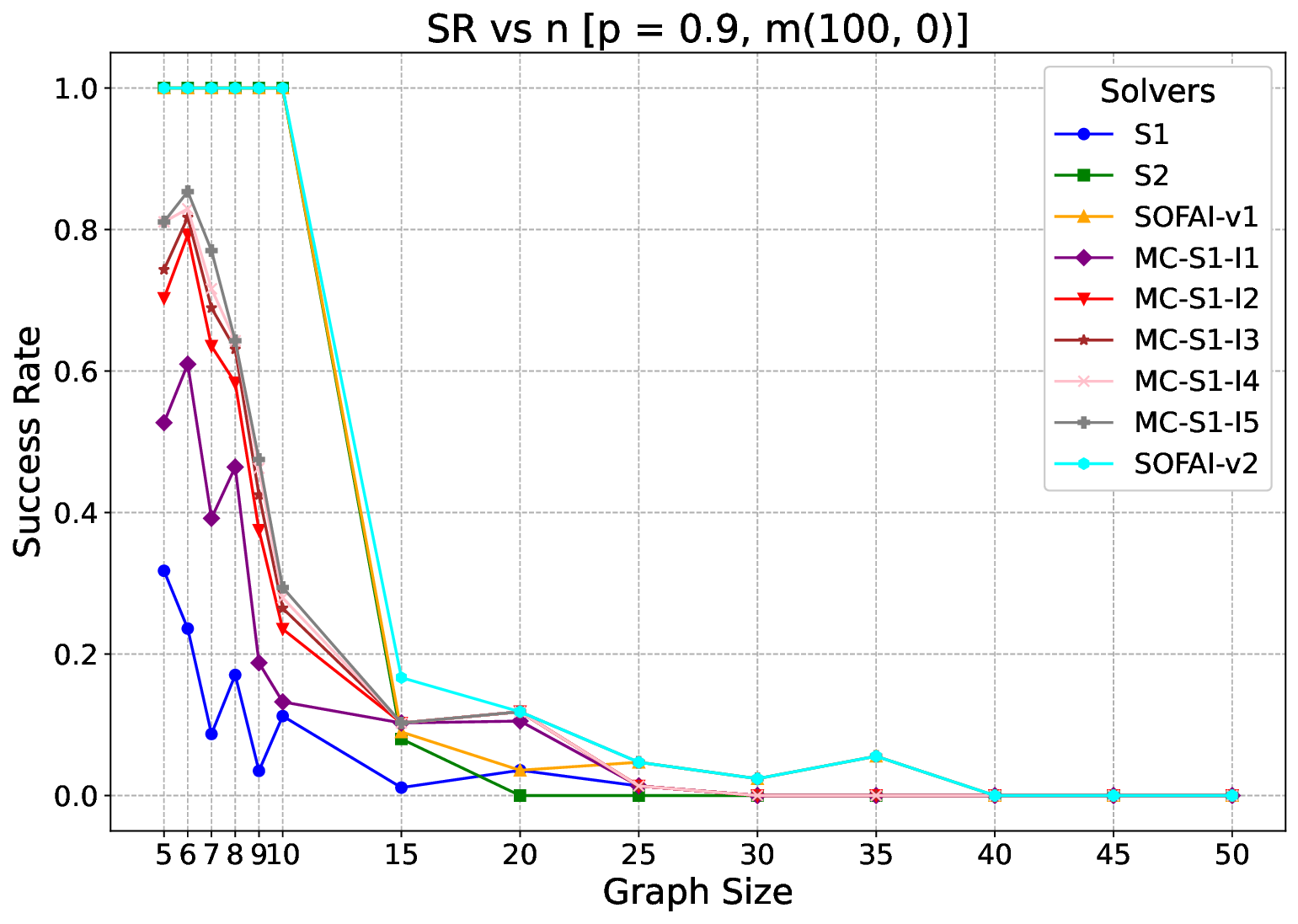}
\label{fig:a}
\end{subfigure}
\caption{Success Rate (SR) of different solvers vs Graph size ($n$) across edge probabilities ($p$) for problem configuration ($m = (100, 0)$)}
\label{fig:sc_s}
\end{figure*}

This section explores the success rates of different solvers in solving graph coloring problems across different problem configurations ($m$). Figure \ref{fig:sc_half}, Figure \ref{fig:sc_un} and Figure \ref{fig:sc_s} demonstrate how each solver—System 1, System 2, SOFAI-v1 and SOFAI-v2 performs across varying edge probabilities ($p$). These graphs show the robustness of SOFAI-v2 in terms of success rate compared to other solvers.

For the above results we set a fixed time limit of 200 seconds per problem. If a solver does not finish within this time, it is counted as a failure.

\section{Results - Average Time Taken}
\label{appendix:averagetime}
\begin{figure*}[!htbp]
\centering
\begin{subfigure}[b]{0.32\textwidth}
\includegraphics[width=\textwidth]{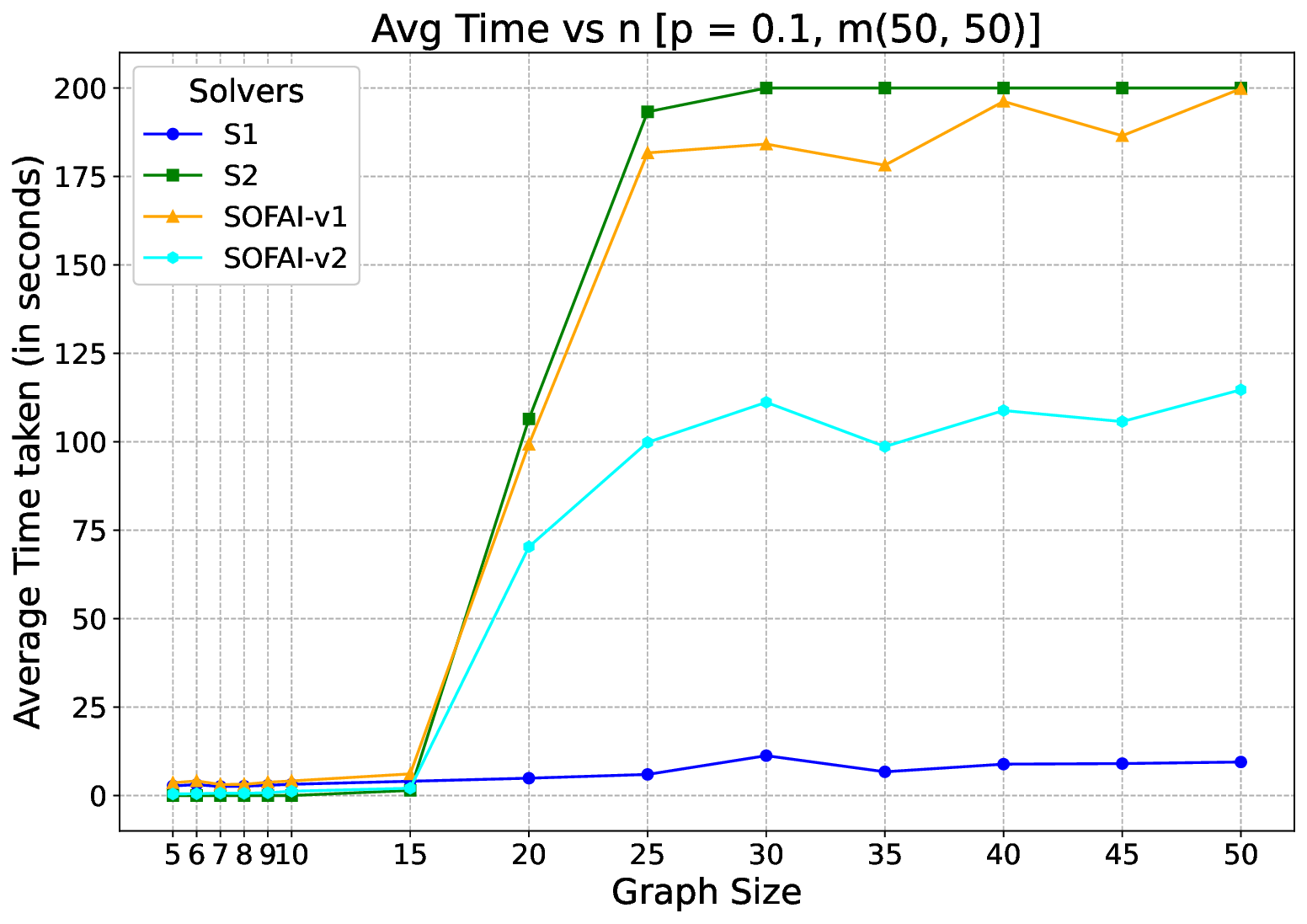}
\label{fig:alls}
\end{subfigure}
\hfill
\begin{subfigure}[b]{0.32\textwidth}
\includegraphics[width=\textwidth]{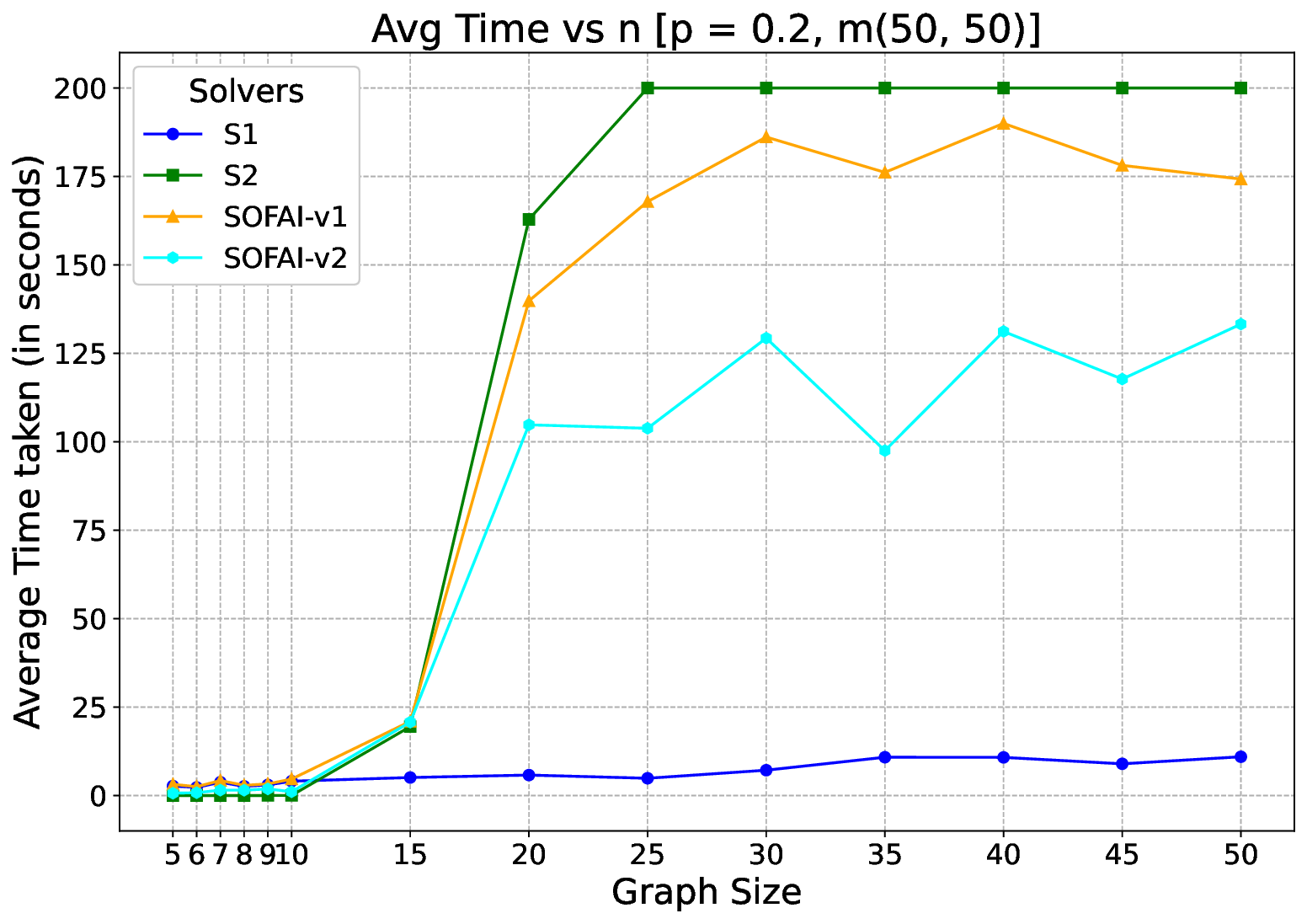}
\label{fig:allus}
\end{subfigure}
\hfill
\begin{subfigure}[b]{0.32\textwidth}
\includegraphics[width=\textwidth]{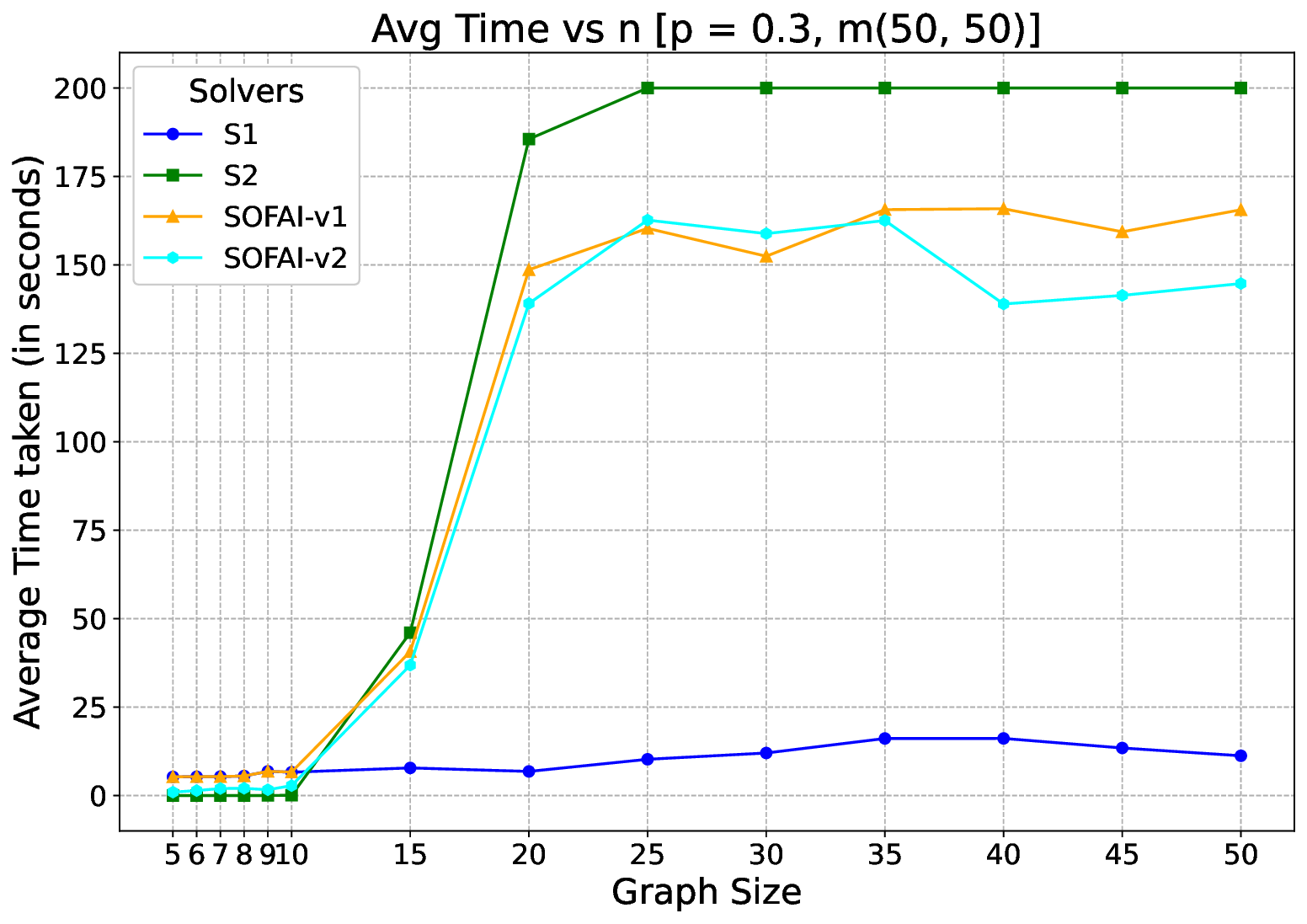}
\label{fig:allhalf}
\end{subfigure}
\begin{subfigure}[b]{0.32\textwidth}
\includegraphics[width=\textwidth]{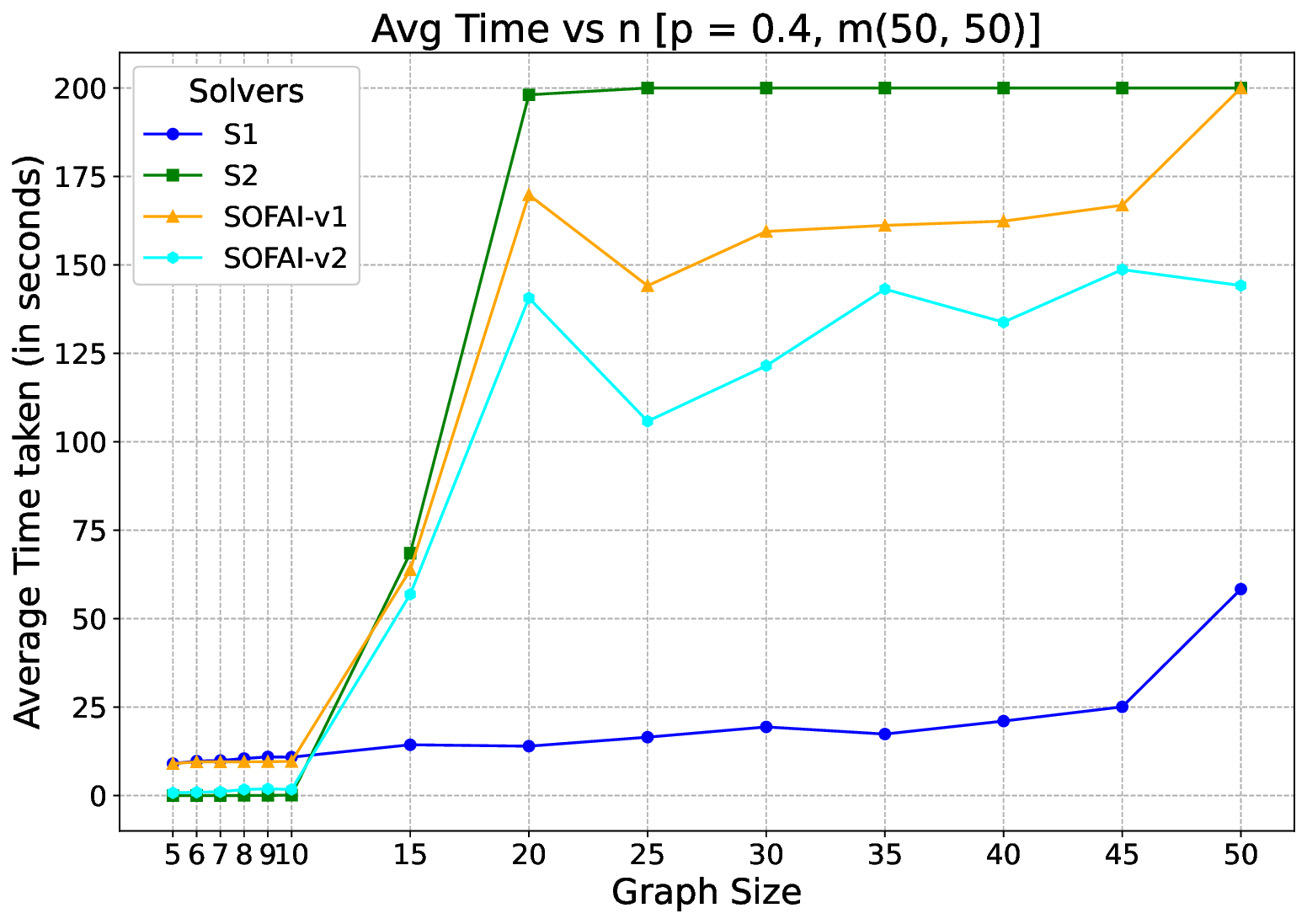}
\label{fig:alls}
\end{subfigure}
\hfill
\begin{subfigure}[b]{0.32\textwidth}
\includegraphics[width=\textwidth]{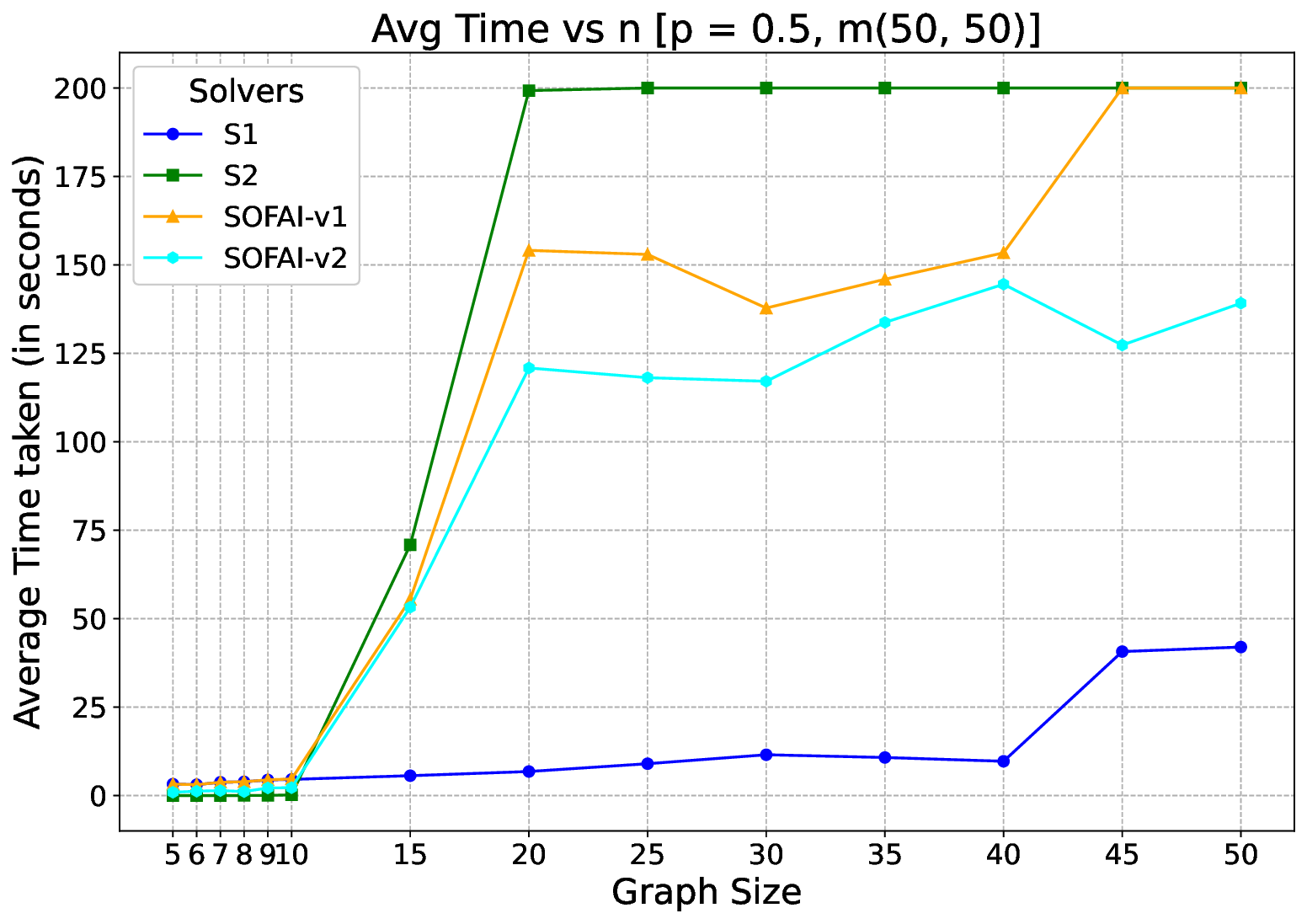}
\label{fig:allus}
\end{subfigure}
\hfill
\begin{subfigure}[b]{0.32\textwidth}
\includegraphics[width=\textwidth]{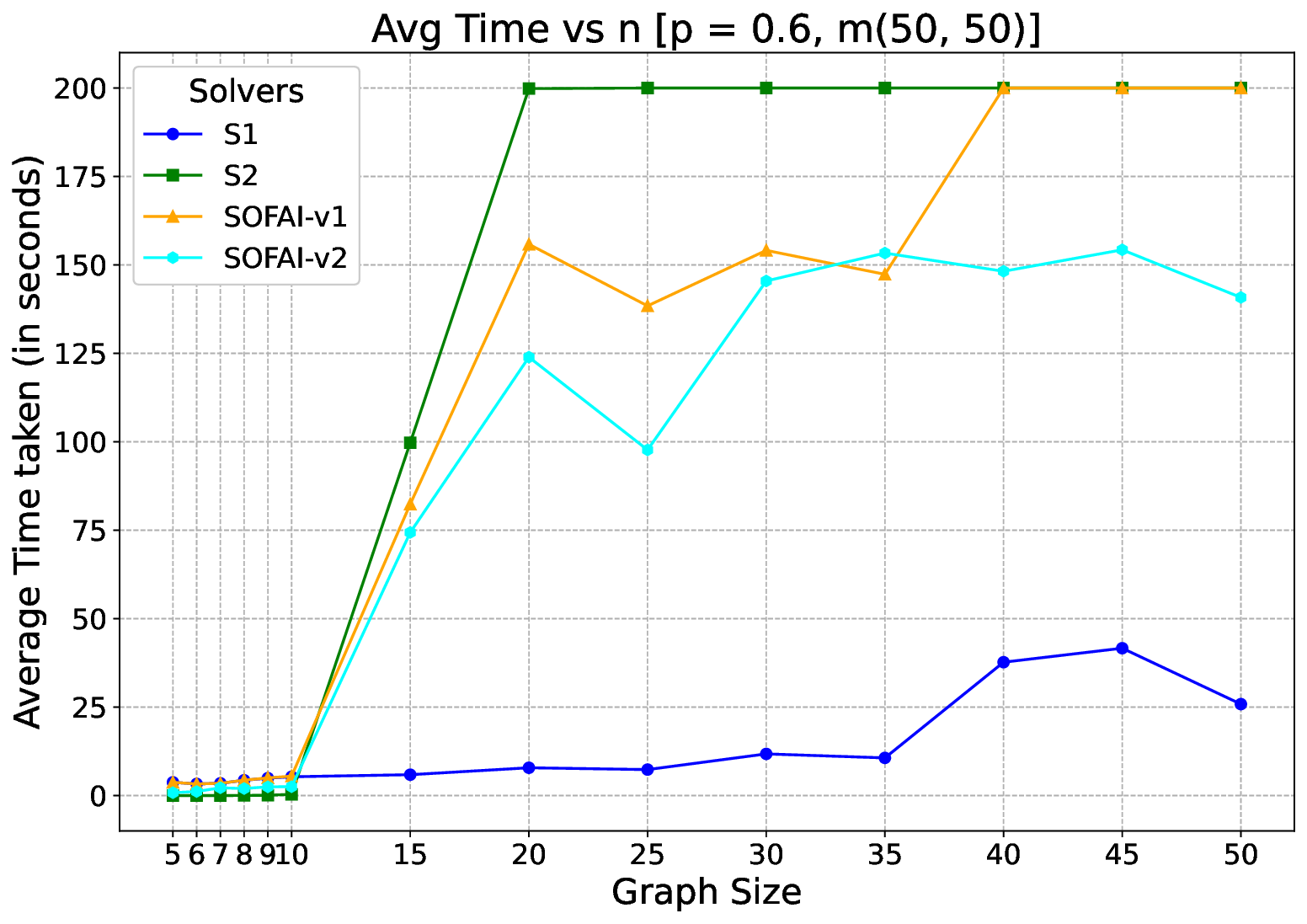}
\label{fig:allhalf}
\end{subfigure}
\begin{subfigure}[b]{0.32\textwidth}
\includegraphics[width=\textwidth]{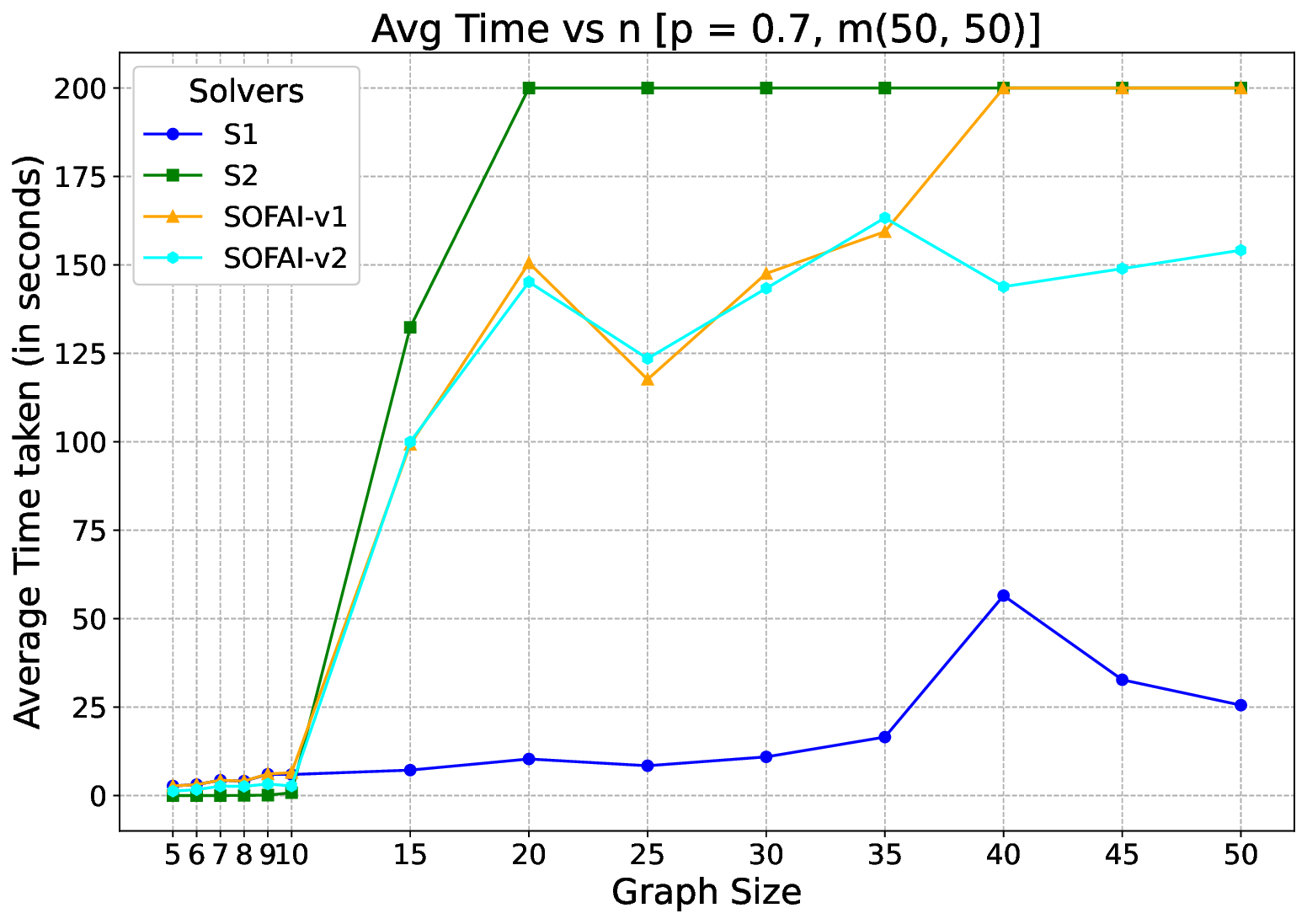}
\label{fig:alls}
\end{subfigure}
\hfill
\begin{subfigure}[b]{0.32\textwidth}
\includegraphics[width=\textwidth]{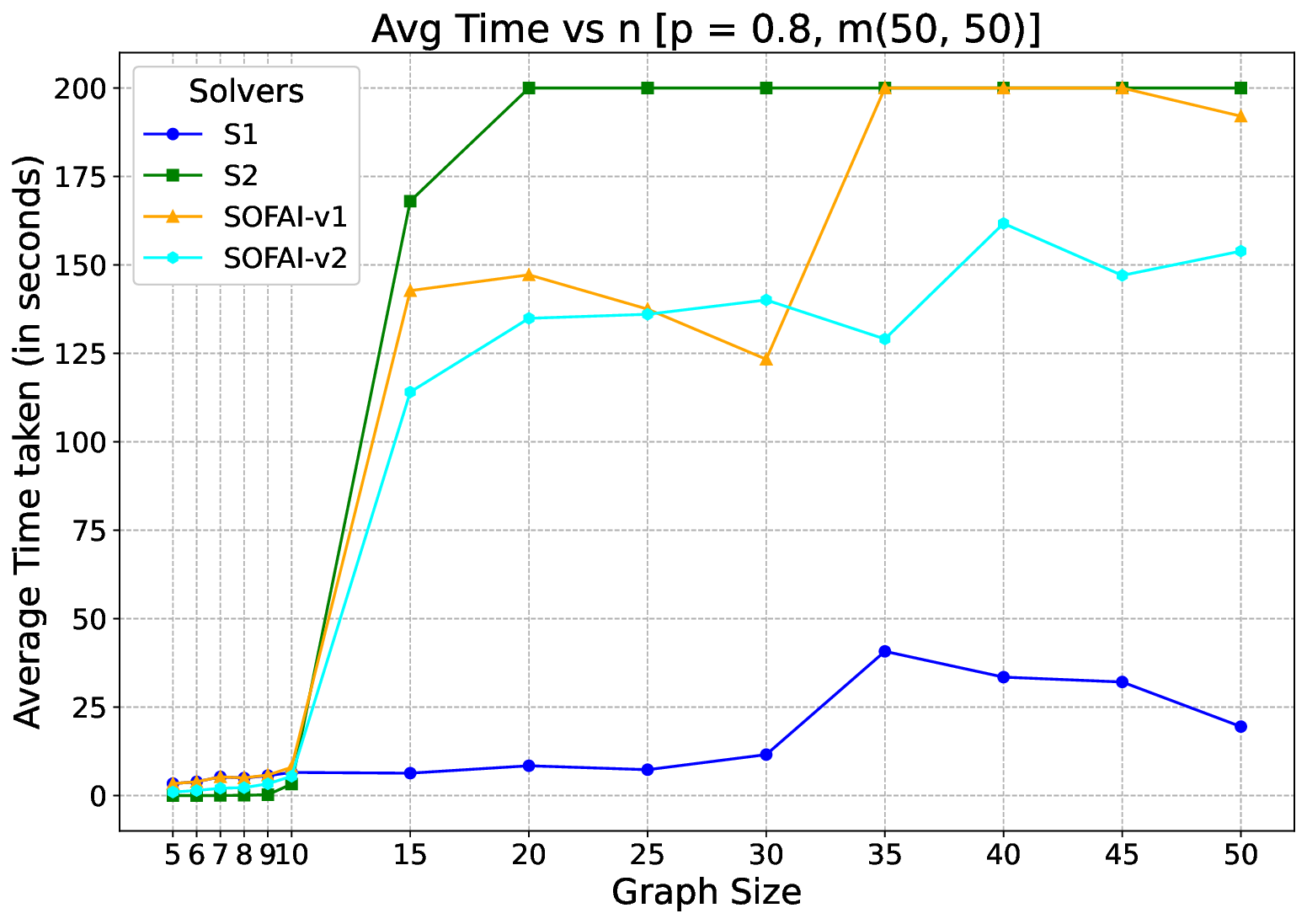}
\label{fig:allus}
\end{subfigure}
\hfill
\begin{subfigure}[b]{0.32\textwidth}
\includegraphics[width=\textwidth]{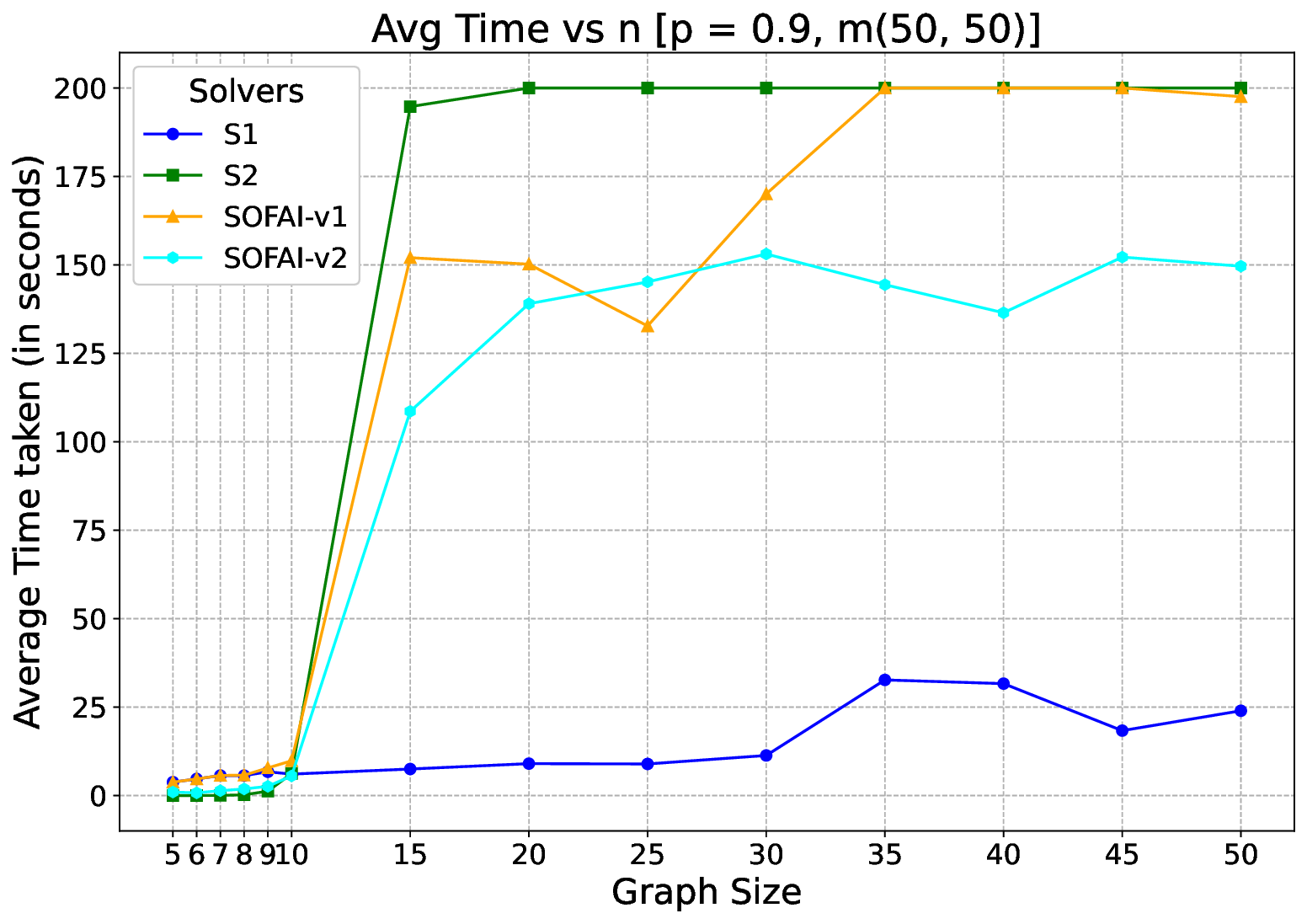}
\label{fig:allhalf}
\end{subfigure}
\caption{Average time of different solvers vs Graph size ($n$) across edge probabilities ($p$) for problem configuration ($m = (50, 50)$)}
\label{fig:t_half}
\end{figure*}

\begin{figure*}[!htbp]
\centering
\begin{subfigure}[b]{0.32\textwidth}
\includegraphics[width=\textwidth]{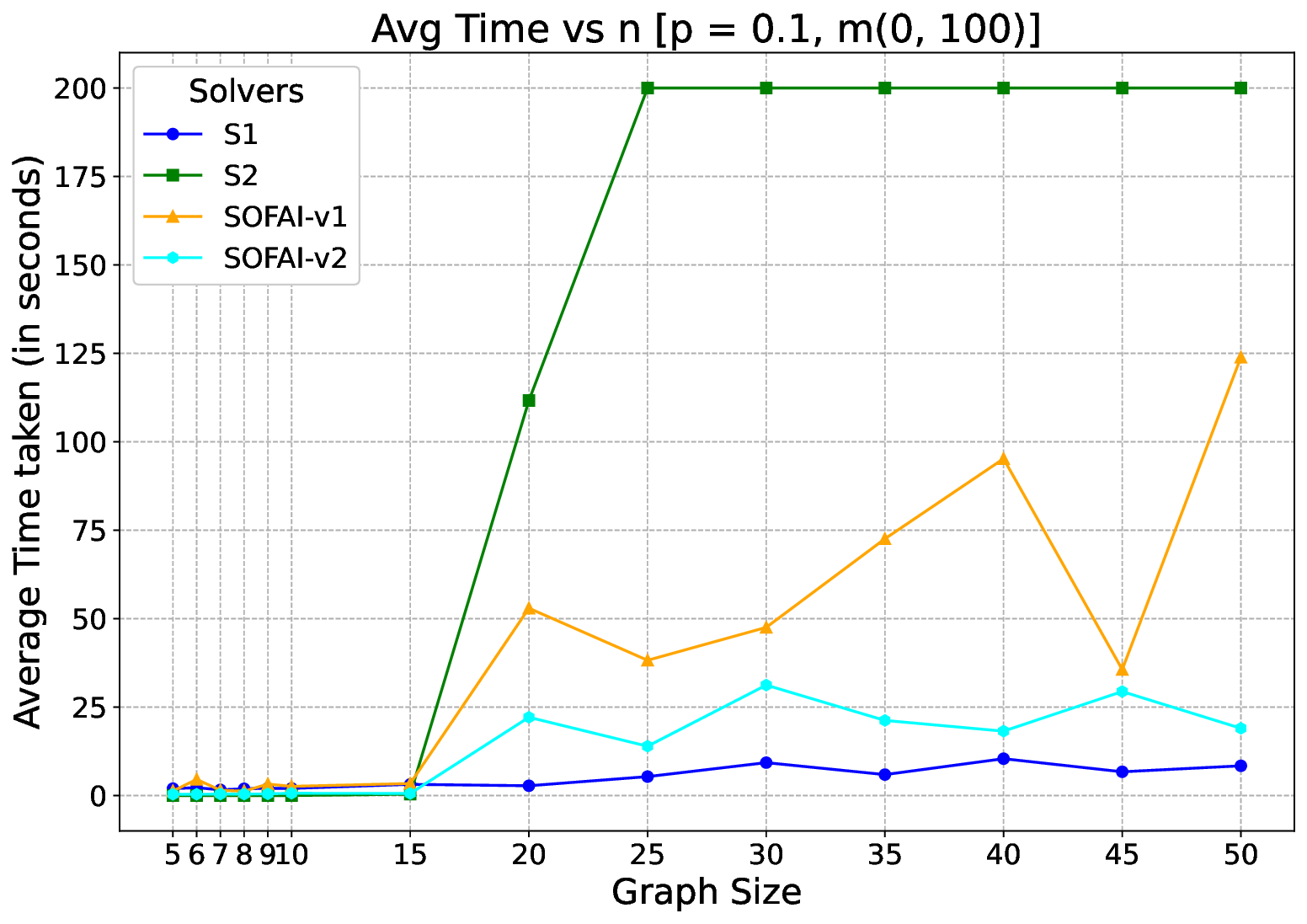}
\label{fig:alls}
\end{subfigure}
\hfill
\begin{subfigure}[b]{0.32\textwidth}
\includegraphics[width=\textwidth]{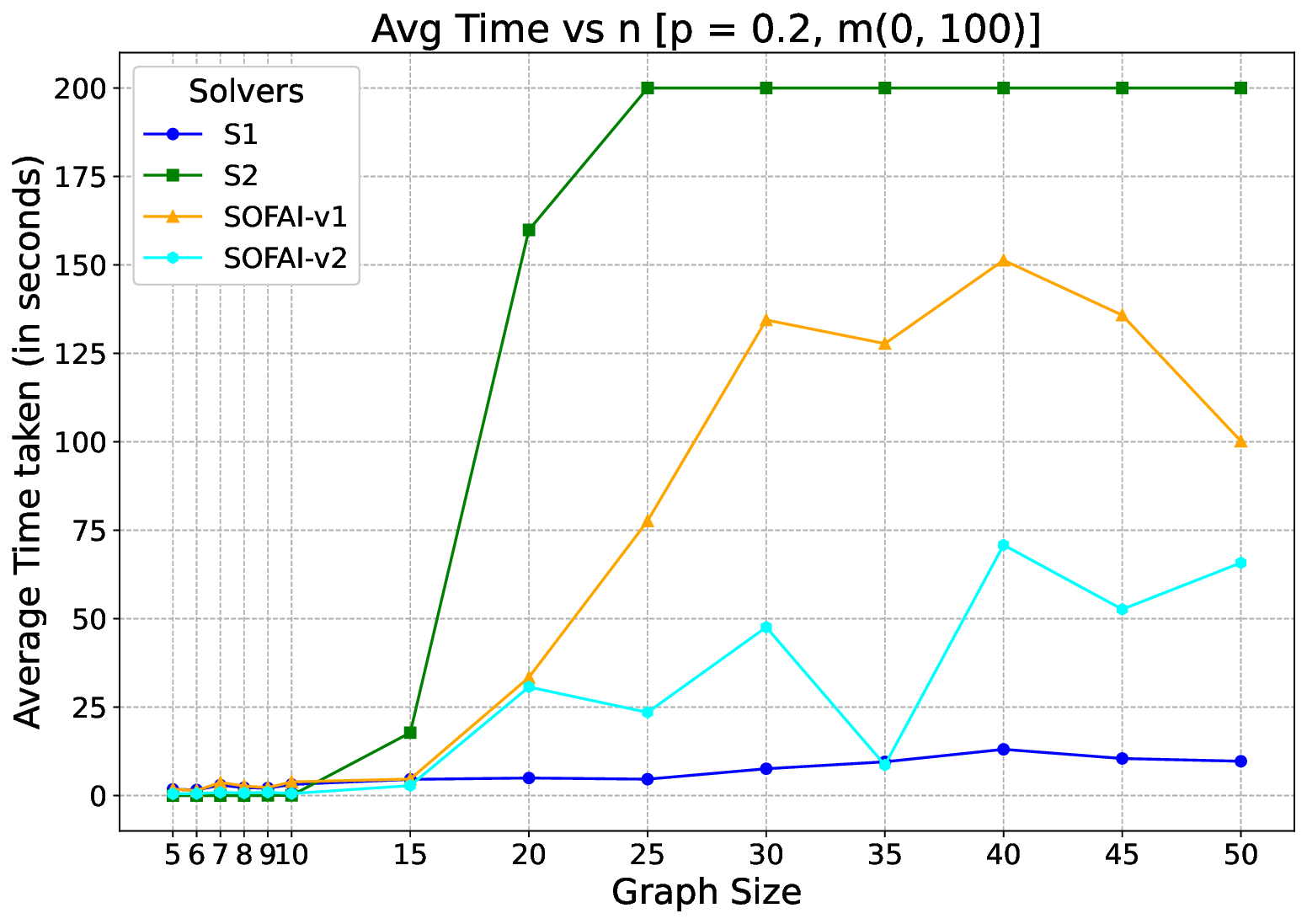}
\label{fig:allus}
\end{subfigure}
\hfill
\begin{subfigure}[b]{0.32\textwidth}
\includegraphics[width=\textwidth]{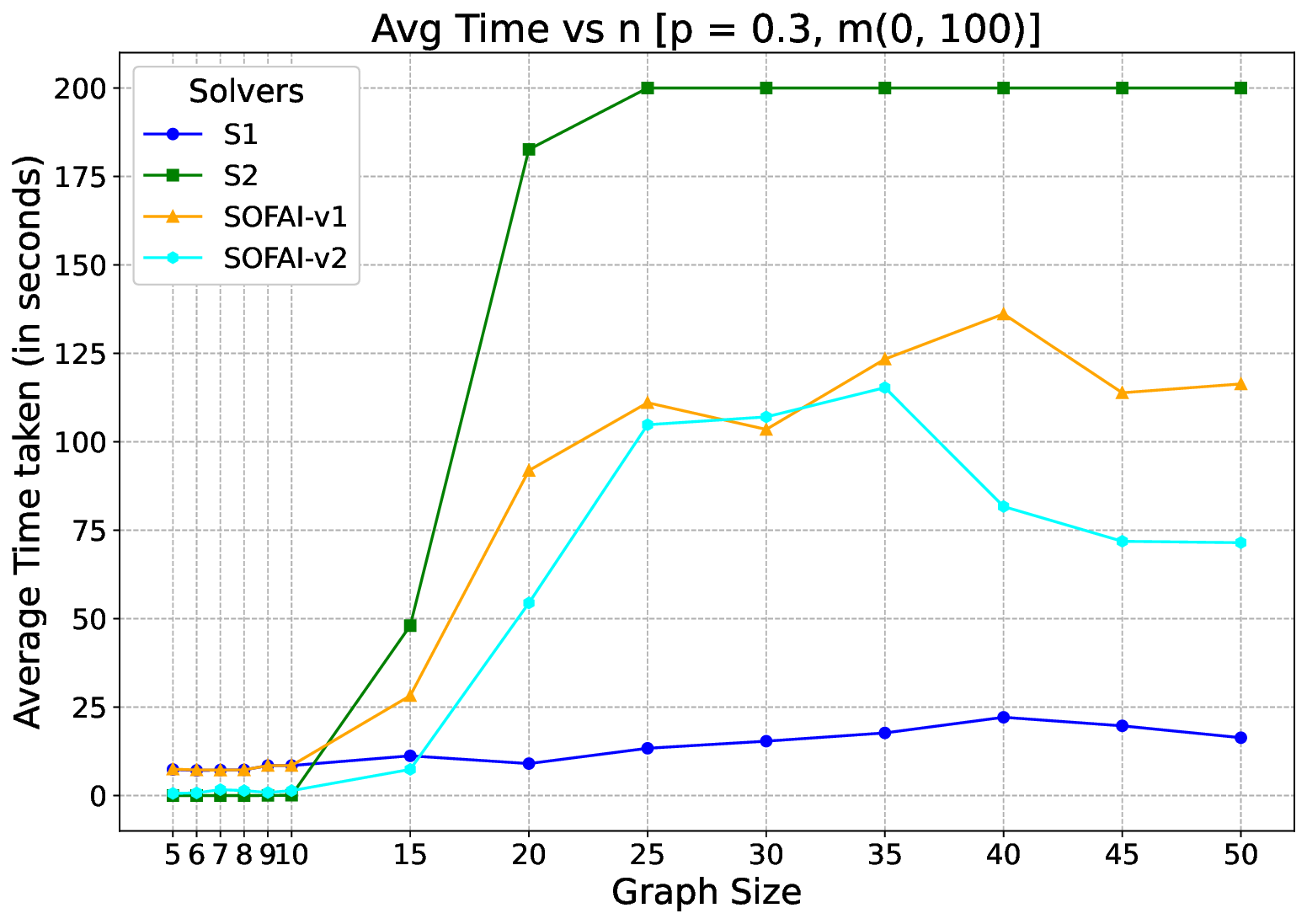}
\label{fig:allhalf}
\end{subfigure}
\begin{subfigure}[b]{0.32\textwidth}
\includegraphics[width=\textwidth]{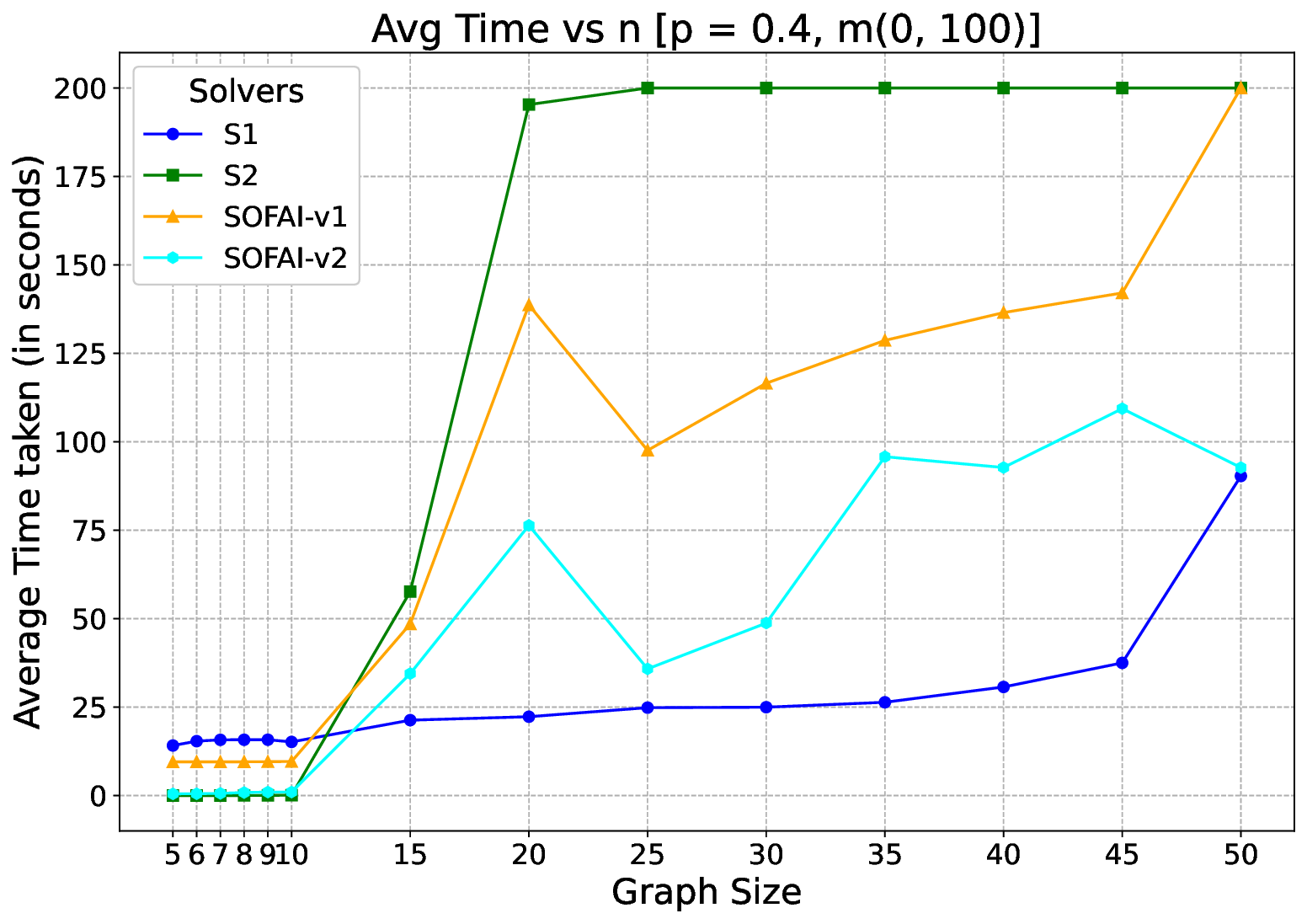}
\label{fig:alls}
\end{subfigure}
\hfill
\begin{subfigure}[b]{0.32\textwidth}
\includegraphics[width=\textwidth]{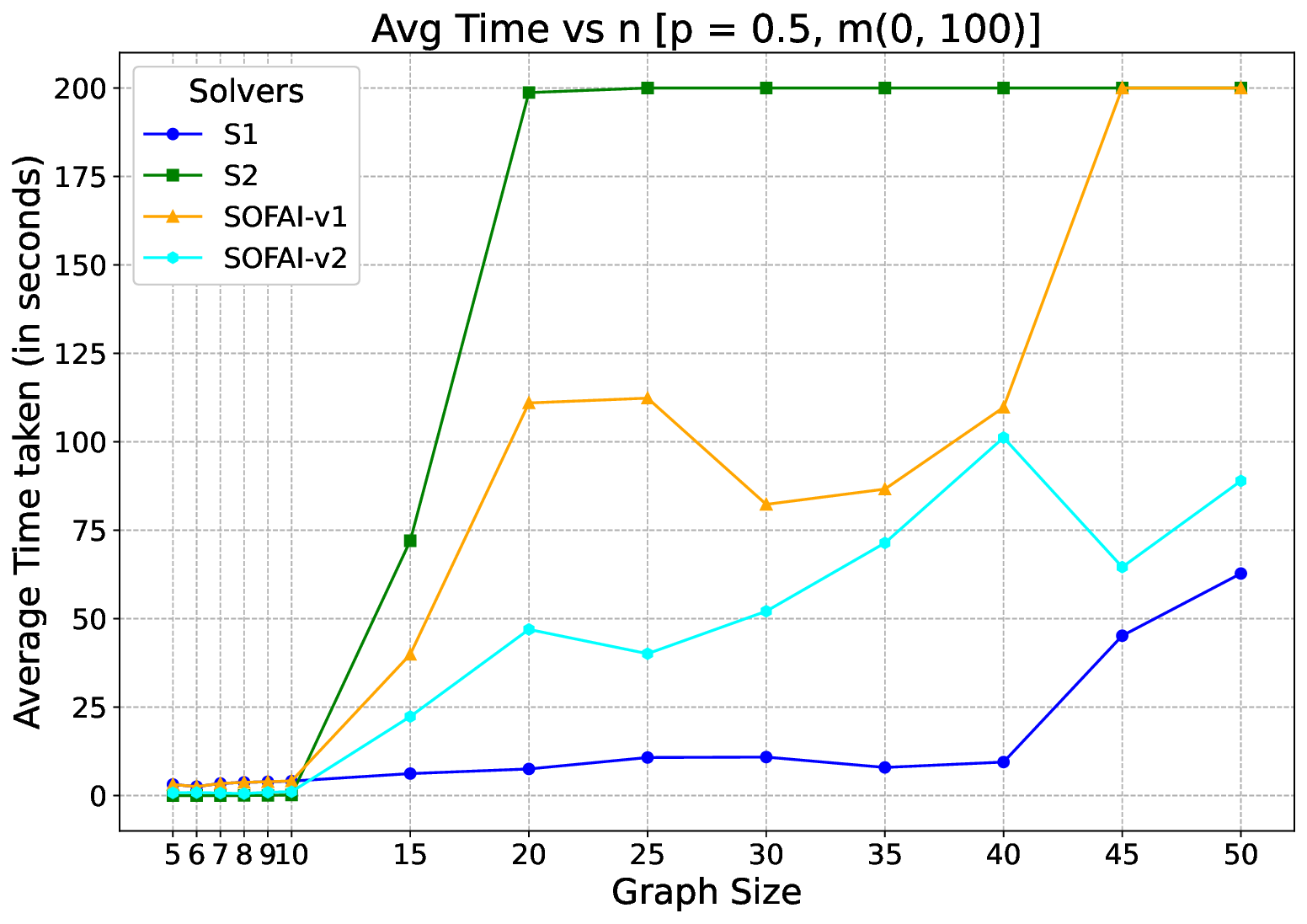}
\label{fig:allus}
\end{subfigure}
\hfill
\begin{subfigure}[b]{0.32\textwidth}
\includegraphics[width=\textwidth]{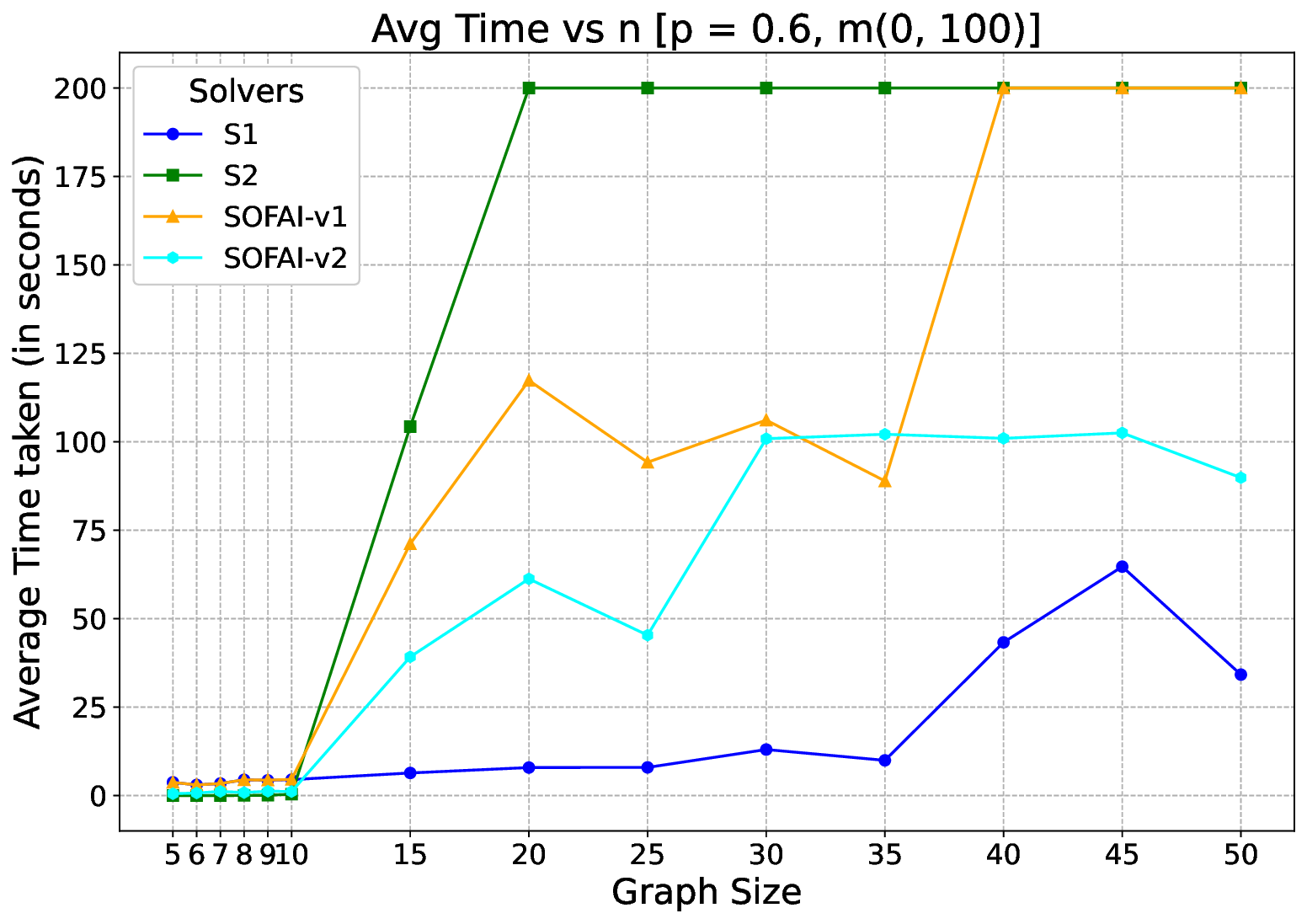}
\label{fig:allhalf}
\end{subfigure}
\begin{subfigure}[b]{0.32\textwidth}
\includegraphics[width=\textwidth]{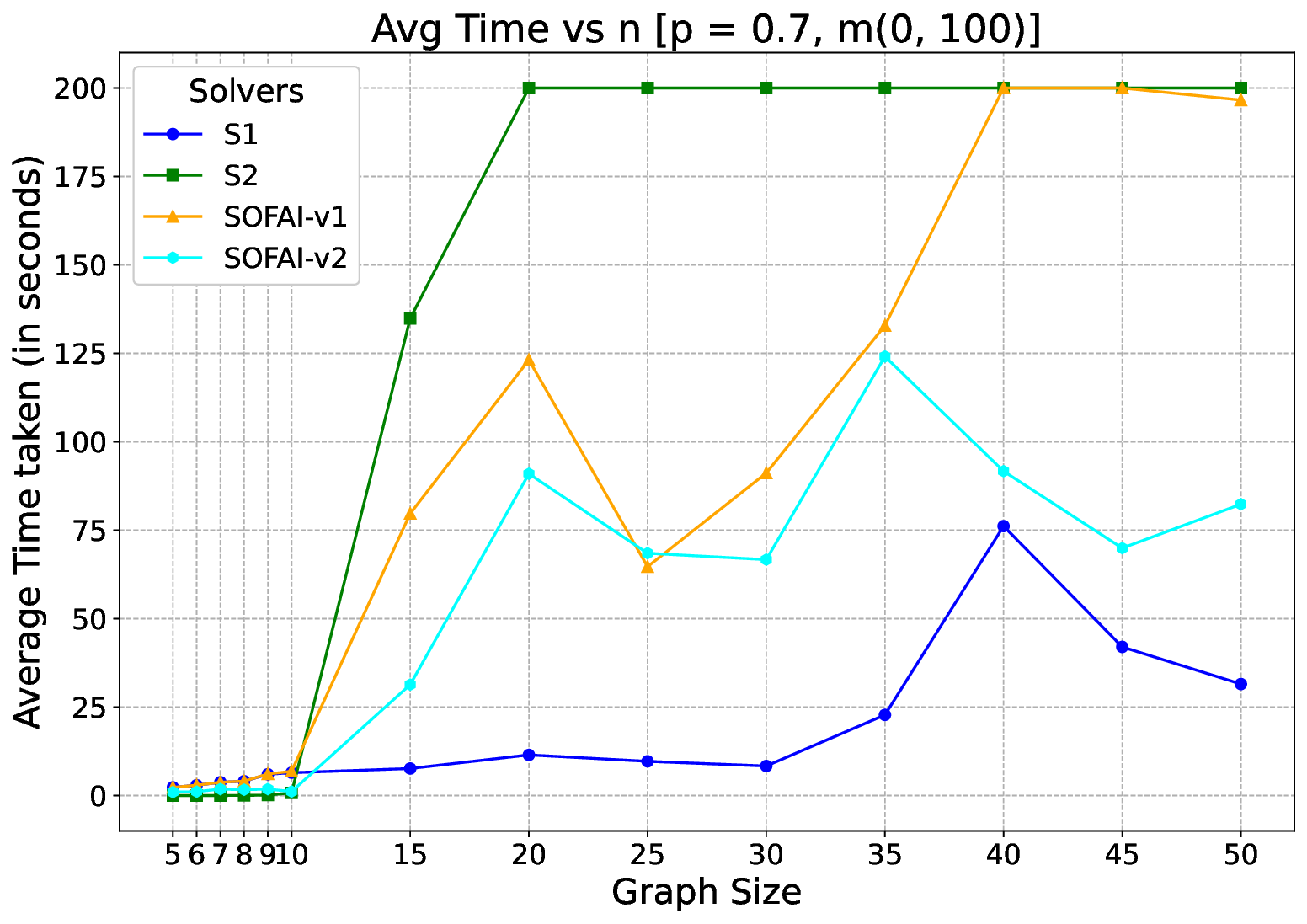}
\label{fig:alls}
\end{subfigure}
\hfill
\begin{subfigure}[b]{0.32\textwidth}
\includegraphics[width=\textwidth]{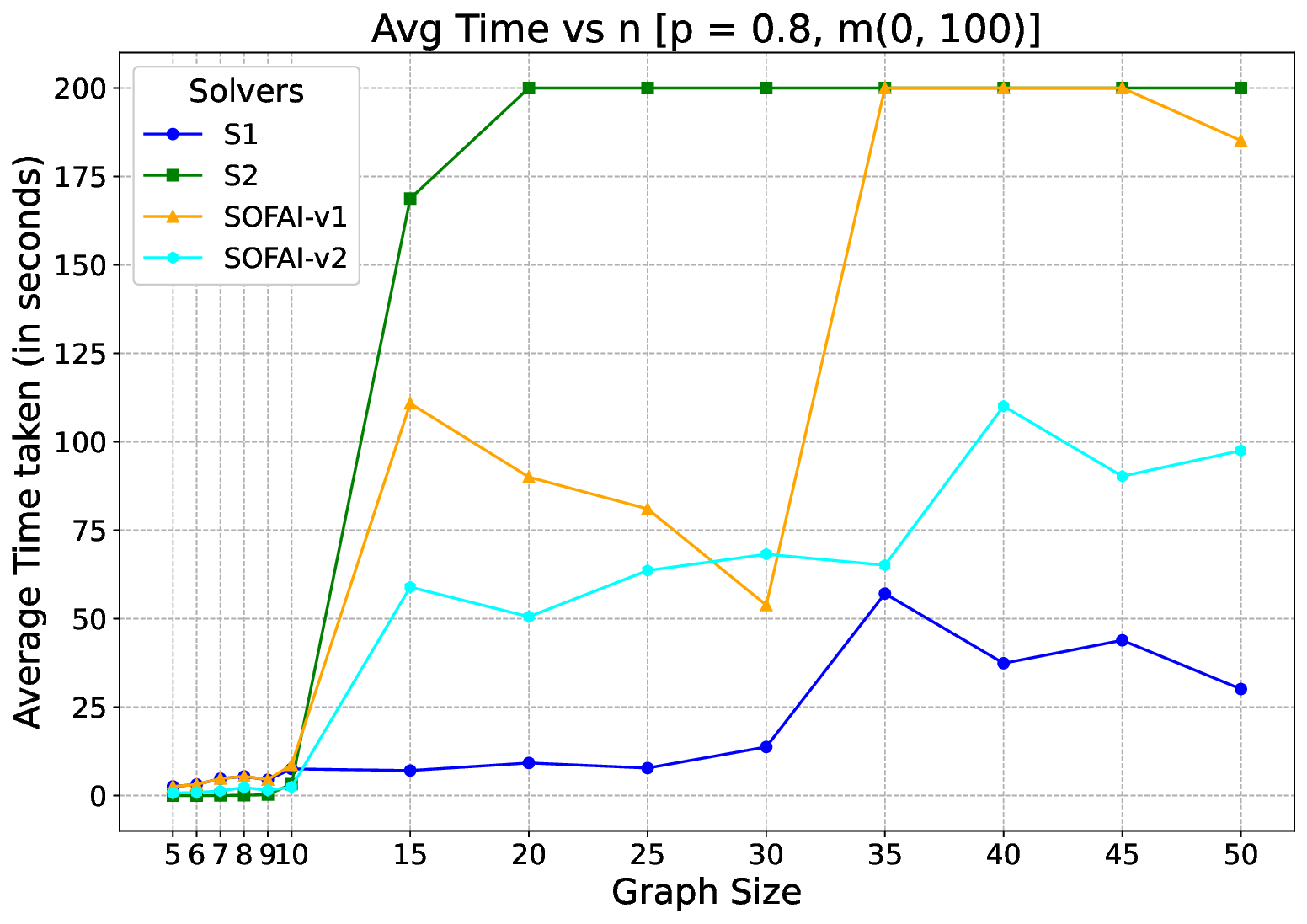}
\label{fig:allus}
\end{subfigure}
\hfill
\begin{subfigure}[b]{0.32\textwidth}
\includegraphics[width=\textwidth]{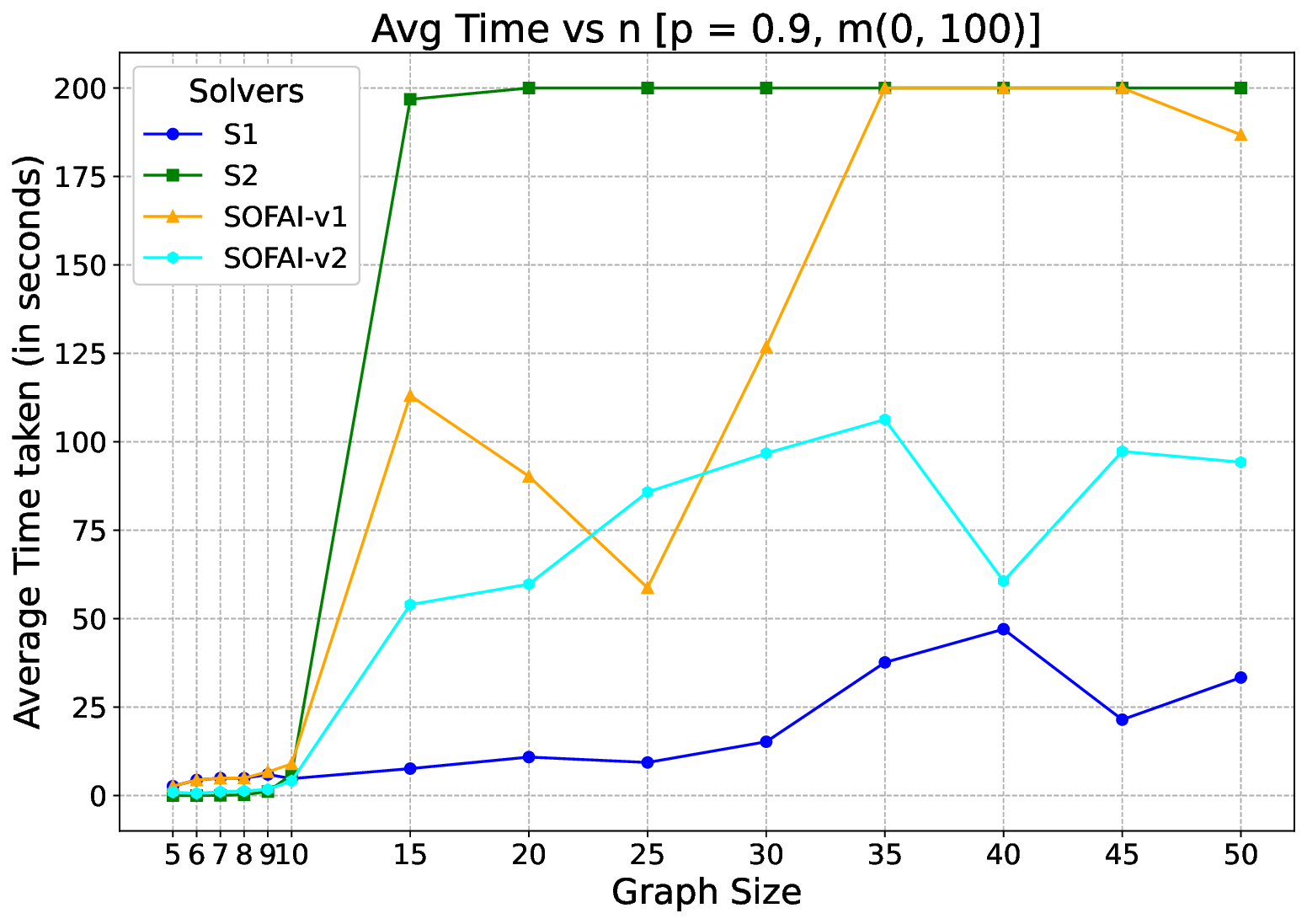}
\label{fig:allhalf}
\end{subfigure}
\caption{Average time of different solvers vs Graph size ($n$) across edge probabilities ($p$) for problem configuration ($m = (0, 100)$)}
\label{fig:t_un}
\end{figure*}

\begin{figure*}[!htbp]
\centering
\begin{subfigure}[b]{0.32\textwidth}
\includegraphics[width=\textwidth]{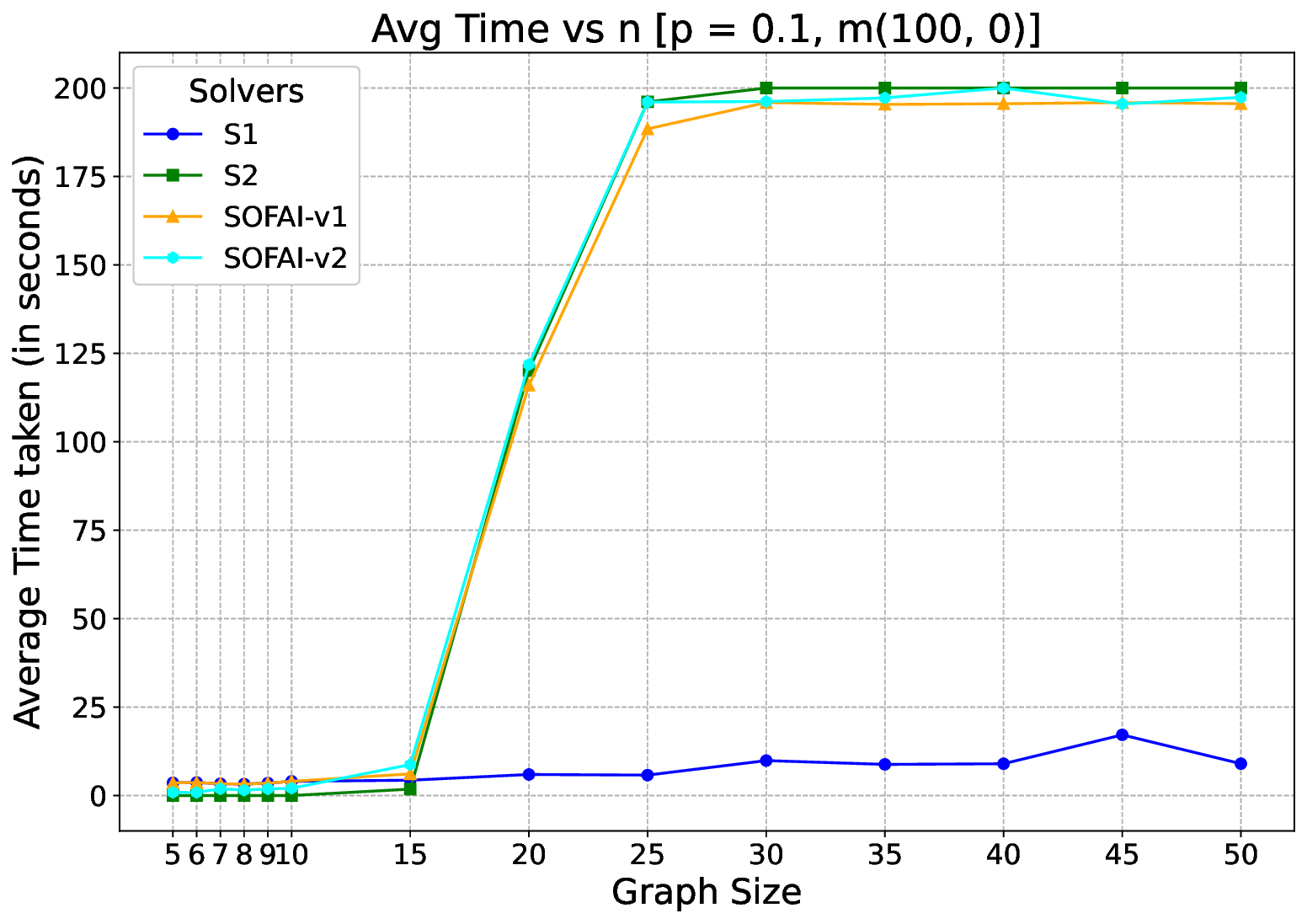}
\label{fig:alls}
\end{subfigure}
\hfill
\begin{subfigure}[b]{0.32\textwidth}
\includegraphics[width=\textwidth]{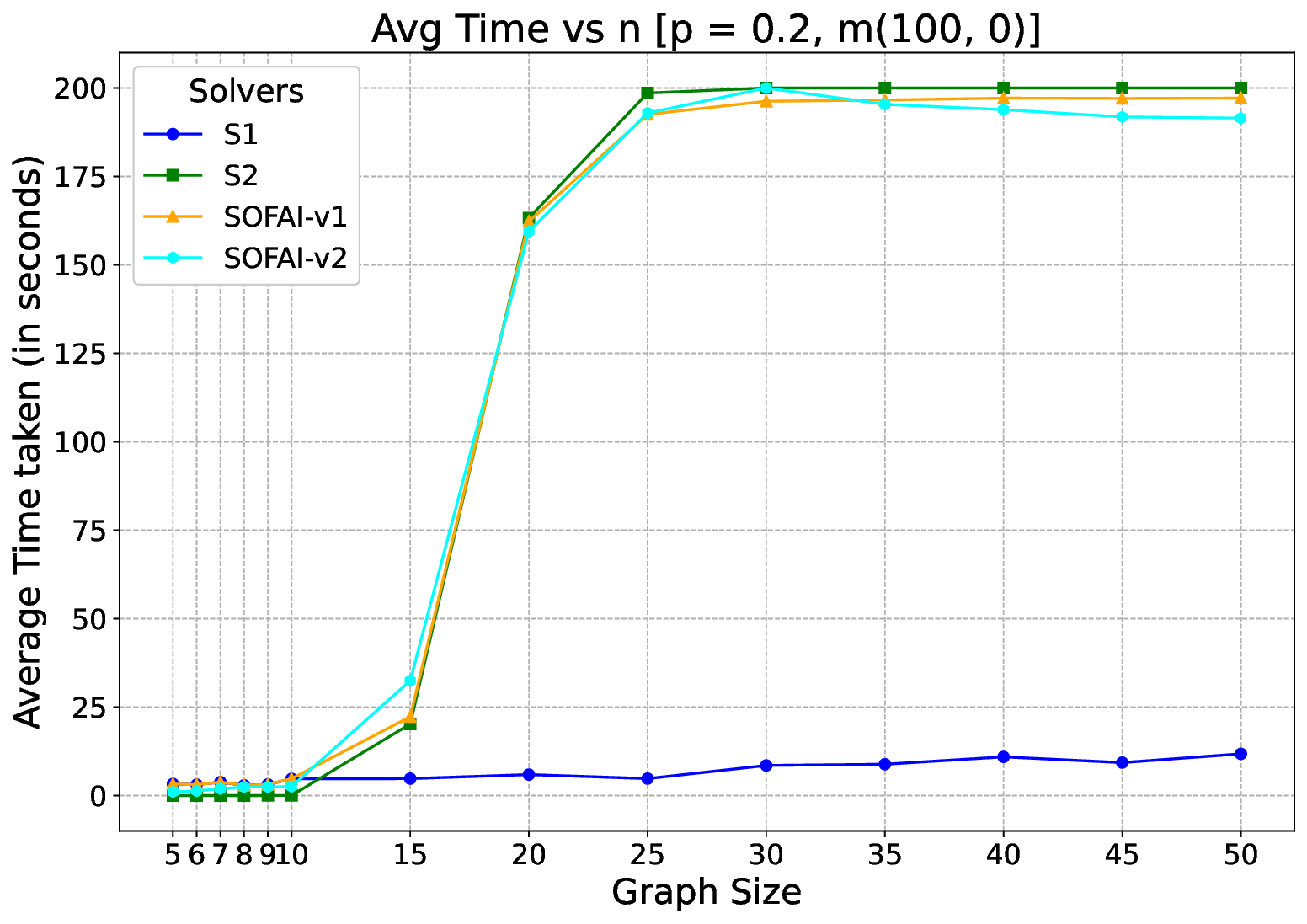}
\label{fig:allus}
\end{subfigure}
\hfill
\begin{subfigure}[b]{0.32\textwidth}
\includegraphics[width=\textwidth]{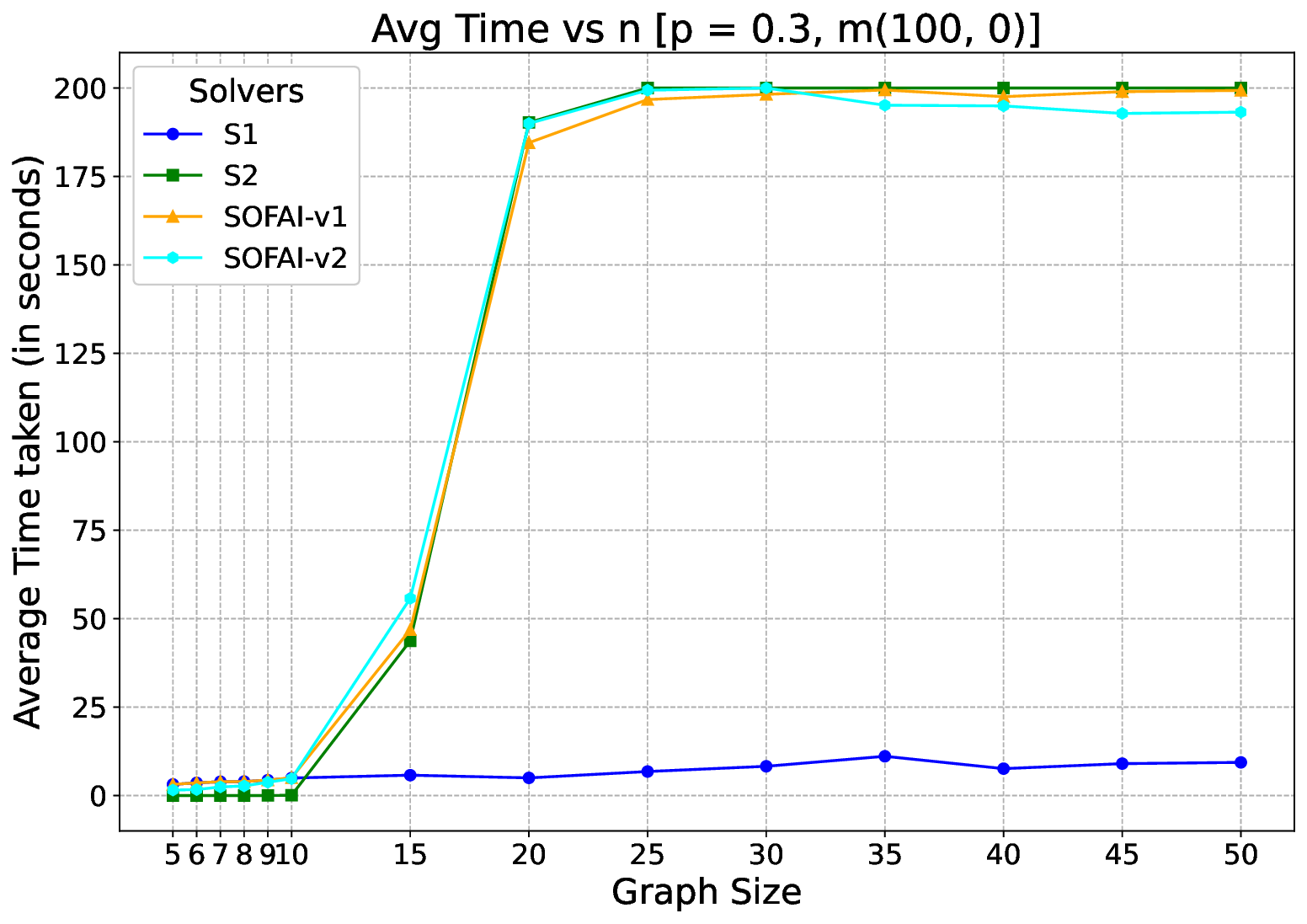}
\label{fig:allhalf}
\end{subfigure}
\begin{subfigure}[b]{0.32\textwidth}
\includegraphics[width=\textwidth]{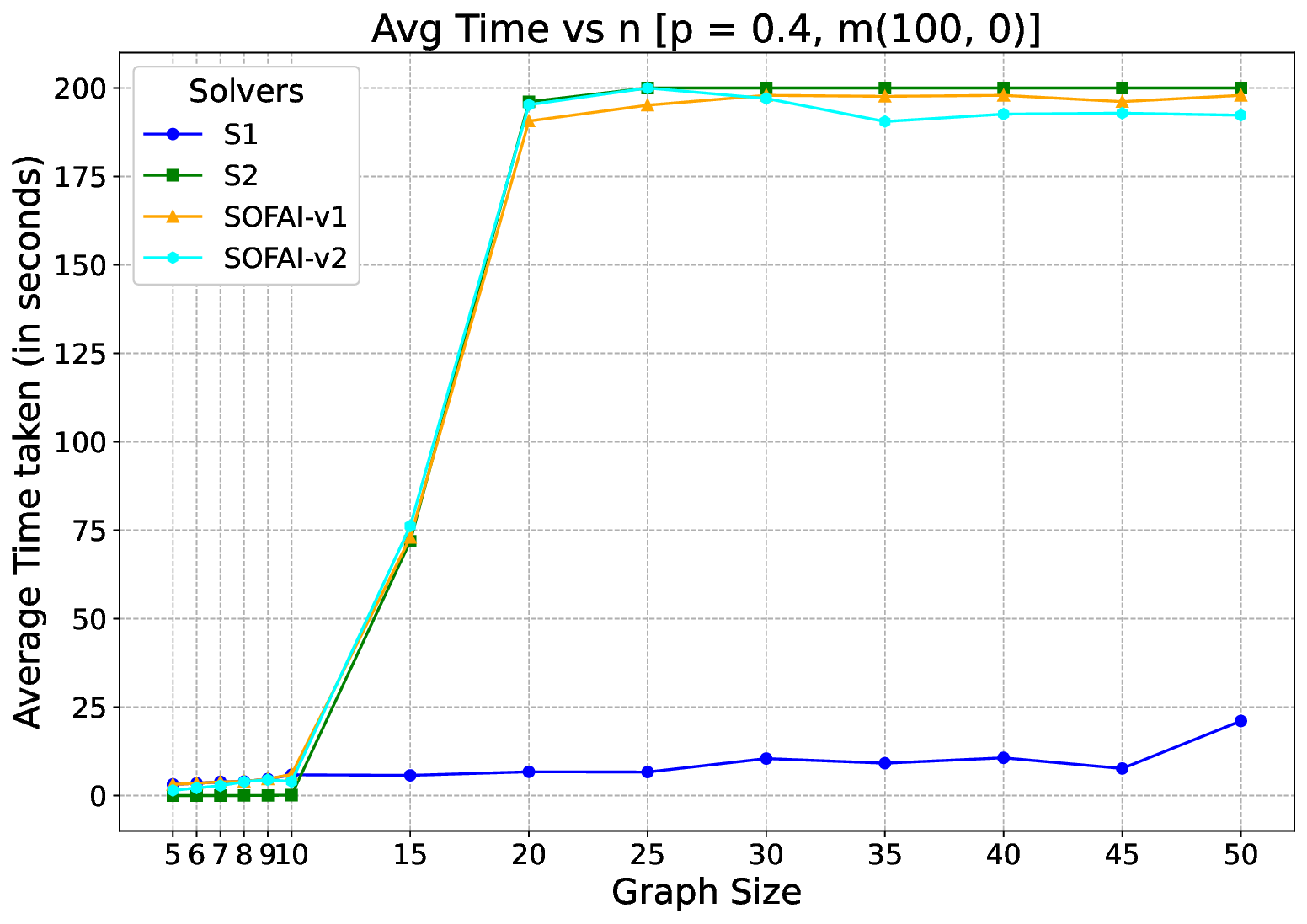}
\label{fig:alls}
\end{subfigure}
\hfill
\begin{subfigure}[b]{0.32\textwidth}
\includegraphics[width=\textwidth]{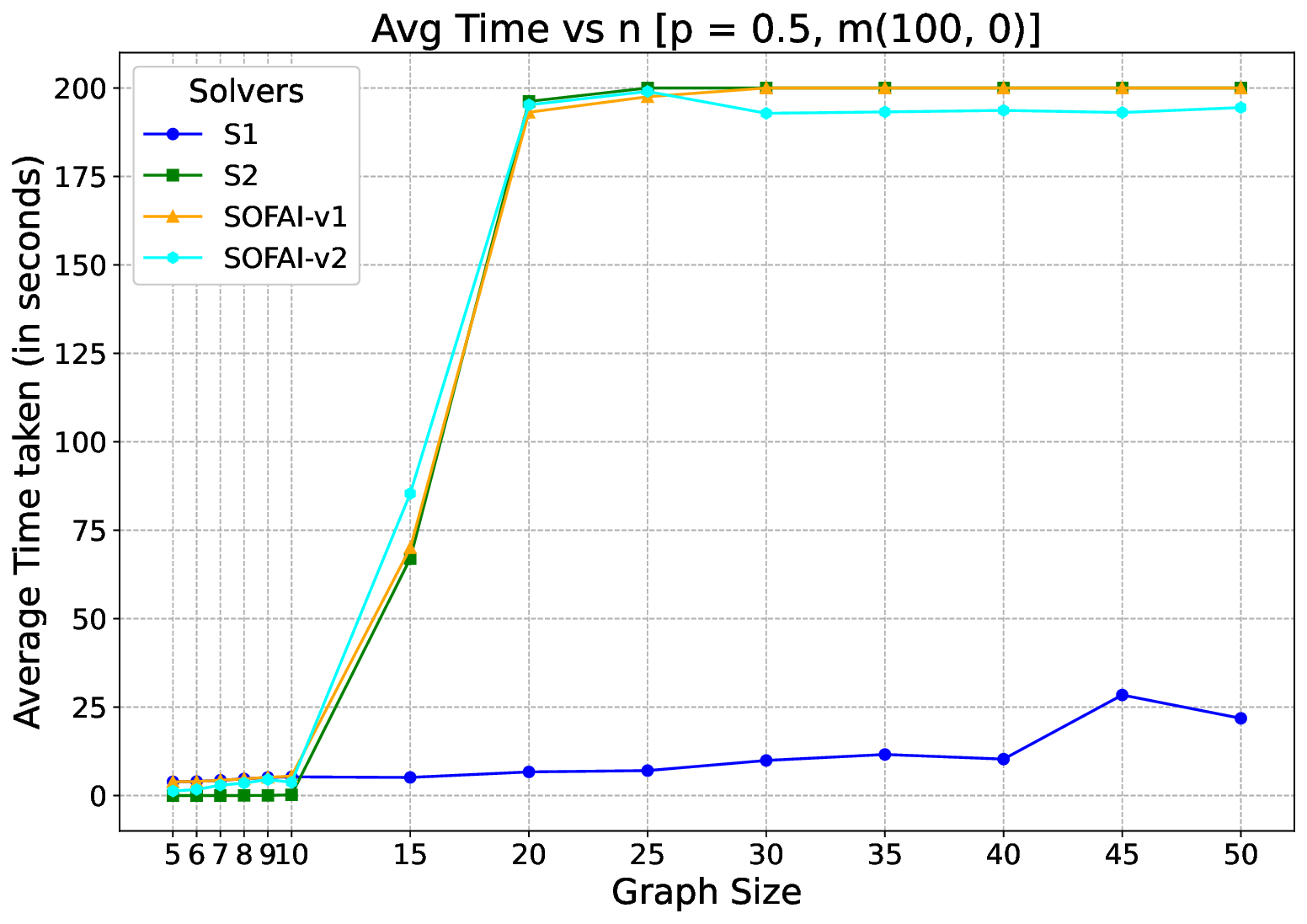}
\label{fig:allus}
\end{subfigure}
\hfill
\begin{subfigure}[b]{0.32\textwidth}
\includegraphics[width=\textwidth]{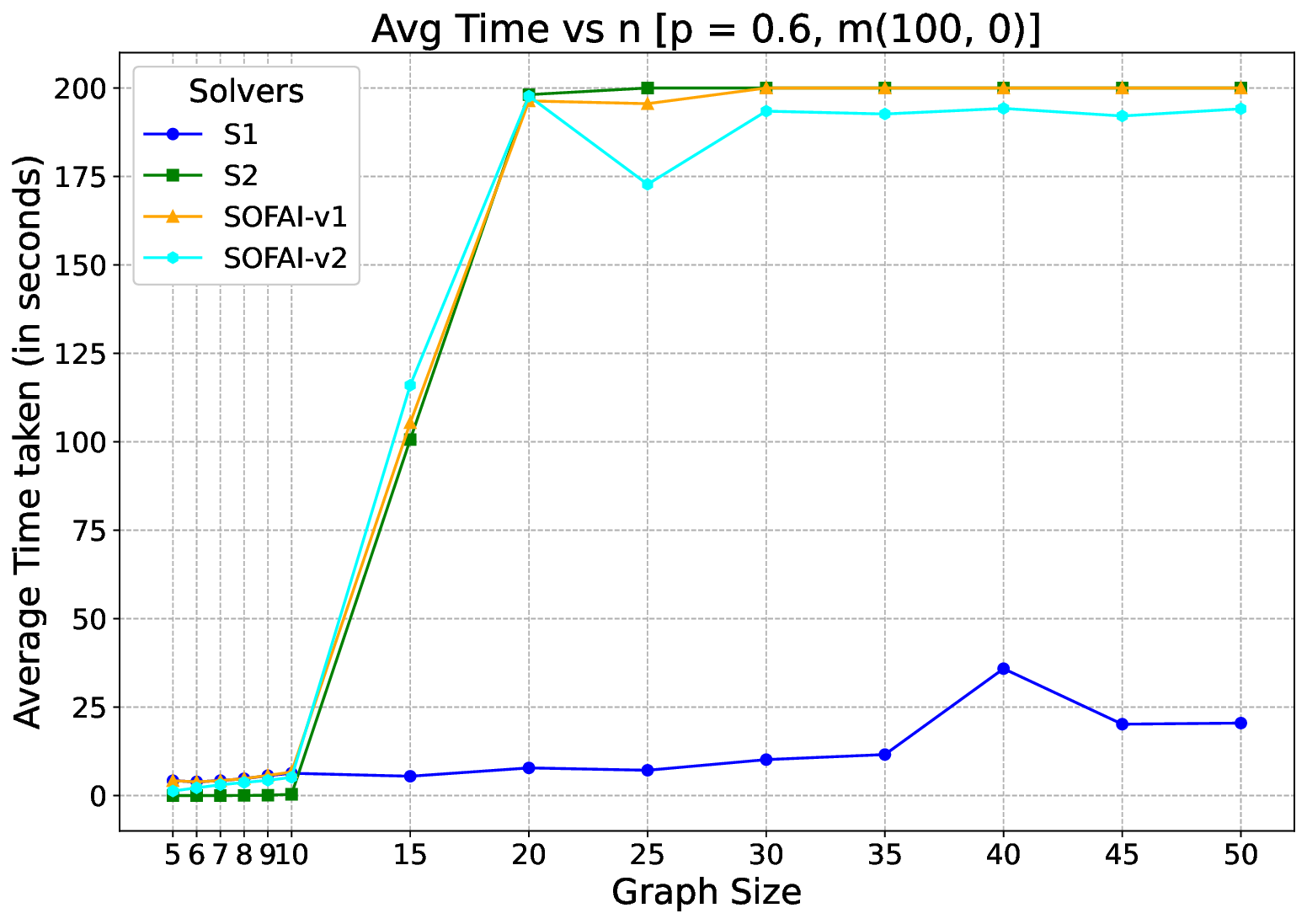}
\label{fig:allhalf}
\end{subfigure}
\begin{subfigure}[b]{0.32\textwidth}
\includegraphics[width=\textwidth]{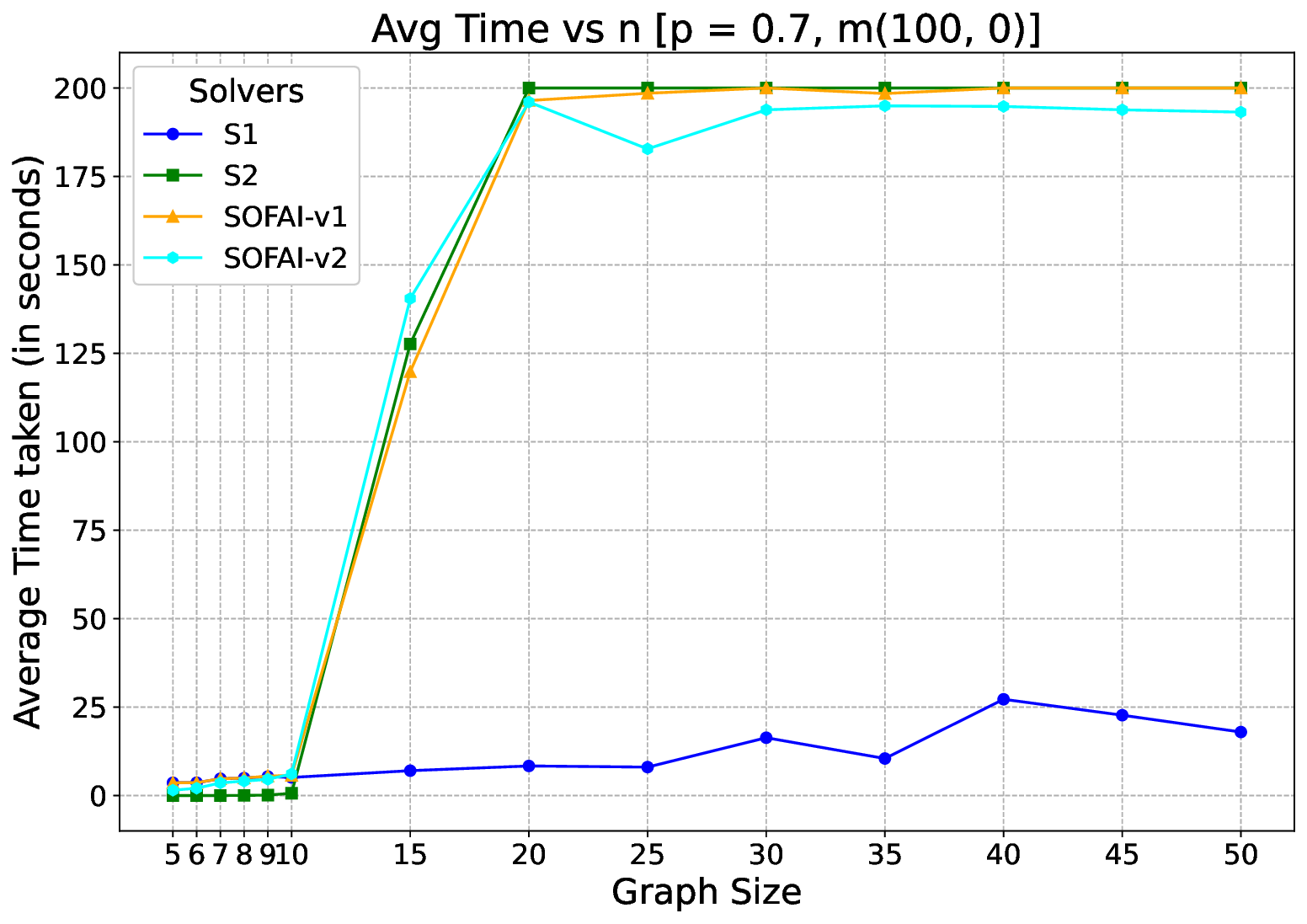}
\label{fig:alls}
\end{subfigure}
\hfill
\begin{subfigure}[b]{0.32\textwidth}
\includegraphics[width=\textwidth]{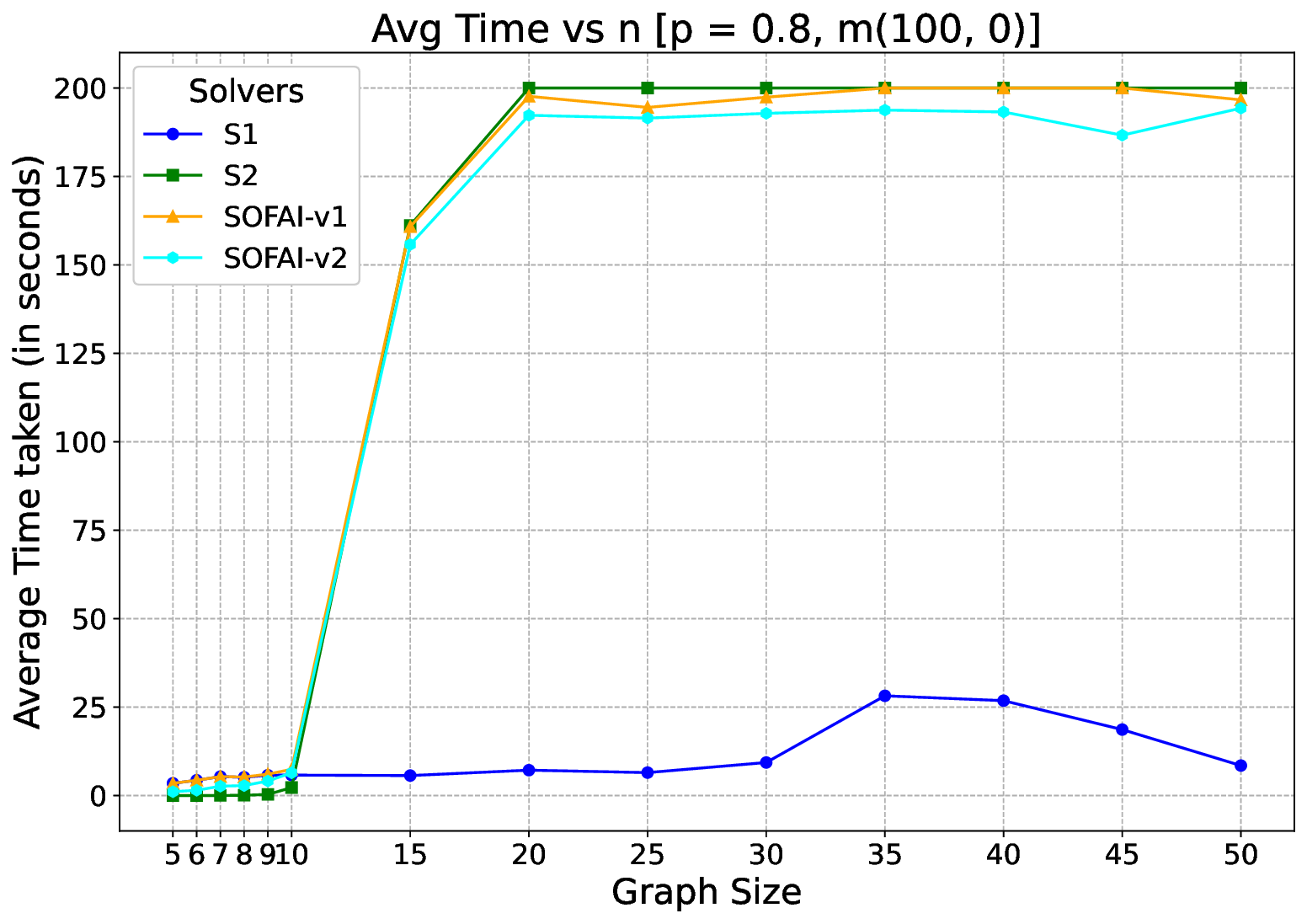}
\label{fig:allus}
\end{subfigure}
\hfill
\begin{subfigure}[b]{0.32\textwidth}
\includegraphics[width=\textwidth]{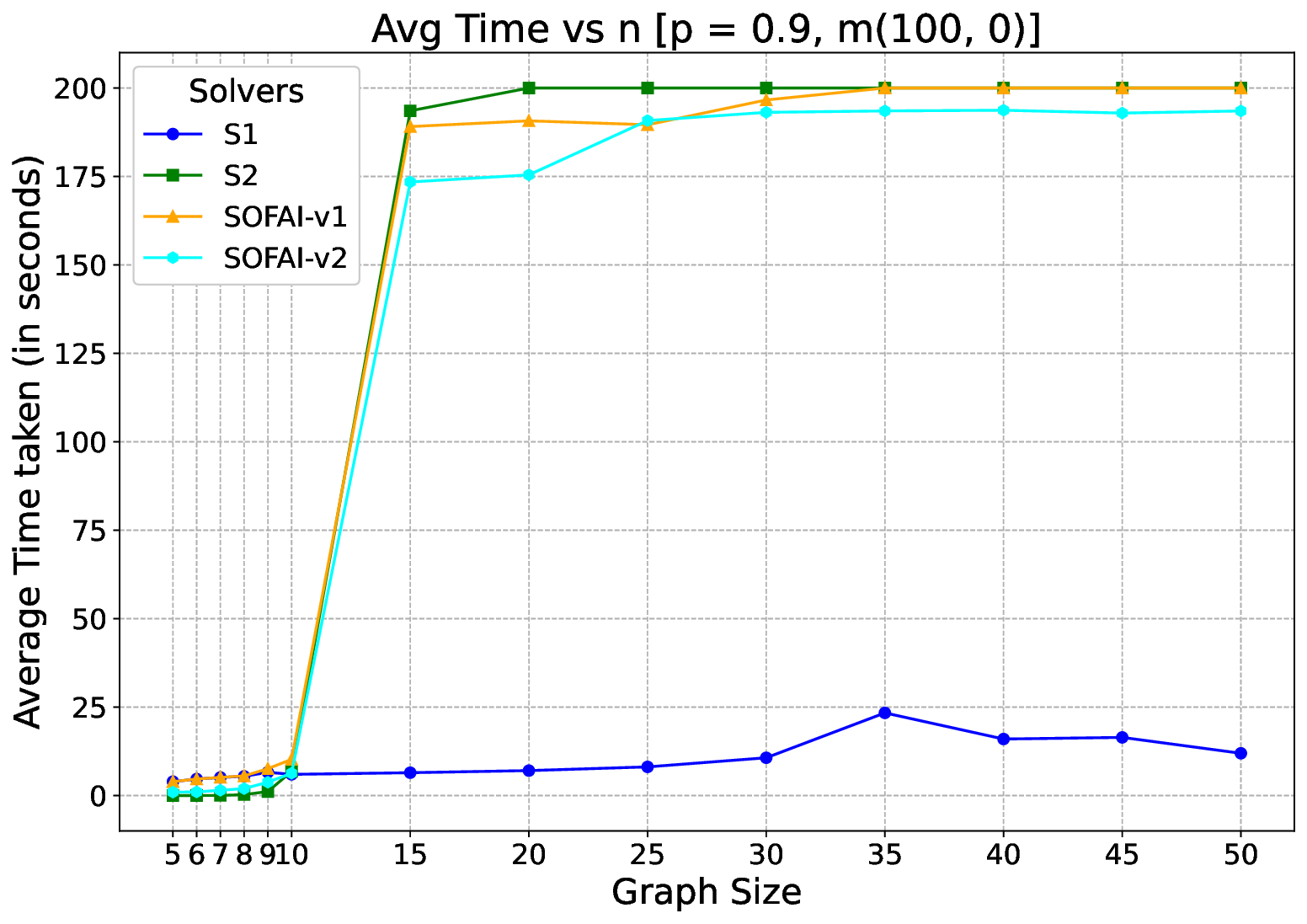}
\label{fig:allhalf}
\end{subfigure}
\caption{Average time of different solvers vs Graph size ($n$) across edge probabilities ($p$) for problem configuration ($m = (100, 0)$)}
\label{fig:t_s}
\end{figure*}

This section explores the time different solvers take to solve graph coloring problems across different problem configurations ($m$). Figure \ref{fig:t_half}, Figure \ref{fig:t_un}, and Figure \ref{fig:t_s} demonstrate how each solver—System 1, System 2, SOFAI-v1, and SOFAI-v2—performs across varying edge probabilities ($p$). These graphs show the superiority in time efficiency of SOFAI-v2 compared to System 2 and SOFAI-v1.

For the above results we set a fixed time limit of 200 seconds per problem. If a solver does not finish within this time, it is counted as a failure.

